%% file: main.tex
\RequirePackage[l2tabu, orthodox]{nag}
\documentclass[12pt]{article}
\usepackage{setspace}
\usepackage{makecell}
\newcolumntype{M}[1]{>{\centering\arraybackslash}m{#1}}
\usepackage[T1]{fontenc}

\usepackage{fouriernc}
\usepackage{amsmath}
\usepackage{amsthm}
\usepackage[scale=0.92]{tgschola}
\usepackage[scaled=0.88]{PTSans}
\usepackage{scalefnt,letltxmacro}
\LetLtxMacro{\oldtextsc}{\textsc}
\renewcommand{\textsc}[1]{\oldtextsc{\scalefont{1.25}#1}}
\usepackage{inconsolata}

\usepackage{amssymb}
\usepackage{mathbbol}
\usepackage{PTSans}
\usepackage{inconsolata}

\usepackage[usenames,dvipsnames]{xcolor}
\definecolor{shadecolor}{gray}{0.9}

\usepackage[english]{babel}
\usepackage[parfill]{parskip}
\usepackage{afterpage}
\usepackage{framed}
\usepackage{nicefrac}

{\endMakeFramed}

\DeclareRobustCommand{\parhead}[1]{\textbf{#1}~}

\usepackage{lineno}

\usepackage{ragged2e}

\usepackage{marginnote}

\newcounter{parcount}

\usepackage{graphicx}
\usepackage[labelfont=bf]{caption}
\usepackage[format=hang]{subcaption}
\usepackage{wrapfig}

\usepackage{booktabs}
\usepackage{multirow}
\usepackage{longtable}

\usepackage{natbib}
\usepackage{bibunits}

\usepackage[algoruled]{algorithm2e}
\usepackage{listings}
\usepackage{fancyvrb}
\fvset{fontsize=\normalsize}

\usepackage[colorlinks,linktoc=all]{hyperref}
\usepackage[all]{hypcap}
\hypersetup{citecolor=Violet}
\hypersetup{linkcolor=black}
\hypersetup{urlcolor=MidnightBlue}

\usepackage[nameinlink]{cleveref}

\usepackage[acronym,smallcaps,nowarn]{glossaries}
\usepackage{listings}
\usepackage{lstbayes}
\lstdefinestyle{mystyle}{
    commentstyle=\color{OliveGreen},
    numberstyle=\tiny\color{black!60},
    stringstyle=\color{BrickRed},
    basicstyle=\ttfamily\scriptsize,
    breakatwhitespace=false,
    breaklines=true,
    captionpos=b,
    keepspaces=true,
    numbers=none,
    numbersep=5pt,
    showspaces=false,
    showstringspaces=false,
    showtabs=false,
    tabsize=2
}
\usepackage{enumitem}

\DeclareMathOperator*{\argmax}{arg\,max}
\DeclareMathOperator*{\argmin}{arg\,min}

\crefname{lemma}{lemma}{lemmas}
\Crefname{lemma}{Lemma}{Lemmas}
\crefname{thm}{theorem}{theorems}
\Crefname{thm}{Theorem}{Theorems}
\crefname{prop}{proposition}{propositions}
\Crefname{prop}{Proposition}{Propositions}

\newtheorem{thm}{Theorem} % reset theorem numbering for each chapter
\newtheorem{defn}[thm]{Definition} % definition numbers are dependent on theorem numbers
\newtheorem{lemma}[thm]{Lemma}
\newtheorem{assumption}{Assumption}
\newtheorem{corollary}[thm]{Corollary}
\newtheorem{prop}[thm]{Proposition}

\renewcommand{\mid}{~\vert~}

\newcommand\dif{\mathop{}\!\mathrm{d}}
\newcommand{\diag}{\textrm{diag}}

\newcommand{\E}{\mathbb{E}}

\newcommand{\cN}{\mathcal{N}}

\newcommand{\g}{\, | \,}
\newcommand{\s}{\, ; \,}

\newacronym{ELBO}{elbo}{evidence lower bound}
\newacronym{GMM}{gmm}{Gaussian mixture model}
\newacronym{KL}{kl}{Kullback-Leibler}
\newacronym{LDA}{lda}{latent Dirichlet allocation}
\newacronym{SVI}{svi}{stochastic variational inference}
\newacronym{VB}{vb}{variational Bayes}
\newacronym{TV}{tv}{total variation}
\newacronym{VBE}{vbe}{variational Bayes estimate}
\newacronym{VFE}{vfe}{variational frequentist estimate}
\newacronym{LAN}{lan}{local asymptotic normality}

\newacronym{MLE}{mle}{maximum likelihood estimate}
\newacronym{MAP}{map}{maximum-a-posterior}
\newacronym{MCMC}{mcmc}{Markov chain Monte Carlo}
\newacronym{EM}{em}{expectation maximization}
\newacronym{LBFGS}{l-bfgs}{limited-memory Broyden-Fletcher-Goldfarb-Shanno}
\newacronym{ADVI}{advi}{automatic differentiation variational inference}
\newacronym{NUTS}{nuts}{No-U-Turn sampler}
\newacronym{HMC}{hmc}{Hamiltonian Monte Carlo}

\newacronym{GLM}{glm}{generalized linear model}
\newacronym{GLMM}{glmm}{generalized linear mixed model}
\newacronym{SBM}{sbm}{stochastic block model}

\newacronym{IF}{if}{influence function}

\newacronym{PF}{pf}{Poisson factorization}

\newacronym[\glsshortpluralkey={rpm}]
{RPM}{rpm}{reweighted probabilistic model}

\newacronym{SNR}{snr}{signal-to-noise ratio}

\usepackage[margin=1in]{geometry}
\usepackage{tcolorbox}

\allowdisplaybreaks
\usepackage[defaultlines=4,all]{nowidow}

\usepackage{times}
\title{\textbf{Frequentist Consistency\\of Variational Bayes}}

\author{
  Yixin Wang\\
  Department of Statistics\\
  Columbia University\\
  \texttt{yixin.wang@columbia.edu} \\
\\
  David M.~Blei\\
  Department of  Statistics\\
  Department of Computer Science\\
  Columbia University\\
  \texttt{david.blei@columbia.edu} \\
}
\date{\today}
\begin{document}
\maketitle

\input{sec_abstract}
Keywords: Bernstein--von Mises, Bayesian inference, variational methods, consistency, asymptotic normality, statistical computing
\clearpage
\begin{bibunit}[alp]
\input{sec_intro}
\input{sec_ideal-vb-posterior}
\input{sec_vbconsistency}
\input{sec_applications}
\input{sec_simulation}
\input{sec_conclusion}

\parhead{Acknowledgements. } We thank the associate editor and two
anonymous reviewers for their constructive comments. We thank Adji
Dieng, Prateek Jaiswal, Christian Naesseth, and Dustin Tran for their
valuable feedback on our manuscript. We also thank Richard Nickl for
pointing us to a key reference. This work is supported by ONR
N00014-11-1-0651, DARPA PPAML FA8750-14-2-0009, the Alfred P. Sloan
Foundation, and the John Simon Guggenheim Foundation.

\clearpage
\putbib[BIB1]
\end{bibunit}

\clearpage
\begin{bibunit}[alp]
{\onecolumn
\input{sec_supp}}
\clearpage
\putbib[BIB1]
\end{bibunit}

\end{document}

%% file: sec_abstract.tex
% !TEX root = main.tex

\begin{abstract}
  A key challenge for modern Bayesian statistics is how to perform
  scalable inference of posterior distributions.  To address this
  challenge, \gls{VB} methods have emerged as a popular alternative to
  the classical \gls{MCMC} methods. \gls{VB} methods tend to be faster
  while achieving comparable predictive performance. However, there
  are few theoretical results around \gls{VB}. In this paper, we
  establish frequentist consistency and asymptotic normality of
  \gls{VB} methods. Specifically, we connect \gls{VB} methods to point
  estimates based on variational approximations, called frequentist
  variational approximations, and we use the connection to prove a
  variational Bernstein--von Mises theorem.  The theorem leverages the
  theoretical characterizations of frequentist variational
  approximations to understand asymptotic properties of \gls{VB}.  In
  summary, we prove that (1) the \gls{VB} posterior converges to the
  \gls{KL} minimizer of a normal distribution, centered at the truth
  and (2) the corresponding variational expectation of the parameter
  is consistent and asymptotically normal. As applications of the
  theorem, we derive asymptotic properties of \gls{VB} posteriors in
  Bayesian mixture models, Bayesian generalized linear mixed models,
  and Bayesian stochastic block models. We conduct a simulation study
  to illustrate these theoretical results.
\end{abstract}

%%% Local Variables:
%%% mode: latex
%%% TeX-master: "main"
%%% End:

%% file: sec_intro.tex
% !TEX root = main.tex
\section{Introduction}
\label{sec:introduction}

\glsreset{VB}
\glsreset{MCMC}

Bayesian modeling is a powerful approach for discovering hidden
patterns in data. We begin by setting up a probability model of latent
variables and observations. We incorporate prior knowledge by setting
priors on latent variables and a functional form of the likelihood.
Finally we infer the posterior, the conditional distribution of the
latent variables given the observations.

For many modern Bayesian models, exact computation of the posterior is
intractable and statisticians must resort to approximate posterior
inference.  For decades, \gls{MCMC}
sampling~\citep{Hastings:1970,gelfand1990sampling,Robert:2004} has
maintained its status as the dominant approach to this problem.
\gls{MCMC} algorithms are easy to use and theoretically sound.  In
recent years, however, data sizes have soared.  This challenges
\gls{MCMC} methods, for which convergence can be slow, and calls upon
scalable alternatives. One popular class of alternatives is \gls{VB}
methods.

To describe \gls{VB}, we introduce notation for the posterior
inference problem.  Consider observations $x = x_{1:n}$.  We posit
local latent variables $z = z_{1:n}$, one per observation, and global
latent variables $\theta = \theta_{1:d}$.  This gives a joint,
\begin{align}
  \label{eq:joint}
    p(\theta, x, z) = p(\theta) \prod_{i=1}^{n} p(z_i \g
    \theta) p(x_i   \g z_i, \theta).
\end{align}
The posterior inference problem is to calculate the posterior
$p(\theta, z \g x)$.

This division of latent variables is common in modern Bayesian
statistics.\footnote{In particular, our results are applicable to
  general models with \textit{local} and \textit{global} latent
  variables~\citep{Hoffman:2013}. The number of local variables $z$
  increases with the sample size $n$; the number of global variables
  $\theta$ does not.  We also note that the conditional independence
  of \Cref{eq:joint} is not necessary for our results.  But we use
  this common setup to simplify the presentation.} In the Bayesian
\gls{GMM}~\citep{roberts1998bayesian}, the component means,
covariances, and mixture proportions are global latent variables; the
mixture assignments of each observation are local latent variables. In
the Bayesian \gls{GLMM}~\citep{breslow1993approximate}, the intercept
and slope are global latent variables; the group-specific random
effects are local latent variables. In the Bayesian
\gls{SBM}~\citep{Wiggins:2008}, the cluster assignment probabilities
and edge probabilities matrix are two sets of global latent variables;
the node-specific cluster assignments are local latent variables. In
the \gls{LDA} model ~\citep{blei2003latent}, the topic-specific word
distributions are global latent variables; the document-specific topic
distributions are local latent variables.  We will study all these
examples below.

\gls{VB} methods formulate posterior inference as an optimization
\citep{jordan1999introduction,wainwright2008graphical,Blei:2016}.  We
consider a family of distributions of the latent variables and then
find the member of that family that is closest to the posterior.

Here we focus on mean-field variational inference (though our results
apply more widely).  First, we posit a family of factorizable
probability distributions on latent variables
\begin{align*}
    \mathcal{Q}^{n+d}
    = \left\{q: q(\theta, z)
    =  \textstyle \prod^d_{i=1}q_{\theta_i}(\theta_i)
    \prod_{j=1}^{n} q_{z_j}(z_j) \right\}.
\end{align*}
This family is called \textit{the mean-field family}.  It represents
a joint of the latent variables with $n+d$ (parametric) marginal 
distributions, $\{q_{\theta_1}, \ldots, q_{\theta_d}, q_{z_1},
\ldots, q_{z_n} \}$.

\gls{VB} finds the member of the family closest to the exact posterior
$p(\theta, z \g x)$, where closeness is measured by \gls{KL}
divergence.  Thus \gls{VB} seeks to solve the optimization,
\begin{align}
    q^*(\theta, z) 
    = \argmin_{q(\theta, z)\in\mathcal{Q}^{n+d}}
    \gls{KL}(q(\theta, z) \, || \, p(\theta, z\mid x)).
\end{align}

In practice, \gls{VB} finds $q^*(\theta, z)$ by optimizing an
alternative objective, \textit{the \gls{ELBO}}, 
\begin{align}
\label{eq:elbo}     
    \gls{ELBO}(q(\theta, z))      
    = -\int q(\theta, z)
    \log\frac{q(\theta,z)}{p(\theta, x, z)}     
    \text{d}\theta \text{d}z.
\end{align} 
This objective is called the \gls{ELBO} because it is a lower bound on
the evidence $\log p(x)$.  More importantly, the \gls{ELBO} is equal
to the negative KL plus $\log p(x)$, which does not depend on
$q(\cdot)$.  Maximizing the \gls{ELBO} minimizes the
\gls{KL}~\citep{jordan1999introduction}.

The optimum $q^*(\theta, z) = q^*(\theta) q^*(z)$ approximates the
posterior, and we call it \textit{the \gls{VB}
  posterior}.\footnote{For simplicity we will write
  $q(\theta, z) = \prod^d_{i=1}q(\theta_i) \prod_{j=1}^{n} q(z_j)$,
  omitting the subscript on the factors $q(\cdot)$. The understanding
  is that the factor is indicated by its argument.}  Though it cannot
capture posterior dependence across latent variables, it has hope to
capture each of their marginals. In particular, this paper is about
the theoretical properties of the \gls{VB} posterior $q^*(\theta)$,
the \gls{VB} posterior of $\theta$.  We will also focus on the
corresponding expectation of the global variable, i.e., an estimate of
the parameter.  It is
\begin{align*}
  \hat{\theta}^*_n:=\int \theta \cdot q^*(\theta) \text{d}\theta.
\end{align*}
We call $\theta^*$ \textit{the \gls{VBE}}.

\gls{VB} methods are fast and yield good predictive performance in
empirical experiments \citep{Blei:2016}. However, there are few
rigorous theoretical results.  In this paper, we prove that (1) the
\gls{VB} posterior converges in \gls{TV} distance to the \gls{KL} 
minimizer of a normal distribution centered at the truth and (2) the
\gls{VBE} is consistent and asymptotically normal.

These theorems are frequentist in the sense that we assume the data
come from $p(x, z \, ;\, \theta_0)$ with a true (nonrandom)
$\theta_0$.  We then study properties of the corresponding posterior
distribution $p(\theta \g x)$, when approximating it with variational
inference.  What this work shows is that the \gls{VB} posterior is
consistent even though the mean field approximating family can be a
brutal approximation.  In this sense, \gls{VB} is a theoretically
sound approximate inference procedure.

\subsection{Main ideas}

We describe the results of the paper.  Along the way, we will need to
define some terms: the \gls{VFE}, the variational log likelihood, the
\gls{VB} posterior, the \gls{VBE}, and the \gls{VB} ideal.
Our results center around the \gls{VB} posterior and the \gls{VBE}.
(\Cref{table:glossary} contains a glossary of terms.)

\parhead{The variational frequentist estimate (\gls{VFE}) and the
  variational log likelihood.} The first idea that we define is the
\textit{variational frequentist estimate} (\gls{VFE}).  It is a point
estimate of $\theta$ that maximizes a local variational objective with
respect to an optimal variational distribution of the local
variables. (The \gls{VFE} treats the variable $\theta$ as a parameter
rather than a random variable.)  We call the objective \textit{the
  variational log likelihood},
\begin{align}
\label{eq:var-loglike}
  M_n(\theta \, ; \, x) =
  \max_{q(z)} \, \, \E_{q(z)}
  \left[ \log p(x, z \g \theta)
  - \log q(z) \right].
\end{align}
In this objective, the optimal variational distribution
$q^{\dagger}(z)$ solves the local variational inference problem,
\begin{align}
\label{eq:var-e}
  q^{\dagger}(z) = \arg \min_{q} \, \textrm{KL}(q(z) \, || \, p(z
  \g x, \theta)).
\end{align}
Note that $q^{\dagger}(z)$ implicitly depends on both the data $x$ and
the parameter $\theta$.

With the objective defined, the \gls{VFE} is
\begin{align}
\label{eq:var-m}
    \hat{\theta}_n = \arg \max_{\theta} \, M_n(\theta \, ; \, x).
\end{align}
It is usually calculated with variational
\gls{EM}~\citep{wainwright2008graphical, ormerod2010explaining}, which
iterates between the E step of \Cref{eq:var-e} and the M step of
\Cref{eq:var-m}.  Recent research has explored the theoretical
properties of the \gls{VFE} for stochastic block models
\citep{bickel2013asymptotic}, generalized linear mixed models
\citep{hall2011asymptotic}, and Gaussian mixture models
\citep{westling2015establishing}.

We make two remarks.  First, the maximizing variational distribution
$q^{\dagger}(z)$ of \Cref{eq:var-e} is different from $q^*(z)$ in the
\gls{VB} posterior: $q^{\dagger}(z)$ is implicitly a function of
individual values of $\theta$, while $q^*(z)$ is implicitly a function
of the variational distributions $q(\theta)$. Second, the variational
log likelihood in \Cref{eq:var-loglike} is similar to the original
objective function for the \gls{EM} algorithm~\citep{Dempster:1977}.
The difference is that the \gls{EM} objective is an expectation with
respect to the exact conditional $p(z \g x)$, whereas the variational
log likelihood uses a variational distribution $q(z)$.

\parhead{Variational Bayes and ideal variational Bayes.} While earlier
applications of variational inference appealed to variational \gls{EM}
and the \gls{VFE}, most modern applications do not.  Rather they use
\gls{VB}, as we described above, where there is a prior on $\theta$
and we approximate its posterior with a global variational
distribution $q(\theta)$.  One advantage of \gls{VB} is that it
provides regularization through the prior.  Another is that it
requires only one type of optimization: the same considerations around
updating the local variational factors $q(z)$ are also at play when
updating the global factor $q(\theta)$.

To develop theoretical properties of \gls{VB}, we connect the
\gls{VB} posterior to the variational log likelihood; this is a
stepping stone to the final analysis. In particular, we define the
\textit{\gls{VB} ideal posterior} $\pi^*(\theta \g x)$,
\begin{align}
\label{eq:ideal-vb}
    \pi^*(\theta \g x) =
    \frac{p(\theta)\exp\{M_n(\theta \s x)\}}
    {\int p(\theta)\exp\{M_n(\theta \s x)\} \text{d}\theta}.
\end{align}
Here the local latent variables $z$ are constrained under the
variational family but the global latent variables $\theta$ are
not. Note that because it depends on the variational log likelihood
$M_n(\theta \s x)$, this distribution implicitly contains an optimal
variational distribution $q^{\dagger}(z)$ for each value of $\theta$;
see \Cref{eq:var-loglike,eq:var-e}.

Loosely, the \gls{VB} ideal lies between the
exact posterior $p(\theta \g x)$ and a variational approximation
$q(\theta)$. It recovers the exact posterior when $p(z \g \theta, x)$
degenerates to a point mass and $q^{\dagger}(z)$ is always equal to
$p(z \g \theta, x)$; in that case the variational likelihood is equal
to the log likelihood and \Cref{eq:ideal-vb} is the posterior. But
$q^{\dagger}(z)$ is usually an approximation to the conditional. Thus
the \gls{VB} ideal usually falls short of the exact
posterior.

That said, the \gls{VB} ideal is more complex that a simple
parametric variational factor $q(\theta)$.  The reason is that its
value for each $\theta$ is defined by the optimization within
$M_n(\theta \s x)$.  Such a distribution will usually lie outside the
distributions attainable with a simple family.

In this work, we first establish the theoretical properties of the
\gls{VB} ideal. We then connect it to the \gls{VB}
posterior.

\parhead{Variational Bernstein--von Mises.}  We have set up the main
concepts.  We now describe the main results.

Suppose the data come from a true (finite-dimensional) parameter
$\theta_0$.  The classical Bernstein--von Mises theorem says that,
under certain conditions, the exact posterior $p(\theta \g x)$
approaches a normal distribution, independent of the prior, as the
number of observations tends to infinity.  In this paper, we extend
the theory around Bernstein--von Mises to the variational posterior.
Here we summarize our results.
\begin{itemize}[leftmargin=*]

\item \Cref{lemma:vbvm} shows that the \gls{VB} ideal
  $\pi^*(\theta \g x)$ is consistent and converges to a normal
  distribution around the \gls{VFE}. If the \gls{VFE} is consistent,
  the \gls{VB} ideal $\pi^*(\theta\mid x)$ converges to a
  normal distribution whose mean parameter is a random vector centered
  at the true parameter.  (Note the randomness in the mean parameter
  is due to the randomness in the observations $x$.)

\item We next consider the point in the variational family that is
  closest to the \gls{VB} ideal $\pi^*(\theta\mid x)$ in
  \gls{KL} divergence.  \Cref{lemma:ivbconsist} and
  \Cref{lemma:ivbnormal} show that this \gls{KL} minimizer is
  consistent and converges to the \gls{KL} minimizer
  of a normal distribution around the
  \gls{VFE}. If the \gls{VFE} is consistent
  \citep{bickel2013asymptotic,hall2011asymptotic} then the \gls{KL}
  minimizer converges to the \gls{KL} minimizer of a normal
  distribution with a random mean centered at the true parameter.

\item \Cref{lemma:vb_post} shows that the \gls{VB} posterior
  $q^*(\theta)$ enjoys the same asymptotic properties as the \gls{KL}
  minimizers of the \gls{VB} ideal
  $\pi^*(\theta \g x)$.

\item \Cref{thm:main} is the \textit{variational Bernstein--von Mises
    theorem}.  It shows that the \gls{VB} posterior $q^*(\theta)$ is
  asymptotically normal around the \gls{VFE}.  Again, if the \gls{VFE}
  is consistent then the \gls{VB} posterior converges to a normal with
  a random mean centered at the true parameter.  Further,
  \Cref{thm:estimate} shows that the \gls{VBE} $\hat{\theta}_n^*$ is
  consistent with the true parameter and asymptotically normal.

\item Finally, we prove two corollaries.  First, if we use a full rank
  Gaussian variational family then the corresponding \gls{VB}
  posterior recovers the true mean and covariance.  Second, if we use
  a mean-field Gaussian variational family then the \gls{VB} posterior
  recovers the true mean and the marginal variance, but not the
  off-diagonal terms. The mean-field \gls{VB} posterior is
  underdispersed.

\end{itemize}

\begin{table*}[t]
\setlength\extrarowheight{10pt}
\centering
\begin{tabular}{ll}
\toprule
    Name & Definition\\
\midrule
    Variational log likelihood &
    $M_n(\theta\s x):=\sup_{q(z)\in\mathcal{Q}^{n}}\int
    q(z)\log\frac{p(x, z\mid\theta)}{q(z)}\text{d}z$\\ 
    Variational frequentist estimate (\gls{VFE})&
    $\hat{\theta}_n := \argmax_\theta M_n(\theta\s x)$\\
    \gls{VB} ideal & 
    $\pi^*(\theta\mid
    x):=\frac{p(\theta)\exp\{M_n(\theta \s x)\}}{\int
    p(\theta)\exp\{M_n(\theta\s x)\}\text{d}\theta}$\\ 
    Evidence Lower Bound (\gls{ELBO})& 
    \gls{ELBO}($q(\theta,z)) :=
    \int \int q(\theta)q(z)\log
    \frac{p(x, z, \theta)}{q(\theta) q(z)} \text{d}\theta
    \text{d}z$\\

    \gls{VB} posterior & 
    $q^*(\theta) :=
    \argmax_{q(\theta)\in\mathcal{Q}^d}\sup_{q(z)\in
    \mathcal{Q}^n}\gls{ELBO}(q(\theta,z))$\\
    \gls{VB} estimate (\gls{VBE}) & 
    $\hat{\theta}_n^*:=\int \theta \cdot
    q^*(\theta)\text{d}\theta$\\
\bottomrule
\end{tabular}
\caption{Glossary of terms\label{table:notation}\label{table:glossary}}
\end{table*}

\glsreset{SBM}
\glsreset{LDA}

\textbf{Related work.} This work draws on two themes. The first is the
body of work on theoretical properties of variational inference.
\citet{you2014variational} and \citet{ormerod2014variational} studied
variational Bayes for a classical Bayesian linear model.  They used
normal priors and spike-and-slab priors on the coefficients,
respectively.  \citet{wang2004convergence} studied variational
Bayesian approximations for exponential family models with missing
values. \citet{wang2005inadequacy} and \citet{wang2006convergence}
analyzed variational Bayes in Bayesian mixture models with conjugate
priors. More recently, \citet{zhang2017theoretical} studied mean field
variational inference in \glspl{SBM} with a batch coordinate ascent
algorithm: it has a linear convergence rate and converges to the
minimax rate within $\log n$ iterations. \citet{sheth2017excess}
proved a bound for the excess Bayes risk using variational inference
in latent Gaussian models. \citet{ghorbani2018instability} studied a
version of \gls{LDA} and identified an instability in variational
inference in certain
\gls{SNR} regimes. \citet{zhang2017convergence} characterized the
convergence rate of variational posteriors for nonparametric and 
high-dimensional inference. \citet{pati2017statistical} provided general
conditions for obtaining optimal risk bounds for point estimates
acquired from mean field variational Bayes.
\citet{alquier2016properties} and \citet{alquier2017concentration}
studied the concentration of variational approximations of Gibbs
posteriors and fractional posteriors based on PAC-Bayesian
inequalities. \citet{yang2017alpha} proposed $\alpha$-variational
inference and developed variational inequalities for the Bayes risk
under the variational solution.

On the frequentist side, \citet{hall2011theory,hall2011asymptotic}
established the consistency of Gaussian variational \gls{EM} estimates
in a Poisson mixed-effects model with a single predictor and a grouped
random intercept. \citet{westling2015establishing} studied the
consistency of variational
\gls{EM} estimates in mixture models through a connection to
M-estimation. \citet{celisse2012consistency} and
\citet{bickel2013asymptotic} proved the asymptotic normality of
parameter estimates in the \gls{SBM} under a mean field variational
approximation.

However, many of these treatments of variational methods---Bayesian or
frequentist---are constrained to specific models and priors. Our work
broadens these works by considering more general models.  Moreover,
the frequentist works focus on estimation procedures under a
variational approximation.  We expand on these works by proving a
variational Bernstein--von Mises theorem, leveraging the frequentist
results to analyze \gls{VB} posteriors.

The second theme is the Bernstein--von Mises theorem. The classical
(parametric) Bernstein--von Mises theorem roughly says that the
posterior distribution of $\sqrt{n}(\theta - \theta_0)$ ``converges'',
under the true parameter value $\theta_0$, to $\cN(X, 1/I(\theta_0))$,
where $X\sim \cN(0, 1/I(\theta_0))$ and $I(\theta_0)$ is the Fisher
information \citep{ghosh2003bayesian,van2000asymptotic,le1953some,
le2012asymptotics}. Early forms of this theorem date back to Laplace,
Bernstein, and von Mises \citep{laplace1809memoire, bernstein,
vonmises}. A version also appeared in \citet{lehmann2006theory}.
\citet{kleijn2012bernstein} established the Bernstein--von Mises
theorem under model misspecification.  Recent advances include
extensions to extensions to semiparametric cases
\citep{murphy2000profile, kim2006bernstein, de2009bernstein,
rivoirard2012bernstein, bickel2012semiparametric,
castillo2012semiparametric, castillo2012semiparametric2,
castillo2014bernstein, panov2014critical, castillo2015bernstein,
ghosal2017fundamentals} and nonparametric cases
\citep{cox1993analysis, diaconis1986consistency,
diaconis1997consistency, diaconis1998consistency, freedman1999wald,
kim2004bernstein, boucheron2009bernstein, james2008large,
johnstone2010high, bontemps2011bernstein, kim2009bernstein,
knapik2011bayesian, leahu2011bernstein, rivoirard2012bernstein,
castillo2012nonparametric, castillo2013nonparametric,
spokoiny2013bernstein, castillo2014bayesian, castillo2014bernstein,
ray2014adaptive, panov2015finite, lu2017bernstein}. In particular,
\citet{lu2016gaussapprox} proved a Bernstein--von Mises type result for
Bayesian inverse problems, characterizing Gaussian approximations of
probability measures with respect to the \gls{KL} divergence. Below,
we borrow proof techniques from \citet{lu2016gaussapprox}. But we move
beyond the Gaussian approximation to establish the consistency of
variational Bayes.

\parhead{This paper.} The rest of the paper is organized as follows.
\Cref{sec:ideal-vb-posterior} characterizes theoretical properties of
the \gls{VB} ideal.  \Cref{sec:vbconsistency} contains the central
results of the paper.  It first connects the \gls{VB} ideal and the
\gls{VB} posterior. It then proves the variational Bernstein--von
Mises theorem, which characterizes the asymptotic properties of the
\gls{VB} posterior and \gls{VB} estimate. \Cref{sec:app} studies three
models under this theoretical lens, illustrating how to establish
consistency and asymptotic normality of specific \gls{VB} estimates.
\Cref{sec:simulation} reports simulation studies to illustrate these
theoretical results. Finally, \Cref{sec:discussion} concludes with
paper with a discussion.

%%% Local Variables:
%%% mode: latex
%%% TeX-master: "main"
%%% End:

%% file: sec_ideal-vb-posterior.tex
% !TEX root = main.tex
\section{The \gls{VB} ideal}
\label{sec:ideal-vb-posterior}

To study the \gls{VB} posterior $q^*(\theta)$, we first study the
\gls{VB} ideal of \Cref{eq:ideal-vb}.  In the next section
we connect it to the \gls{VB} posterior.

Recall the \gls{VB} ideal is
\begin{align*}
\pi^*(\theta\mid x) = \frac{p(\theta)\exp(M_n(\theta \s x))}{\int
p(\theta)\exp(M_n(\theta \s x))\dif \theta},
\end{align*}
where $M_n(\theta \s x)$ is the variational log likelihood of
\Cref{eq:var-loglike}.  If we embed the variational log likelihood
$M_n(\theta \s x)$ in a statistical model of $x$, this model has
likelihood
\begin{align*}
\ell(\theta\s x) \propto \exp(M_n(\theta;x)).
\end{align*}
We call it the \textit{frequentist variational model}. The \gls{VB}
ideal $\pi^*(\theta \mid x)$ is thus the classical posterior under
the frequentist variational model $\ell(\theta\s x)$; the \gls{VFE}
is the classical \gls{MLE}.

Consider the results around frequentist estimation of $\theta$ under
variational approximations of the local variables $z$
\citep{bickel2013asymptotic,hall2011asymptotic,westling2015establishing}.
These works consider asymptotic properties of estimators that maximize
$M_n(\theta \s x)$ with respect to $\theta$.  We will first leverage
these results to prove properties of the \gls{VB} ideal and
their \gls{KL} minimizers in the mean field variational family
$\mathcal{Q}^d$. Then we will use these properties to study the
\gls{VB} posterior, which is what is estimated in practice.

This section relies on the consistent testability and the \gls{LAN} of
$M_n(\theta\s x)$ (defined later) to show the \gls{VB} ideal
is consistent and asymptotically normal. We will then show that its
\gls{KL} minimizer in the mean field family is also consistent and
converges to the \gls{KL} minimizer of a normal distribution in
\gls{TV} distance.

These results are not surprising. Suppose the variational log
likelihood behaves similarly to the true log likelihood, i.e., they
produce consistent parameter estimates. Then, in the spirit of the
classical Bernstein--von Mises theorem under model misspecification
\citep{kleijn2012bernstein}, we expect the \gls{VB} ideal to
be consistent as well. Moreover, the approximation through a
factorizable variational family should not ruin this consistency---
point masses are factorizable and thus the limiting distribution lies
in the approximating family.

\subsection{The \gls{VB} ideal}

The lemma statements and proofs adapt ideas from
\citet{ghosh2003bayesian,
  van2000asymptotic,bickel1967asymptotically,kleijn2012bernstein,
  lu2016gaussapprox} to the variational log likelihood.  Let $\Theta$
be an open subset of $\mathbb{R}^d$. Suppose the observations
$x = x_{1:n}$ are a random sample from the measure $P_{\theta_0}$ with
density $\int p(x, z\mid \theta = \theta_0)\dif z$ for some fixed,
nonrandom value $\theta_0\in\Theta$. $z = z_{1:n}$ are local latent
variables, and $\theta = \theta_{1:d}\in\Theta$ are global latent
variables.  We assume that the density maps
$(\theta, x)\mapsto \int p(x, z\mid \theta)\dif z$ of the true model
and $(\theta, x)\mapsto \ell(\theta\s x)$ of the variational
frequentist models are measurable. For simplicity, we also assume that
for each $n$ there exists a single measure that dominates all measures
with densities $\ell(\theta\s x), \theta \in \Theta$ as well as the
true measure
$P_{\theta_0}$.\\

\begin{assumption}
\label{assumption:classicbvm}
We assume the following conditions for the rest of the paper:
\begin{enumerate}
    \item (Prior mass) The prior measure with Lebesgue-density
    $p(\theta)$ on $\Theta$ is continuous and positive on a neighborhood
    of $\theta_0$. There exists a constant $M_p>0$ such that $|(\log
    p(\theta))''|\leq M_pe^{|\theta|^2}$.

    \item (Consistent testability) For every $\epsilon > 0$ there exists
    a sequence of tests $\phi_n$ such that
    \[
        \int \phi_n(x)p(x, z\mid \theta_0)
        \emph{d}z\emph{d}x
        \rightarrow 0
    \]
    and
    \[
        \sup_{\theta:||\theta -\theta_0||\geq \epsilon} 
        \int (1-\phi_n(x))
        \frac{\ell(\theta\s x)}{\ell(\theta_0\s x)}
        p(x, z\mid \theta_0)
        \emph{d}z\emph{d}x
        \rightarrow 0,
    \]

    \item (Local asymptotic normality (\gls{LAN})) For every compact set
    $K\subset
    \mathbb{R}^d$, there exist random vectors $\Delta_{n,\theta_0}$ bounded
    in probability and nonsingular matrices $V_{\theta_0}$ such that
    \[
        \sup_{h\in K}
        |M_n(\theta + \delta_nh\s x) - M_n(\theta\s x) 
        - h^\top V_{\theta_0}\Delta_{n,\theta_0} 
        + \frac{1}{2}h^\top V_{\theta_0}h|
        \stackrel{P_{\theta_0}}{\rightarrow} 0,
    \]
    where $\delta_n$ is a $d\times d$ diagonal matrix. We have
    $\delta_n\rightarrow 0$ as $n \rightarrow \infty$. For $d = 1$, we
    commonly have $\delta_n = 1/\sqrt{n}$.\\
\end{enumerate}
\end{assumption}
These three assumptions are standard for Bernstein--von Mises theorem.
The first assumption is a prior mass assumption. It says the prior on
$\theta$ puts enough mass to sufficiently small balls around
$\theta_0$. This allows for optimal rates of convergence of the
posterior. The first assumption further bounds the second derivative
of the log prior density. This is a mild technical assumption
satisfied by most non-heavy-tailed distributions.

The second assumption is a consistent testability assumption. It says
there exists a sequence of uniformly consistent (under
$P_{\theta_0}$) tests for testing $H_0:\theta = \theta_0$ against
$H_1:||\theta - \theta_0||\geq\epsilon$ for every $\epsilon > 0$
based on the frequentist variational model. This is a weak
assumption. For example, it suffices to have a compact $\Theta$ and
continuous and identifiable $M_n(\theta\s x)$. It is also true when
there exists a consistent estimator $T_n$ of $\theta$. In this case,
we can set $\phi_n:=1\{T_n-\theta \geq \epsilon/2\}.$

The last assumption is a local asymptotic normality assumption on
$M_n(\theta\s x)$ around the true value $\theta_0$. It says the
frequentist variational model can be asymptotically approximated by a
normal location model centered at $\theta_0$ after a rescaling of
$\delta_n^{-1}$. This normalizing sequence $\delta_n$ determines the
optimal rates of convergence of the posterior. For example, if
$\delta_n = 1/\sqrt{n}$, then we commonly have $\theta - \theta_0 =
O_p(1/\sqrt{n})$. We often need model-specific analysis to verify
this condition, as we do in \Cref{sec:app}. We discuss sufficient
conditions and general proof strategies in \Cref{subsec:lan}.

In the spirit of the last assumption, we perform a change-of-variable
step:
\begin{align}
  \label{eq:transformed-theta}
\tilde{\theta} = \delta_n^{-1} (\theta - \theta_0).
\end{align} We center $\theta$ at the true value $\theta_0$ and
rescale it by the reciprocal of the rate of convergence
$\delta_n^{-1}.$ This ensures that the asymptotic distribution of
$\tilde{\theta}$ is not degenerate, i.e., it does not converge to a
point mass. We define
$\pi^*_{\tilde{\theta}}(\cdot \mid x)$ as the density of
$\tilde{\theta}$ when $\theta$ has density $\pi^*(\cdot \mid x)$:
\begin{align*}
    \pi^*_{\tilde{\theta}}(\tilde{\theta}\mid x)
    = 
    \pi^*(\theta_0 + \delta_n\tilde{\theta}\mid x)
    \cdot
    |\det(\delta_n)|.
\end{align*}

Now we characterize the asymptotic properties of the \gls{VB} ideal.\\
\begin{lemma}
\label{lemma:vbvm}

The \gls{VB} ideal converges in total variation to a sequence of normal
distributions,
\[
    ||
    \pi^*_{\tilde{\theta}}(\cdot\mid x)
    - 
    \cN(\cdot;\Delta_{n,\theta_0}, V^{-1}_{\theta_0})
    ||
    _{\gls{TV}}\stackrel{P_{\theta_0}}{\rightarrow} 0.\\
\]
\end{lemma}

\emph{Proof sketch of \cref{lemma:vbvm}.} This is a consequence of
the classical finite-dimensional Bernstein--von Mises theorem under
model misspecification \citep{kleijn2012bernstein}.  Theorem 2.1 of
\citet{kleijn2012bernstein} roughly says that the posterior is
consistent if the model is locally asymptotically normal around the
true parameter value $\theta_0$. Here the true data generating measure
is $P_{\theta_0}$ with density
$\int p(x, z\mid \theta = \theta_0)\dif z$, while the frequentist
variational model has densities $\ell(\theta\s x), \theta\in\Theta$.

What we need to show is that the consistent testability assumption in
\Cref{assumption:classicbvm} implies assumption (2.3) in
\citet{kleijn2012bernstein}:
\[\int_{|\tilde{\theta}|>M_n} \pi^*_{\tilde{\theta}}(\tilde{\theta}\mid x) 
\dif \tilde{\theta}\stackrel{P_{\theta_0}}{\rightarrow} 0\]
for every sequence of constants $M_n \rightarrow \infty.$. To show
this, we mimic the argument of Theorem 3.1 of
\citet{kleijn2012bernstein}, where they show this implication for the
iid case with a common convergence rate for all dimensions of
$\theta$. See \Cref{sec:vbvmproof} for details.\hfill \qedsymbol\\

This lemma says the \gls{VB} ideal of the rescaled $\theta$,
$\tilde{\theta} = \delta_n^{-1}(\theta - \theta_0)$, is asymptotically
normal with mean $\Delta_{n,\theta_0}$. The mean, $\Delta_{n,\theta_0}$,
as assumed in \Cref{assumption:classicbvm}, is a random vector bounded
in probability. The asymptotic distribution
$\cN(\cdot;\Delta_{n,\theta_0}, V^{-1}_{\theta_0})$ is thus also random,
where randomness is due to the data $x$ being random draws from the
true data generating measure $P_{\theta_0}$. We notice that
if the \gls{VFE}, $\hat{\theta}_n$, is consistent and asymptotically
normal, we commonly have
$\Delta_{n,\theta_0} = \delta_n^{-1}(\hat{\theta}_n - \theta_0)$ with
$\mathbb{E}(\Delta_{n,\theta_0}) = 0$. Hence, the \gls{VB} ideal 
will converge to a normal distribution with a random mean
centered at the true value $\theta_0$.

\subsection{The KL minimizer of the \gls{VB} ideal}

Next we study the \gls{KL} minimizer of the \gls{VB} ideal
in the mean field variational family. We show its consistency and
asymptotic normality. To be clear, the asymptotic normality is in the
sense that the \gls{KL} minimizer of the \gls{VB} ideal
converges to the \gls{KL} minimizer of a normal distribution in
\gls{TV}
distance.\\

\begin{lemma}
\label{lemma:ivbconsist}
The \gls{KL} minimizer of the \gls{VB} ideal over the mean
field family is consistent: it
converges weakly to a point mass in $P_{\theta_0}$-probability, 
\[
    \argmin_{q(\theta)\in\mathcal{Q}^d} 
    \gls{KL}(q(\theta)||\pi^*(\theta\mid x))
    \stackrel{d}{\rightarrow} \delta_{\theta_0} \qquad\text{in $P_{\theta_0}$-probability}.\\
\]

\end{lemma}
\emph{Proof sketch of \cref{lemma:ivbconsist}.} The key insight here
is that point masses are factorizable. \Cref{lemma:vbvm} above
suggests that the \gls{VB} ideal converges in distribution
to a point mass. We thus have its \gls{KL} minimizer also
converging to a point mass, because point masses reside within the
mean field family. In other words, there is no loss, in the limit,
incurred by positing a factorizable variational family for
approximation.

To prove this lemma, we bound the mass of $B^c(\theta_0, \eta_n)$
under $q(\theta)$, where $B^c(\theta_0, \eta_n)$ is the complement of
an $\eta_n$-sized ball centered at $\theta_0$ with $\eta_n\rightarrow
0$ as $n\rightarrow\infty$. In this step, we borrow ideas from the
proof of Lemma 3.6 and Lemma 3.7 in \citet{lu2016gaussapprox}.
See \Cref{sec:ivbconsistproof} for details.
\hfill\qedsymbol\\

\begin{lemma}
\label{lemma:ivbnormal}
The \gls{KL} minimizer of the \gls{VB} ideal of $\tilde{\theta}$
converges to that of $\cN(\cdot
\s\Delta_{n,\theta_0}, V^{-1}_{\theta_0})$ in total variation: under
mild technical conditions on the tail behavior of $\mathcal{Q}^d$
(see \Cref{assumption:scoreint} in \Cref{sec:ivbnormalproof}),
\[
    \left\|
    \argmin_{q\in\mathcal{Q}^d} 
    \gls{KL}(q(\cdot)||\pi^*_{\tilde{\theta}}(\cdot\mid x))
    -
    \argmin_{q\in\mathcal{Q}^d} 
    \gls{KL}(q(\cdot)|| \cN(\cdot \s\Delta_{n,\theta_0}, V^{-1}_{\theta_0}))
    \right\|
    _{\gls{TV}}\stackrel{P_{\theta_0}}{\rightarrow} 0.\\
\]

\end{lemma}

\emph{Proof sketch of \cref{lemma:ivbnormal}.} The intuition here is
that, if the two distribution are close in the limit, their \gls{KL}
minimizers should also be close in the limit.
\Cref{lemma:vbvm} says that the \gls{VB} ideal of
$\tilde{\theta}$ converges to $\cN(\cdot;\Delta_{n,\theta_0},
V^{-1}_{\theta_0})$ in total variation. We would expect their
\gls{KL} minimizer also converges in some metric. This
result is also true for the (full-rank) Gaussian variational family if
rescaled appropriately.

Here we show their convergence in total variation. This is achieved
by showing the $\Gamma$-convergence of the functionals of $q$:
$\gls{KL}(q(\cdot)||\pi^*_{\tilde{\theta}}(\cdot\mid x))$ to 
$\gls{KL}(q(\cdot)|| 
\cN(\cdot \s\Delta_{n,\theta_0}, V^{-1}_{\theta_0}))$, 
for parametric $q$'s. $\Gamma$-convergence is a classical tool for
characterizing variational problems; $\Gamma$-convergence of
functionals ensures convergence of their minimizers
\citep{dal2012introduction,braides2006handbook}.
See
\Cref{sec:ivbnormalproof} for proof details and a review of
$\Gamma$-convergence. \hfill\qedsymbol\\

We characterized the limiting properties of the \gls{VB} ideal 
and their \gls{KL} minimizers. We will next show
that the \gls{VB} posterior is close to the \gls{KL} divergence
minimizer of the \gls{VB} ideal. \Cref{sec:vbconsistency}
culminates in the main theorem of this paper -- the variational
Bernstein--von Mises theorem -- showing the \gls{VB} posterior share
consistency and asymptotic normality with the \gls{KL} divergence
minimizer of \gls{VB} ideal.

%%% Local Variables:
%%% mode: latex
%%% TeX-master: "main"
%%% End:

%% file: sec_vbconsistency.tex
% !TEX root = main.tex
\section{Frequentist consistency of variational Bayes}
\label{sec:vbconsistency}

We now study the \gls{VB} posterior. In the previous section, we
proved theoretical properties for the \gls{VB} ideal and its
\gls{KL} minimizer in the variational family. Here we first connect
the \gls{VB} ideal to the \gls{VB} posterior, the quantity
that is used in practice.  We then use this connection to understand
the theoretical properties of the \gls{VB} posterior.

We begin by characterizing the optimal variational distribution in a
useful way.  Decompose the variational family as
\begin{align*}
    q(\theta, z) = q(\theta) q(z),
\end{align*}
where $q(\theta) = \prod_{i=1}^{d} q(\theta_i)$ and
$q(z) = \prod_{i=1}^{n} q(z_i)$.  Denote the prior $p(\theta)$.  Note
$d$ does not grow with the size of the data.  We will develop a theory
around \gls{VB} that considers asymptotic properties of the \gls{VB}
posterior $q^*(\theta)$.

We decompose the \gls{ELBO} of \Cref{eq:elbo} into the portion
associated with the global variable and the portion associated with
the local variables,
\begin{align*}
    \gls{ELBO}(q(\theta)q(z))
    &= \int \int q(\theta)q(z)\log \frac{p(\theta, x, z)}{q(\theta)
    q(z)} \dif \theta \dif z\\
    &= \int \int q(\theta) q(z) \log \frac{p(\theta)
    p(x, z\mid \theta)}{q(\theta) q(z)} \dif \theta \dif z\\
    &= \int q(\theta) \log \frac{p(\theta)}{q(\theta)} \dif \theta
    + \int q(\theta) \int q(z) \log \frac{p(x, z \g \theta)}{q(z)}\dif \theta \dif z.
\end{align*}

The optimal variational factor for the global variables, i.e., the
\gls{VB} posterior, maximizes the \gls{ELBO}.  From the decomposition,
we can write it as a function of the optimized local variational
factor,
\begin{align}
    q^*(\theta) =
    \argmax_{q(\theta)} \,\,
    \sup_{q(z)}
    \int q(\theta)
    \left(
    \log \left[p(\theta)
    \exp\left\{
    \int q(z) \log \frac{p(x, z\mid \theta)}{q(z)}\dif z
    \right\}\right]
    - \log q(\theta) \right) \dif \theta.
    \label{eq:vb_est_eq}
\end{align}

One way to see the objective for the \gls{VB} posterior is as the
\gls{ELBO} profiled over $q(z)$, i.e., where the optimal $q(z)$ is a
function of $q(\theta)$ \citep{Hoffman:2013}.  With this perspective,
the \gls{ELBO} becomes a function of $q(\theta)$ only.  We denote it
as a functional $\gls{ELBO}_p(\cdot)$:
\begin{align}
    \label{eq:vb_post}
    \gls{ELBO}_p(q(\theta)):=\sup_{q(z)}
    \int q(\theta)
    \left(
    \log \left[p(\theta)
    \exp\left\{
    \int q(z) \log \frac{p(x, z\mid \theta)}{q(z)}\dif z
    \right\}\right]
    - \log q(\theta) \right) \dif \theta.
\end{align}
We then rewrite \Cref{eq:vb_est_eq} as
$q^*(\theta) = \argmax_{q(\theta)}\gls{ELBO}_p(q(\theta))$.  This
expression for the \gls{VB} posterior is key to our results.

\subsection{\gls{KL} minimizers of the \gls{VB} ideal}

Recall that the \gls{KL} minimization objective to the ideal \gls{VB}
posterior is the functional $\gls{KL}(\cdot || \pi^*(\theta\mid x))$.
We first show that the two optimization objectives
$\gls{KL}(\cdot || \pi^*(\theta\mid x))$ and $\gls{ELBO}_p(\cdot)$ are
close in the limit.  Given the continuity of both
$\gls{KL}(\cdot || \pi^*(\theta\mid x))$ and $\gls{ELBO}_p(\cdot)$,
this implies the asymptotic properties of optimizers of
$\gls{KL}(\cdot || \pi^*(\theta\mid x))$ will be shared
by the optimizers of $\gls{ELBO}_p(\cdot)$.\\

\begin{lemma} The negative \gls{KL} divergence to the
\gls{VB} ideal is equivalent to the profiled \gls{ELBO} in the limit:
under mild technical conditions on the tail behavior of
$\mathcal{Q}^d$ (see for example \Cref{assumption:vbpost_tech} in
\Cref{sec:vbpostproof}), for $q(\theta)\in
\mathcal{Q}^d,$
\label{lemma:vb_post}
\[
    \gls{ELBO}_p(q(\theta)) = 
    -\gls{KL}(q(\theta) || \pi^*(\theta\mid x)) + o_P(1).
\]

\end{lemma}

\emph{Proof sketch of \Cref{lemma:vb_post}}. We first notice that
\begin{align}
&-\gls{KL}(q(\theta)||\pi^*(\theta\mid x))\\
=&\int q(\theta)\log\frac{p(\theta)\exp(M_n(\theta\s x))}{q(\theta)}\dif \theta\\
= &\int q(\theta)
\left(
\log \left[p(\theta)
\exp\left\{\sup_{q(z)}
\int q(z) \log \frac{p(x, z\mid\theta)}{q(z)}
\dif z\right\}\right]
- \log q(\theta) \right) \dif \theta.
\label{eq:ivb_post}
\end{align}
Comparing \Cref{eq:ivb_post} with \Cref{eq:vb_post}, we can see that
the only difference between $-\gls{KL}(\cdot||\pi^*(\theta\mid x))$
and $\gls{ELBO}_p(\cdot)$ is in the position of $\sup_{q(z)}$.
$\gls{ELBO}_p(\cdot)$ allows for a single choice of optimal $q(z)$
given $q(\theta)$, while $-\gls{KL}(\cdot||\pi^*(\theta\mid x))$
allows for a different optimal $q(z)$ for each value of $\theta$. In
this sense, if we restrict the variational family of $q(\theta)$ to
be point masses, then $\gls{ELBO}_p(\cdot)$ and
$-\gls{KL}(\cdot||\pi^*(\theta\mid x))$ will be the same.

The only members of the variational family of $q(\theta)$ that admit
finite $-\gls{KL}(q(\theta)||\pi^*(\theta\mid x))$ are ones that
converge to point masses at rate $\delta_n$, so we expect
$\gls{ELBO}_p(\cdot)$ and $-\gls{KL}(\cdot||\pi^*(\theta\mid x))$ to
be close as $n\rightarrow \infty.$ We prove this by bounding the
remainder in the Taylor expansion of $M_n(\theta\s x)$ by a sequence
converging to zero in probability. See \Cref{sec:vbpostproof} for
details.
\hfill\qedsymbol\\

\subsection{The \gls{VB} posterior}

\Cref{sec:ideal-vb-posterior} characterizes the asymptotic behavior
of the \gls{VB} ideal $\pi^*(\theta\mid x)$ and their
\gls{KL} minimizers. \Cref{lemma:vb_post} establishes the
connection between the \gls{VB} posterior $q^*(\theta)$ and the
\gls{KL} minimizers of the \gls{VB} ideal $\pi^*(\theta\mid
x)$.  Recall $\argmin_{q(\theta)\in\mathcal{Q}^d}
\gls{KL}(q(\theta)||\pi^*(\theta\mid x))$ is consistent and converges
to the \gls{KL} minimizer of a normal distribution. We now build on
these results to study the \gls{VB} posterior $q^*(\theta)$.

Now we are ready to state the main theorem. It establishes the
asymptotic behavior of the \gls{VB} posterior $q^*(\theta)$.\\

\begin{thm}
\label{thm:main}
\emph{(Variational Bernstein-von-Mises Theorem)}
\begin{enumerate}
\item The \gls{VB} posterior is consistent: 
\[
    q^*(\theta)
    \stackrel{d}{\rightarrow} \delta_{\theta_0}\qquad\text{in $P_{\theta_0}$-probability}.\\
\]

\item The \gls{VB} posterior is asymptotically normal in the sense
that it converges to the \gls{KL} minimizer of a normal distribution:
\begin{align}
  \left\|q^*_{\tilde{\theta}}(\cdot) - \argmin_{q\in\mathcal{Q}^d}
    \gls{KL}(q(\cdot)|| \cN(\cdot \s\Delta_{n,\theta_0},
    V^{-1}_{\theta_0})) \right\|
  _{\gls{TV}}\stackrel{P_{\theta_0}}{\rightarrow} 0.
\end{align}
Here we transform $q^*(\theta)$ to
$q_{\tilde{\theta}}(\tilde{\theta})$, which is centered around the
true $\theta_0$ and scaled by the convergence rate; see
\Cref{eq:transformed-theta}. When $\mathcal{Q}^d$ is the mean field variational family, then the limiting \gls{VB} posterior is normal:
\begin{align}
\label{eq:normalvarfactor}
\argmin_{q\in\mathcal{Q}^d}
    \gls{KL}(q(\cdot)|| \cN(\cdot \s\Delta_{n,\theta_0},
    V^{-1}_{\theta_0})) = \cN(\cdot \s\Delta_{n,\theta_0},
    V'^{-1}_{\theta_0})),
\end{align}
where $V'_{\theta_0}$ is diagonal and has the same diagonal terms as $V_{\theta_0}$.
\end{enumerate}
\end{thm}

\emph{Proof sketch of \Cref{thm:main}.} This theorem is a direct
consequence of \Cref{lemma:ivbconsist}, \Cref{lemma:ivbnormal},
\Cref{lemma:vb_post}. We need the same mild technical conditions on
$\mathcal{Q}^d$ as in \Cref{lemma:ivbnormal} and
\Cref{lemma:vb_post}. \Cref{eq:normalvarfactor} can be proved by first establishing the normality of the optimal variational factor (see Section 10.1.2 of \citet{bishop2006pattern} for details) and proceeding with \Cref{lemma:normalkl}. See \Cref{sec:mainproof} for
details.\hfill\qedsymbol\\

Given the convergence of the \gls{VB} posterior, we can now establish
the asymptotic properties of the \gls{VBE}.\\

\begin{thm} (Asymptotics of the \gls{VBE})

Assume $\int |\theta|^2\pi(\theta)\dif \theta<\infty$. Let
$\hat{\theta}_n^* = \int \theta \cdot  q_1^*(\theta)\dif \theta$ denote the
\gls{VBE}.

\begin{enumerate}
    \item The \gls{VBE}
    is consistent: 
    \[\hat{\theta}_n^*\stackrel{P_{\theta_0}}{\rightarrow}\theta_0.\]
    \item The \gls{VBE} is asymptotically normal in the sense that it
    converges in distribution to the mean of the \gls{KL}
    minimizer:\footnote{The randomness in the mean of the \gls{KL}
    minimizer comes from $\Delta_{n,\theta_0}$.} if $\Delta_{n,\theta_0}\stackrel{d}{\rightarrow}X$ for some $X$, 
    \[\delta_n^{-1}(\hat{\theta}_n^*-\theta_0)\stackrel{d}{\rightarrow}
    \int \tilde{\theta} \cdot \argmin_{q\in\mathcal{Q}^d} 
    \gls{KL}(q(\tilde{\theta})|| \cN(\tilde{\theta} \s X, V^{-1}_{\theta_0} ))
    \dif\tilde{\theta}.
    \]
\end{enumerate}
\label{thm:estimate}
\end{thm}
\emph{Proof sketch of \Cref{thm:estimate}.} As the posterior mean is
a continuous function of the posterior distribution, we would expect
the \gls{VBE} is consistent given the
\gls{VB} posterior is. We also know that the posterior mean is the
Bayes estimator under squared loss. Thus we would expect the
\gls{VBE} to converge in distribution to squared loss minimizer of
the \gls{KL} minimizer of the \gls{VB} ideal. The result follows from
a very similar argument from Theorem 2.3 of
\citet{kleijn2012bernstein}, which shows that the posterior mean
estimate is consistent and asymptotically normal under model
misspecification as a consequence of the Bernsterin--von Mises
theorem and the argmax theorem. See \Cref{sec:mainproof} for details.
\hfill
\qedsymbol\\

We remark that $\Delta_{n,\theta_0}$, as in
\Cref{assumption:classicbvm}, is a random vector bounded in
$P_{\theta_0}$ probability. The randomness is due to $x$ being a
random sample generated from $P_{\theta_0}$.

In cases where \gls{VFE} is consistent, like in all the examples we
will see in \Cref{sec:app}, $\Delta_{n,\theta_0}$ is a zero mean
random vector with finite variance. For particular realizations of
$x$ the value of $\Delta_{n,\theta_0}$ might not be zero; however,
because we scale by $\delta_n^{-1}$, this does not destroy the
consistency of
\gls{VB} posterior or the \gls{VBE}.

\subsection{Gaussian \gls{VB} posteriors}

We illustrate the implications of \Cref{thm:main} and
\Cref{thm:estimate} on two choices of variational families: a full
rank Gaussian variational family and a factorizable Gaussian
variational family. In both cases, the \gls{VB} posterior and the
\gls{VBE} are consistent and asymptotically normal with different
covariance matrices. The \gls{VB} posterior under the factorizable
family is underdispersed.\\

\begin{corollary}
\label{corollary:fullrankgaussian}
  Posit a full rank Gaussian variational family, that is
  \begin{align}
    \mathcal{Q}^d = \{q:q(\theta) =\cN(m, \Sigma)\},
  \end{align}
  with $\Sigma$ positive definite. Then
\begin{enumerate}
    \item $q^*(\theta)\stackrel{d}{\rightarrow} \delta_{\theta_0}$ 
    in $P_{\theta_0}$-probability.
    \item 
    $
    ||q^*_{\tilde{\theta}}(\cdot)
    -
    \cN(\cdot\s\Delta_{n,\theta_0}, V^{-1}_{\theta_0})
    ||
    _{\gls{TV}}\stackrel{P_{\theta_0}}{\rightarrow} 0.
    $
    \item $\hat{\theta}_n^*\stackrel{P_{\theta_0}}{\rightarrow}\theta_0.$
    \item $\delta_n^{-1}( \hat{\theta}_n^* - \theta_0) 
    - \Delta_{n,\theta_0} = o_{P_{\theta_0}}(1)$.\\
\end{enumerate}
\end{corollary}

\emph{Proof sketch of \cref{corollary:fullrankgaussian}.} This is a direct consequence of \Cref{thm:main} and \Cref{thm:estimate}. We only need to show that \Cref{lemma:ivbnormal} is also true for the full rank Gaussian variational family. The last conclusion implies $\delta_n^{-1}( \hat{\theta}_n^* - \theta_0) \stackrel{d}{\rightarrow}X$ if $\Delta_{n,\theta_0}\stackrel{d}{\rightarrow}X$ for some random variable $X$. We defer the proof to \Cref{sec:fullrankgaussianproof}.
\hfill\qedsymbol\\

This corollary says that under a full rank Gaussian variational
family, \gls{VB} is consistent and asymptotically normal in the
classical sense. It accurately recovers the asymptotic normal
distribution implied by the local asymptotic normality of
$M_n(\theta\s x)$.

Before stating the corollary for the factorizable Gaussian
variational family, we first present a lemma on the \gls{KL}
 minimizer of a Gaussian distribution over the factorizable
Gaussian family. We show that the minimizer keeps the mean but has a
diagonal covariance matrix that matches the precision. We also show
the minimizer has a smaller entropy than the original distribution.
This echoes the well-known phenomenon of \gls{VB} algorithms
underestimating the variance.\\

\begin{lemma}
\label{lemma:normalkl}
The factorizable \gls{KL}  minimizer of a Gaussian
distribution keeps the mean and matches the precision:
\[
    \argmin_{\mu_0\in\mathbb{R}^d,
    \Sigma_0\in\text{diag}(d\times d)}
    \gls{KL}(\cN(\cdot;\mu_0,\Sigma_0)||\cN(\cdot;\mu_1,\Sigma_1))
    = \mu_1, \Sigma^*_1,
\]
where $\Sigma_1^*$ is diagonal with $\Sigma^*_{1, ii} =
((\Sigma_1^{-1})_{ii})^{-1}$ for $i = 1, 2, ..., d$. Hence, the
entropy of the factorizable \gls{KL} minimizer is smaller
than or equal to that of the original distribution:
\[\mathbb{H}(\cN(\cdot;\mu_0,\Sigma^*_1))\leq \mathbb{H}(\cN(\cdot;\mu_0,\Sigma_1)).\\\]
\end{lemma}

\emph{Proof sketch of \Cref{lemma:normalkl}.} The first statement is
consequence of a technical calculation of the \gls{KL} divergence
between two normal distributions. We differentiate the \gls{KL}
divergence over $\mu_0$ and the diagonal terms of $\Sigma_0$ and
obtain the result. The second statement is due to the inequality of
the determinant of a positive matrix being always smaller than or
equal to the product of its diagonal terms
\citep{amir1969product,beckenbach2012inequalities}. In this sense,
mean field variational inference underestimates posterior variance.
See \Cref{sec:normalklproof} for details.\hfill\qedsymbol\\

The next corollary studies the \gls{VB} posterior and the \gls{VBE}
under a factorizable Gaussian variational family.\\

\begin{corollary}
\label{corollary:meanfieldgaussian}
  Posit a factorizable Gaussian variational family,
  \begin{align}
    \mathcal{Q}^d = \{q:q(\theta) =\cN(m, \Sigma)\}
  \end{align}
  where $\Sigma$ positive definite and diagonal. Then
\begin{enumerate}
\item $q^*(\theta)\stackrel{d}{\rightarrow} \delta_{\theta_0}$ 
in $P_{\theta_0}$-probability.
\item 
$
    ||q^*_{\tilde{\theta}}(\cdot)
    -
    \cN(\cdot\s\Delta_{n,\theta_0}, V'^{-1}_{\theta_0})
    ||
    _{\gls{TV}}\stackrel{P_{\theta_0}}{\rightarrow} 0,
$

where $V'$ is diagonal and has the same diagonal entries as
$V_{\theta_0}$.
\item $\hat{\theta}_n^*\stackrel{P_{\theta_0}}{\rightarrow}\theta_0.$
\item $\delta_n^{-1}( \hat{\theta}_n^* - \theta_0) 
- \Delta_{n,\theta_0} = o_{P_{\theta_0}}(1)$.\\
\end{enumerate}
\end{corollary}

\emph{Proof of \cref{corollary:meanfieldgaussian}.} This is a
direct consequence of \Cref{lemma:normalkl}, \Cref{thm:main}, and
\Cref{thm:estimate}. \hfill\qedsymbol\\

This corollary says that under the factorizable Gaussian variational
family, \gls{VB} is consistent and asymptotically normal in the
classical sense. The rescaled asymptotic distribution for
$\tilde{\theta}$ recovers the mean but underestimates the covariance.
This underdispersion is a common phenomenon we see in mean field
variational Bayes.

As we mentioned, the \gls{VB} posterior is underdispersed.  One
consequence of this property is that its credible sets can suffer from
under-coverage.  In the literature on \gls{VB}, there are two main
ways to correct this inadequacy. One way is to increase the
expressiveness of the variational family $\mathcal{Q}$ to one that
accounts for dependencies among latent variables. This approach is
taken by much of the recent \gls{VB} literature,
e.g. \citet{tran2015copula,tran2015variational,ranganath2016hierarchical,ranganath2016operator,liu2016stein}. As
long as the expanded variational family $\mathcal{Q}$ contains the
mean field family, \Cref{thm:main} and \Cref{thm:estimate} remain
true.

Alternative methods to handling underdispersion center around
sensitivity analysis and bootstrap.  \citet{giordano2017covariances}
identified the close relationship between Bayesian sensitivity and
posterior covariance. They estimated the covariance with the
sensitivity of the \gls{VB} posterior means with respect to
perturbations of the data. \citet{chen2017use} explored the use of
bootstrap in assessing the uncertainty of a variational point
estimate. They also studied the underlying bootstrap theory.
\citet{giordano2017measuring} assessed the clutering stability in
Bayesian nonparametric models based on an approximation to the
infinitesimal jackknife.

\subsection{The \gls{LAN} condition of the the variational log likelihood}

\label{subsec:lan}

Our results rest on \Cref{assumption:classicbvm}.3, the \gls{LAN}
expansion of the variational log likelihood $M_n(\theta\s x)$.  For
models without local latent variables $z$, their variational log
likelihood $M_n(\theta\s x)$ is the same as their log likelihood
$\log p(x\mid\theta)$. The \gls{LAN} expansion for these models have
been widely studied. In particular, iid sampling from a regular
parametric model is locally asymptotically normal; it satisfies
\Cref{assumption:classicbvm}.3 \citep{van2000asymptotic}. When models
do contain local latent variables, however, as we will see in
\Cref{sec:app}, finding the \gls{LAN} expansion requires model-specific characterization.

For a certain class of models with local latent variables, the
\gls{LAN} expansion for the (complete) log likelihood $\log p(x,
z\mid\theta)$ concurs with the expansion of the variational log
likelihood $M_n(\theta\s x)$. Below we provide a sufficient condition
for such a shared \gls{LAN} expansion.  It is satisfied, for example,
by the stochastic block model \citep{bickel2013asymptotic} under mild
identifiability conditions.\\

\begin{prop}
\label{prop:variationallan}
The log likelihood $\log p(x, z\mid \theta)$ and the variational log
likelihood $M_n(\theta\s x)$ will have the same \gls{LAN}
expansion if:
\begin{enumerate}
    \item The conditioned nuisance posterior is consistent under
    $\delta_n$-perturbation at some rate $\rho_n$ with
    $\rho_n\downarrow 0$ and $\delta_n^{-2}\rho_n\rightarrow 0$:

    For all bounded, stochastic $h_n = O_{P_{\theta_0}}(1)$, the conditional nuisance posterior converges as
    \[\int_{D^c(\theta, \rho_n)} p(z\mid x, \theta = \theta_0 + \delta_n h_n)\dif z = o_{P_{\theta_0}}(1),\]
    where $D^c(\theta, \rho_n) = \{z:d_H(z, z_{\text{profile}})\geq \rho_n\}$ is the Hellinger ball of radius $\rho_n$ around $z_{\text{profile}} = \argmax_z p(x, z \mid \theta)$, the maximum profile likelihood estimate of $z$. 

    \item $\rho_n$ should also satisfy that the likelihood ratio is dominated:
    \[\sup_{z\in \{z:d_H(z, z_{\text{profile}})<\rho_n\}} \mathbb{E}_{\theta_0}\frac{p(x, z \mid \theta_0+\delta_n h_n)}{p(x, z \mid \theta_0)} = O(1),\]
    where the expectation is taken over $x$.

\end{enumerate}
\end{prop}

\emph{Proof sketch of \Cref{prop:variationallan}.} The first
condition roughly says the posterior of the local latent variables
$z$ contracts faster than the global latent variables $\theta$. The
second conidtion is a regularity condition. The two conditions
together ensure the log marginal likelihood $\log \int p(x, z\mid
\theta)\dif z$ and the complete log likelihood $\log p(x, z \mid
\theta)$ share the same \gls{LAN} expansion. This condition shares a
similar flavor with the condition (3.1) of the semiparametric
Bernstein--von Mises theorem in \citet{bickel2012semiparametric}.
This implication can be proved by a slight adaptation of the proof of
Theorem 4.2 in \citet{bickel2012semiparametric}: We view the
collection of local latent variables $z$ as an infinite-dimensional
nuisance parameter. 

This proposition is due to the following key inequality. For
simplicity, we state the version with only discrete local latent
variables $z$:
\begin{align}
\label{eq:boundvi}
  \log p(x, z\mid \theta)\leq M_n(\theta\s x) \leq \log \int p(x, z\mid \theta)\dif z.
\end{align}
The continuous version of this inequality can be easily adapted. The lower bound is due to
\[p(x, z\mid \theta) = \left.\int q(z)\log \frac{p(x, z\mid
      \theta)}{q(x)}\dif z\right\vert_{q(z) = \delta_z},\] and
\[M_n(\theta \s x) = \sup_{q\in\mathcal{Q}^d}\int q(z)\log \frac{p(x, z\mid \theta)}{q(x)}\dif z.\] The upper bound is due to the Jensen's inequality. This inequality ensures that the same \gls{LAN} expansion for the leftmost and the rightmost terms would imply the same \gls{LAN} expansion for the middle term, the variational log likelihood $M_n(\theta\s x)$. See \Cref{sec:lanproof} for details. \hfill\qedsymbol\\

In general we can appeal to Theorem 4 of \citet{le2012asymptotics} to
argue for the preservation of the \gls{LAN} condition, showing that if
it holds for the complete log likelihood then it holds for the
variational log likelihood too.  In their terminology, we need to
establish the \gls{VFE} as a ``distinguished'' statistic.

%%% Local Variables:
%%% mode: latex
%%% TeX-master: "main"
%%% End:

%% file: sec_applications.tex
% !TEX root = main.tex
\section{Applications}
\label{sec:app}
\glsresetall

We proved consistency and asymptotic normality of the \gls{VB}
posterior (in \gls{TV} distance) and the \gls{VBE}.  We mainly relied
on the prior mass condition, the local asymptotic normality of the
variational log likelihood $M_n(x\s\theta)$ and the consistent
testability assumption of the data generating parameter.

We now apply this argument to three types of Bayesian models: Bayesian
mixture models \citep{bishop2006pattern,murphy2012machine}, Bayesian
generalized linear mixed models
\citep{mcculloch2001generalized,jiang2007linear}, and Bayesian
stochastic block models
\citep{wang1987stochastic,snijders1997estimation,
  mossel2012stochastic,abbe2015community,Wiggins:2008}.  For each
model class, we illustrate how to leverage the known asymptotic
results for frequentist variational approximations to prove asymptotic
results for \gls{VB}. We assume the prior mass condition for the rest
of this section: the prior measure of a parameter $\theta$ with
Lebesgue density $p(\theta)$ on $\Theta$ is continuous and positive
on a neighborhood of the true data generating value $\theta_0$. For
simplicity, we posit a mean field family for the local latent
variables and a factorizable Gaussian variational family for the
global latent variables.

\subsection{Bayesian Mixture models}

The Bayesian mixture model is a versatile class of models for density
estimation and clustering
\citep{bishop2006pattern,murphy2012machine}.

Consider a Bayesian mixture of $K$ unit-variance univariate Gaussians
with means $\mu = \{\mu_1, ..., \mu_K\}$.  For each observation
$x_i, i = 1, ..., n$, we first randomly draw a cluster assignment
$c_i$ from a categorical distribution over $\{1, ..., K\}$; we then
draw $x_i$ randomly from a unit-variance Gaussian with mean
$\mu_{c_i}$.  The model is
\begin{align*}
    \mu_k &\sim p_{\mu},& k = 1, ..., K,\\
    c_i &\sim \textrm{Categorical}(1/K, ..., 1/K), &i = 1, ..., n,\\
    x_i\mid c_i, \mu &\sim \cN(c_i^\top \mu, 1). &i = 1, ..., n.
\end{align*}
For a sample of size $n$, the joint distribution is
\begin{align*}
    p(\mu, c, x) = 
    \prod^K_{i=1}p_\mu(\mu_i)\prod^n_{i=1}p(c_i)p(x_i\mid c_i,\mu).
\end{align*}
Here $\mu$ is a $K$-dimensional global latent vector and $c_{1:n}$ are
local latent variables. We are interested inferring the posterior of
the $\mu$ vector.

We now establish asymptotic properties of \gls{VB} for Bayesian
\gls{GMM}.\\
\begin{corollary}
Assume the data generating measure $P_{\mu_0}$ has density $\int
p(\mu_0, c, x)\dif c$. Let $q^*(\mu)$ and $\mu^*$ denote the
\gls{VB} posterior and the
\gls{VBE}. Under regularity conditions (A1-A5) and (B1,2,4) of \citet{westling2015establishing}, we have
\begin{align*}
    \left\|q^*(\mu) - \cN\left(\mu_0 + \frac{Y}{\sqrt{n}},
    \frac{1}{n}V_0(\mu_0)\right)\right\|_{\gls{TV}}\stackrel{P_{\mu_0}}{
    \rightarrow} 0,
\end{align*}
and
\begin{align*}
    \sqrt{n}(\mu^* - \mu_0)\stackrel{d}{\rightarrow} Y,
\end{align*}
where $\mu_0$ is the true value of $\mu$ that generates the
data. We have
    \[Y\sim\cN(0, V(\mu_0)),\] 
    \[V(\mu_0) = A(\mu_0)^{-1}B(\mu_0)A(\mu_0)^{-1},\]
    \[A(\mu) = 
    \mathbb{E}_{P_{\mu_0}}[D^2_\mu m(\mu\s x)],\]
    \[B(\mu) = 
    \mathbb{E}_{P_{\mu_0}}[D_\mu m(\mu\s x)D_\mu m(\mu\s x)^\top],\]
    \[m(\mu\s x) = 
    \sup_{q(c)\in\mathcal{Q}^n}\int q(c)
    \log\frac{p(x,c\mid\mu)}{q(c)}\dif c.\]
The diagonal matrix $V_0(\mu_0)$ satisfies $(V_{0}(\mu_0)^{-1})_{ii}
= (A(\mu_0))_{ii}$. The specification of Gaussian mixture model is
invariant to permutation among $K$ components; this corollary is true
up to permutations among the $K$ components.\\
\label{corollary:gmm}
\end{corollary}

\emph{Proof sketch for \Cref{corollary:gmm}}. The consistent
testability condition is satisfied by the existence of a consistent
estimate due to Theorem 1 of \citet{westling2015establishing}. The
local asymptotic normality is proved by a Taylor expansion of
$m(\mu\s x)$ at $\mu_0$. This result then follows directly from our
\Cref{thm:main} and \Cref{thm:estimate} in \Cref{sec:vbconsistency}.
The technical conditions inherited from
\citet{westling2015establishing} allow us to use their Theorems 1 and
2 for properties around \gls{VFE}. See \Cref{sec:gmmproof} for proof
details. \hfill
\qedsymbol\\

\subsection{Bayesian Generalized linear mixed models}

Bayesian \glspl{GLMM} are a powerful class of models for analyzing
grouped data or longitudinal data
\citep{mcculloch2001generalized,jiang2007linear}.

Consider a Poisson mixed model with a simple linear relationship and
group-specific random intercepts.  Each observation reads $(X_{ij},
Y_{ij}), 1\leq i\leq m, 1\leq j \leq n$, where the $Y_{ij}$'s are
non-negative integers and the $X_{ij}$'s are unrestricted real
numbers. For each group of observations $(X_{ij}, Y_{ij}), 1\leq j
\leq n$, we first draw the random effect $U_i$ independently from
$N(0, \sigma^2)$. We follow by drawing $Y_{ij}$ from a Poisson
distribution with mean $\exp(\beta_0+ \beta_1X_{ij}+U_i)$. The
probability model is
\begin{align*}
    \beta_0 &\sim p_{\beta_0},\\
    \beta_1 &\sim p_{\beta_1},\\
    \sigma^2 &\sim p_{\sigma^2},\\
    U_i &\stackrel{iid}{\sim} \cN(0, \sigma^2),\\
    Y_{ij}\mid X_{ij}, U_i &\sim 
    \textrm{Poi}(\exp(\beta_0+\beta_1X_{ij}+U_i)).
\end{align*}
The joint distribution is
\begin{align*}
    p(\beta_0, \beta_1, &
    \sigma^2, U_{1:m}, Y_{1:m, 1:n}\mid X_{1:m, 1:n})= \\ 
    & p_{\beta_0}(\beta_0)p_{\beta_1}(\beta_1)p_{\sigma^2}(\sigma^2)
    \prod^m_{i=1}\cN(U_i;0, \sigma^2)\times
    \prod^m_{i=1}\prod^n_{j=1} \textrm{Poi}(Y_{ij};
    \exp(\beta_0+\beta_1X_{ij}+U_i)).
\end{align*}

We establish asymptotic properties of \gls{VB} in Bayesian Poisson
linear mixed models.\\
\begin{corollary}
Consider the true data generating distribution $P_{\beta_0^0,
\beta_1^0, (\sigma^2)^0}$ with the global latent variables taking the
true values $\{\beta_0^0, \beta_1^0, (\sigma^2)^0\}$. Let
$q^*_{\beta_0}(\beta_0)$, $q^*_{\beta_1}(\beta_1)$,
$q^*_{\sigma^2}(\sigma^2)$ denote the
\gls{VB} posterior of $\beta_0,
\beta_1, \sigma^2$. Similarly, let $\beta_0^*, \beta_1^*,
(\sigma^2)^*$ be the \gls{VBE}s accordingly. Consider $m = O(n^2)$.
Under regularity conditions (A1-A5) of \citet{hall2011asymptotic}, we
have
\begin{align*}
    \left\|q^*_{\beta_0}(\beta_0)
    q^*_{\beta_1}(\beta_1)
    q^*_{\sigma^2}(\sigma^2)
     - \cN\left((\beta^0_0,\beta^0_1, (\sigma^2)^0)+
    (\frac{Z_1}{\sqrt{n}}, \frac{Z_2}{\sqrt{mn}}, \frac{Z_3}{\sqrt{n}}), 
    \text{diag}(V_1, V_2, V_3)\right)\right\|_{\gls{TV}}
    \stackrel{P_{\beta_0^0, \beta_1^0, (\sigma^2)^0}}{\rightarrow}0,
\end{align*}
where
\[Z_1\sim \cN(0, (\sigma^2)^0), Z_2\sim 
\cN(0, \tau^2), Z_3\sim \cN(0, 2\{(\sigma^2)^0\}^2),\]
\[V_1 = \exp(-\beta_0+\frac{1}{2}(\sigma^2)^0)/\phi(\beta_1^0),\] 
\[V_2 = \exp(-\beta_0^0+\frac{1}{2}\sigma^2)/\phi''(\beta_1^0),\]
\[V_3 = 2\{(\sigma^2)^0\}^2, \]
\[\tau^2 =
\frac{\exp\{-(\sigma^2)^0/2-\beta_0^0\}\phi(\beta_1^0)}
{\phi''(\beta_1^0)\phi(\beta_1^0)-\phi'(\beta_1^0)^2}.\] Here
$\phi(\cdot)$ is the moment generating function of $X$.

Also,
\begin{align*}
    (\sqrt{m}(\beta_0^* - \beta_0^0),
    \sqrt{mn}(\beta_1^* - \beta_1^0),
    \sqrt{m}((\sigma^2)^* - (\sigma^2)^0))
    \stackrel{d}{\rightarrow} 
    (Z_1, Z_2, Z_3).
\end{align*}
\label{corollary:glmm}
\end{corollary}

\emph{Proof sketch for \Cref{corollary:glmm}.} The consistent
testability assumption is satisfied by the existence of consistent
estimates of the global latent variables shown in Theorem 3.1 of
\citet{hall2011asymptotic}. The local asymptotic normality is proved
by a Taylor expansion of the variational log likelihood based on
estimates of the variational parameters based on equations (5.18) and
(5.22) of \citet{hall2011asymptotic}. The technical conditions
inherited from \citet{hall2011asymptotic} allow us to leverage their
Theorem 3.1 for properties of the \gls{VFE}. The result then follows
directly from \Cref{thm:main} and \Cref{thm:estimate} in
\Cref{sec:vbconsistency}. See \Cref{sec:glmmproof} for proof details.
\hfill \qedsymbol\\

\subsection{Bayesian stochastic block models}
Stochastic block models are an important methodology for community
detection in network data \citep{wang1987stochastic,
snijders1997estimation,mossel2012stochastic,abbe2015community}.

Consider $n$ vertices in a graph.  We observe pairwise linkage
between nodes $A_{ij}\in\{0, 1\}, 1\leq i, j\leq n$.  In a stochastic
block model, this adjacency matrix is driven by the following
process: first assign each node $i$ to one of the $K$ latent classes
by a categorical distribution with parameter $\pi$. Denote the class
membership as $Z_i\in \{1, ..., K\}$. Then draw $A_{ij}\sim
\textrm{Bernoulli} (H_{Z_i,Z_j})$.  The parameter $H$ is a symmetric
matrix in $[0,1]^{K\times K}$ that specifies the edge probabilities
between two latent classes; the parameter $\pi$ are the proportions
of the latent classes. The Bayesian stochastic block model is
\begin{align*}
    \pi &\sim p(\pi),\\
    H &\sim p(H),\\
    Z_i\mid \pi &\stackrel{iid}{\sim} \textrm{Categorical}(\pi),\\
    A_{ij}\mid Z_i,Z_j, H &\stackrel{iid}{\sim} 
    \textrm{Bernoulli}(H_{Z_iZ_j}).
\end{align*}
The dependence in stochastic block model is more complicated than the
Bayesian \gls{GMM} or the Bayesian \gls{GLMM}.

Before establishing the result, we reparameterize $(\pi, H)$ by
$\theta = (\omega, \nu)$, where $\omega\in\mathbb{R}^{K-1}$ is the
log odds ratio of belonging to classes $1,...,K-1$, and
$\nu\in\mathbb{R}^{K\times K}$ is the log odds ratio of an edge
existing between all pairs of the $K$ classes.  The
reparameterization is
\begin{align*}
    \omega(a) 
    &= \log\frac{\pi(a)}{1-\sum^{K-1}_{b=1}\pi(b)}, 
    &a = 1,...,K-1,\\
    \nu(a,b) 
    &= \log\frac{H(a,b)}{1-H(a,b)}, & a,b = 1, ..., K.
\end{align*}

The joint distribution is
\begin{align*}
    p(\theta, &Z, A) = \\ 
    & \prod^{K-1}_{a=1}
    [e^{\omega(a)n_a}(1+\sum^{K-1}_{a=1}e^{\omega(a)})^{-n}]
    \times \prod^K_{a=1}\prod^K_{b=1}
    [e^{\nu(a,b)O_{ab}}(1+e^{\nu(a,b)})^{n_{ab}}]^{1/2},
\end{align*}
where
\begin{align*}
    n_a(Z) = &\sum^n_{i=1}1\{Z_i=a\},\\
    n_{ab}(Z) = & \sum^n_{i=1}\sum^n_{j\ne i}1\{Z_i=a, Z_j=b\},\\
    O_{ab}(A, Z) = &\sum^n_{i=1}\sum^n_{j\ne i}1\{Z_i=a, Z_j=b\}A_{ij}.
\end{align*}

We now establish the asymptotic properties of \gls{VB} for stochastic
block models.\\

\begin{corollary}
Consider $\nu_0, \omega_0$ as true data generating parameters. Let
$q^*_{\nu}(\nu), q^*_{\omega}(\omega)$ denote the \gls{VB} posterior
of $\nu$ and $\omega$. Similarly, let $\nu^*, \omega^*$ be the
\gls{VBE}. Then
\begin{align*}
    \left\|q^*_{\nu}(\nu)q^*_{\omega}(\omega)
    -\cN\left((\nu, \omega);
    (\nu_0, \omega_0)+
    (\frac{\Sigma^{-1}_1Y_1}{\sqrt{n\lambda_0}}, 
    \frac{\Sigma^{-1}_2Y_2}{\sqrt{n}}), V_n(\nu_0, \omega_0)
    \right)\right\|_{\gls{TV}}
    \stackrel{P_{\nu_0, \omega_0}}{\rightarrow}0
\end{align*}
where $\lambda_0 = \mathbb{E}_{P_{\nu_0, \omega_0}}$(\text{degree of
each node}), $(\log n)^{-1}\lambda_0\rightarrow \infty.$ $Y_1$ and
$Y_2$ are two zero mean random vectors with covariance matrices
$\Sigma_1$ and $\Sigma_2$, where $\Sigma_1, \Sigma_2$ are known
functions of $\nu_0, \omega_0$. The diagonal matrix $V(\nu_0,
\omega_0)$ satisfies $V^{-1}(\nu_0, \omega_0)_{ii} =
\text{diag}(\Sigma_1, \Sigma_2)_{ii}$. Also,
\begin{align*}
    (\sqrt{n\lambda_0}(\nu^* - \nu_0), 
    \sqrt{n}(\omega^* - \omega_0))
    \stackrel{d}{\rightarrow}
    (\Sigma^{-1}_1Y_1, \Sigma^{-1}_2Y_2),
\end{align*}
The specification of classes in \gls{SBM} is permutation invariant.
So the convergence above is true up to permutation with the $K$
classes. We follow \citet{bickel2013asymptotic} to consider the
quotient space of $(\nu, \omega)$ over permutations.
\label{corollary:sbm}
\end{corollary}

\emph{Proof sketch of \Cref{corollary:sbm}}. The consistent testability
assumption is satisfied by the existence of consistent estimates by
Lemma 1 of \citet{bickel2013asymptotic}. The local asymptotic
normality,
\begin{multline}
    \sup_{q(z)\in\mathcal{Q}^K}
    \int q(z)\log\frac{p(A, z\mid \nu_0+\frac{t}{\sqrt{n^2\rho_n}}, 
    \omega_0+\frac{s}{\sqrt{n}})}{q(z)}\dif z \\
    = \sup_{q(z)\in\mathcal{Q}^K}
    \int q(z)\log\frac{p(A, z\mid \nu_0, \omega_0)}{q(z)}\dif z + 
    s^\top Y_1 + t^\top Y_2 -\frac{1}{2}s^\top\Sigma_1 s - 
    \frac{1}{2}t^\top \Sigma_2 t + o_P(1), 
\end{multline}
for $(\nu_0, \omega_0)\in\mathcal{T}$ for compact $\mathcal{T}$ with
$\rho_n = \frac{1}{n}\mathbb{E}(\text{degree of each node})$, is
established by Lemma 2, 3 and Theorem 3 of \citet{bickel2013asymptotic}. The result
then follows directly from our \Cref{thm:main} and
\Cref{thm:estimate} in \Cref{sec:vbconsistency}. See \Cref{sec:sbmproof} for proof
details. \hfill
\qedsymbol\\

%%% Local Variables:
%%% mode: latex
%%% TeX-master: "main"
%%% End:

%% file: sec_simulation.tex
% !TEX root = main.tex
\section{Simulation studies}
\label{sec:simulation}

We illustrate the implication of \Cref{thm:main} and
\Cref{thm:estimate} by simulation studies on Bayesian \gls{GLMM}
\citep{mccullagh1984generalized}. We also study the \gls{VB}
posteriors of \gls{LDA} \citep{blei2003latent}. This is a model that
shares similar structural properties with \gls{SBM} but has no
consistency results established for its \gls{VFE}.

We use two automated inference algorithms offered in Stan, a
probabilistic programming system \citep{carpenter2015stan}: \gls{VB}
through \gls{ADVI} \citep{kucukelbir2016automatic} and \gls{HMC}
simulation through \gls{NUTS} \citep{hoffman2014nuts}. We note that
optimization algorithms used for \gls{VB} in practice only find local
optima.

In both cases, we observe the \gls{VB} posteriors get closer to the
truth as the sample size increases; when the sample size is large
enough, they coincide with the truth. They are underdispersed,
however, compared with \gls{HMC} methods.

\subsection{Bayesian Generalized Linear Mixed Models}

We consider the Poisson linear mixed model studied in \Cref{sec:app}.
Fix the group size as $n=10$. We simulate data sets of size $N =$
(50, 100, 200, 500, 1000, 2000, 5000, 10000, 20000). As the size of
the data set grows, the number of groups also grows; so does the
number of local latent variables $U_i, 1\leq i\leq m$. We generate a
four-dimensional covariate vector for each $X_{ij}, 1\leq i\leq m,
1\leq j \leq n$, where the first dimension follows i.i.d $\cN(0,1)$,
the second dimension follows i.i.d $\cN(0, 25)$, the third dimension
follows i.i.d Bernoulli$(0.4)$, and the fourth dimension follows
i.i.d. Bernoulli$(0.8)$. We wish to study the behaviors of
coefficient efficients for underdispersed/overdispersed continuous
covariates and balanced/imbalanced binary covariates. We set the true
parameters as $\beta_0 = 5$, $\beta_1 = (0.2, -0.2, 2, -2)$, and
$\sigma^2 = 2$.

\begin{figure}
\begin{subfigure}{.5\textwidth}
    \centering
    \includegraphics[width=1.1\linewidth]{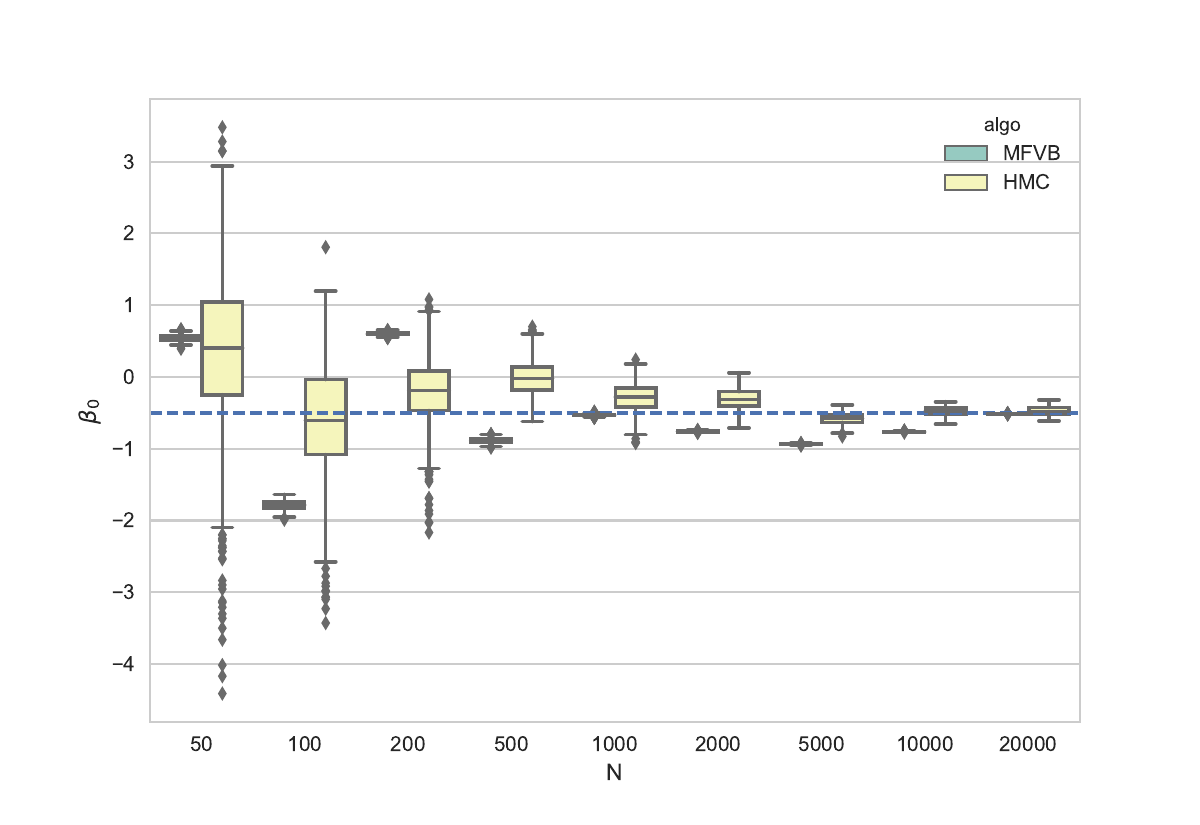}
    \caption{Posterior of $\beta_0$}
    \label{fig:plmmbeta0}
\end{subfigure}%
\begin{subfigure}{.5\textwidth}
    \centering
    \includegraphics[width=1.1\linewidth]{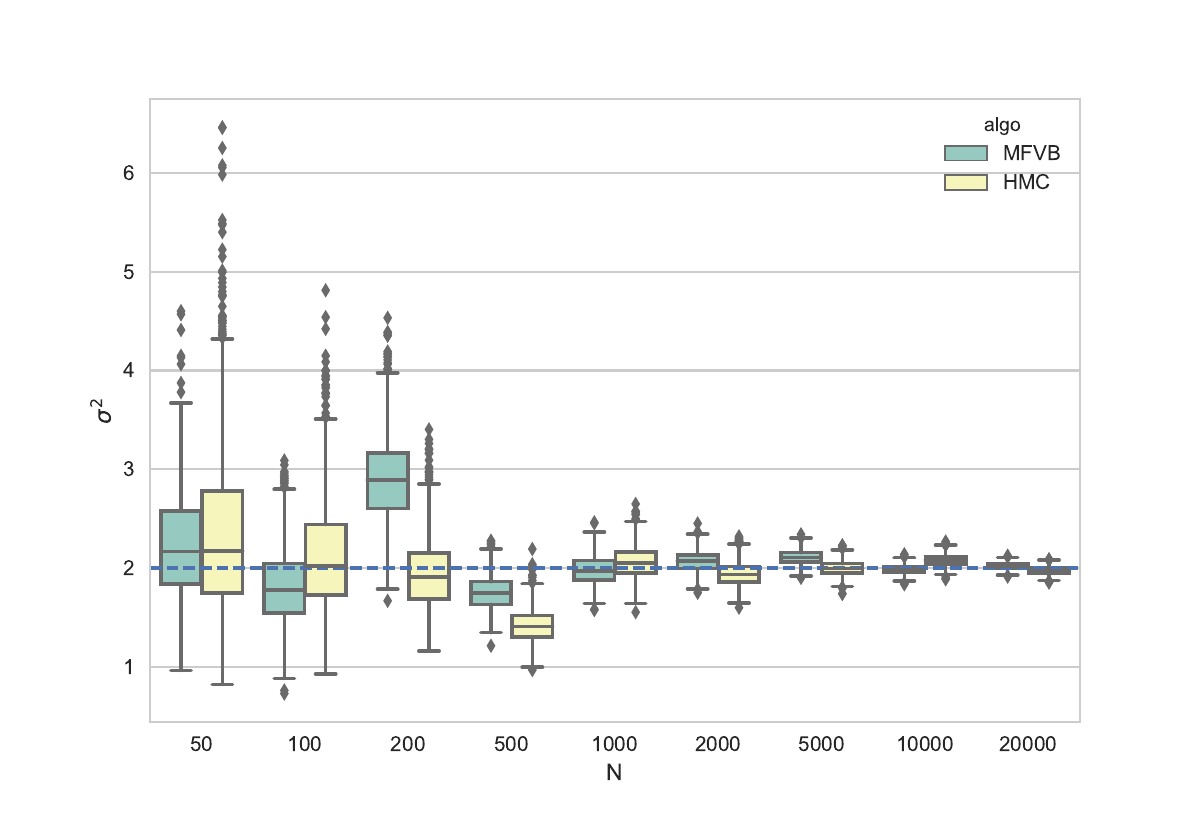}
    \caption{Posterior of $\sigma^2$}
    \label{fig:plmmsigma}
\end{subfigure}
\begin{subfigure}{.5\textwidth}
    \centering
    \includegraphics[width=1.1\linewidth]{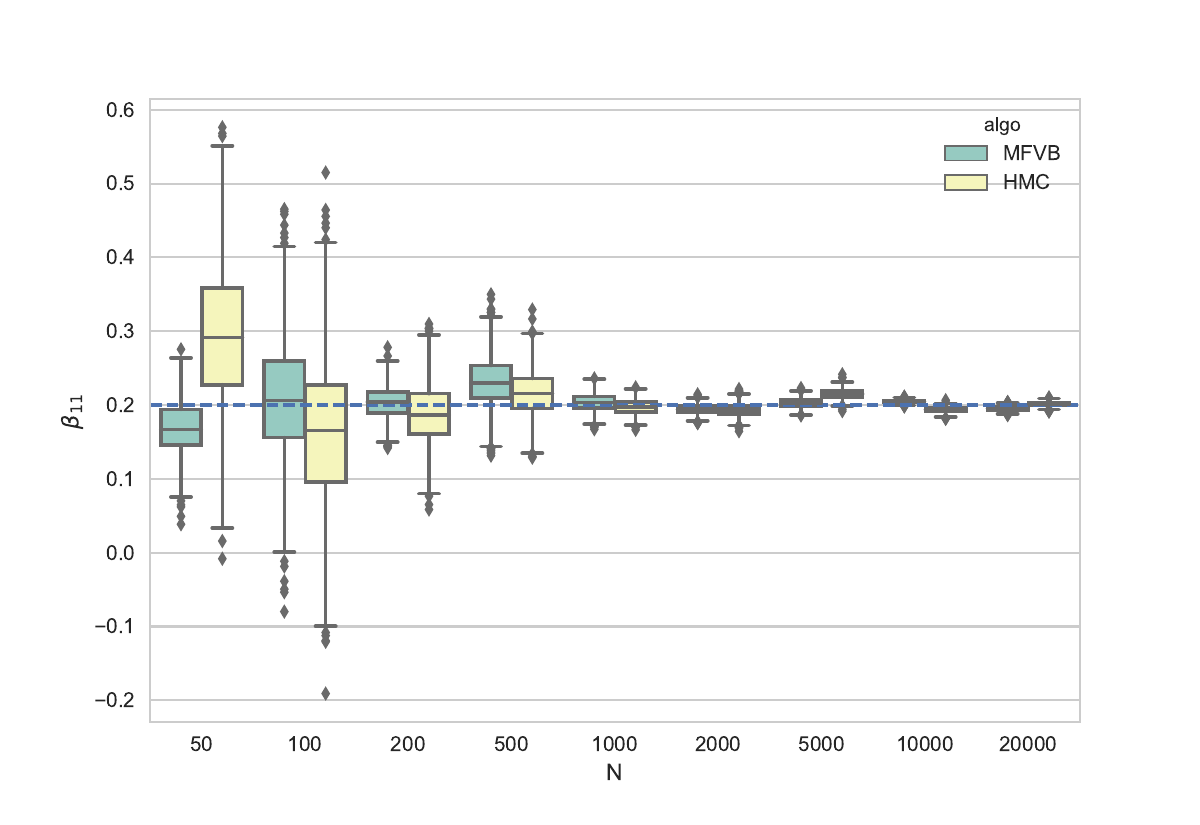}
    \caption{Posterior of $\beta_{11}$}
    \label{fig:plmmbeta11}
\end{subfigure}%
\begin{subfigure}{.5\textwidth}
    \centering
    \includegraphics[width=1.1\linewidth]{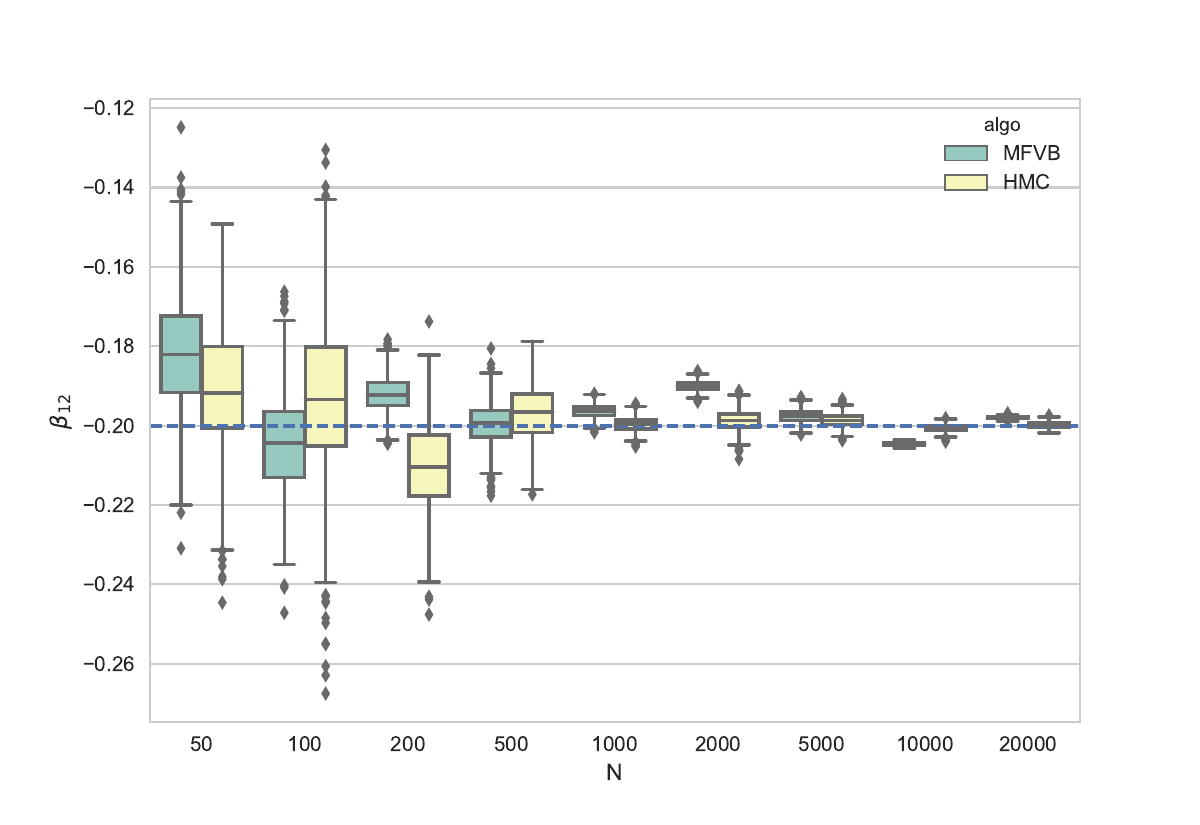}
    \caption{Posterior of $\beta_{12}$}
    \label{fig:plmmbeta12}
\end{subfigure}
\begin{subfigure}{.5\textwidth}
    \centering
    \includegraphics[width=1.1\linewidth]{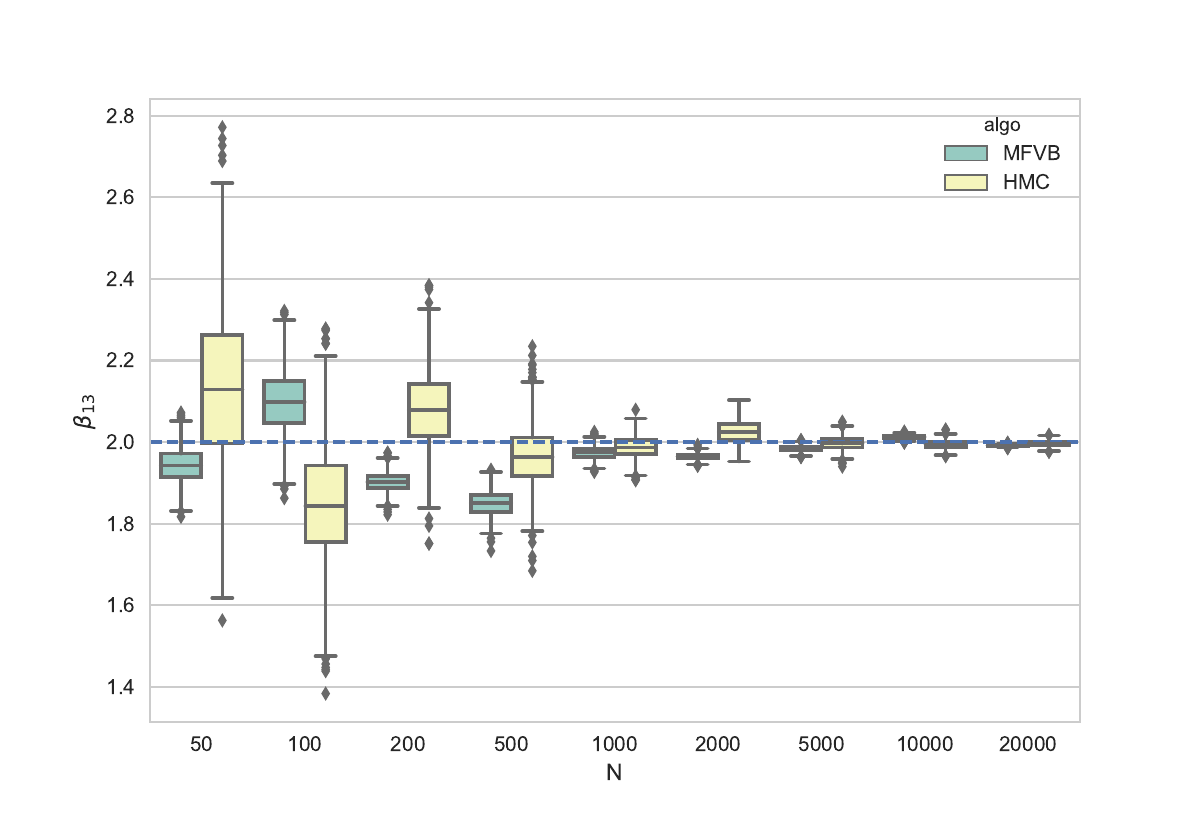}
    \caption{Posterior of $\beta_{13}$}
    \label{fig:plmmbeta13}
\end{subfigure}%
\begin{subfigure}{.5\textwidth}
    \centering
    \includegraphics[width=1.1\linewidth]{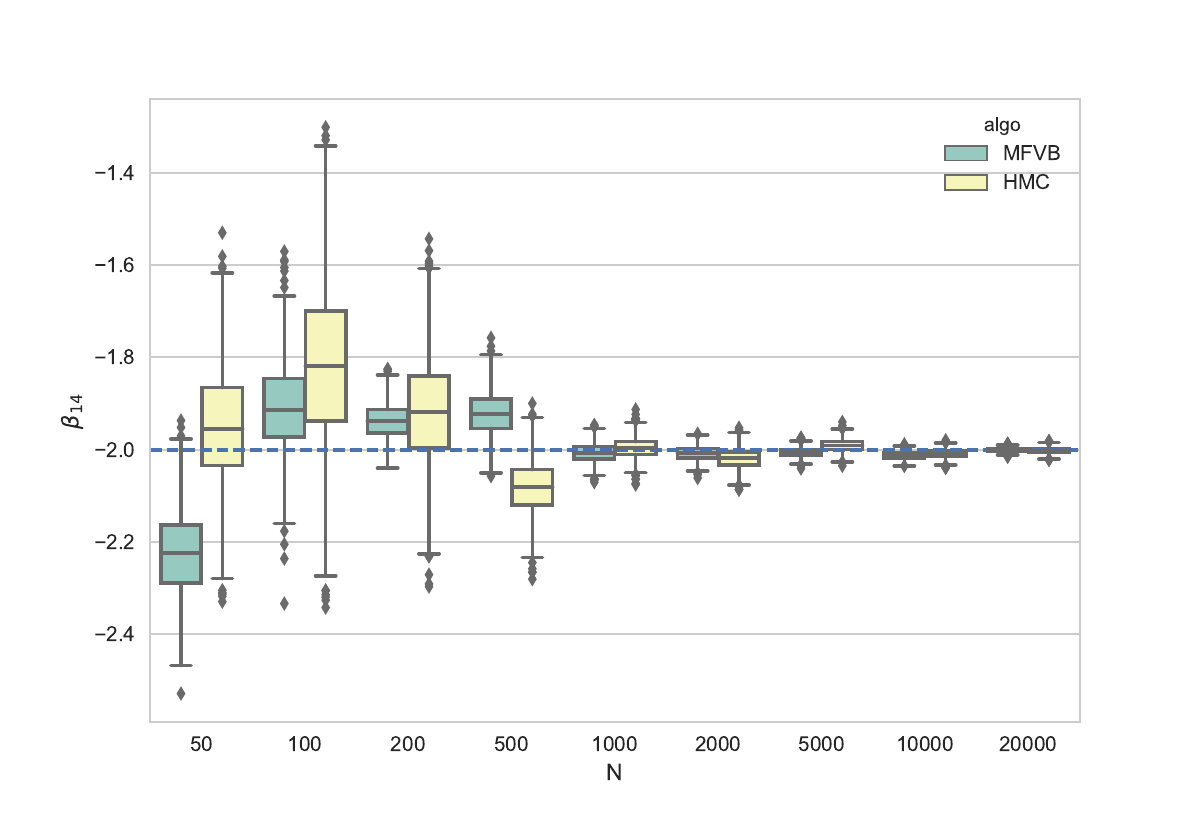}
    \caption{Posterior of $\beta_{14}$}
    \label{fig:plmmbeta14}
\end{subfigure}
\caption{\gls{VB} posteriors and \gls{HMC} posteriors of Poisson
Generalized Linear Mixed Model versus size of datasets. \gls{VB}
posteriors are consistent and asymptotically normal but
underdispersed than \gls{HMC} posteriors. $\beta_0$ and $\sigma^2$
converge to the truth slower than $\beta_1$ does. They echo our
conclusions in \Cref{thm:main} and \Cref{corollary:glmm}.}
\label{fig:plmmsim}
\end{figure}

\Cref{fig:plmmsim} shows the boxplots of \gls{VB} posteriors for
$\beta_0, \beta_1,$ and $\sigma^2$. All \gls{VB} posteriors converge
to their corresponding true values as the size of the data set
increases. The box plots present rather few outliers; the lower
fence, the box, and the upper fence are about the same size. This
suggests normal \gls{VB} posteriors. This echoes the consistency and
asymptotic normality concluded from \Cref{thm:main}. The \gls{VB}
posteriors are underdispersed, compared to the posteriors via
\gls{HMC}. This also echoes our conclusion of underdispersion in
\Cref{thm:main} and \Cref{lemma:normalkl}.

Regarding the convergence rate, \gls{VB} posteriors of all dimensions
of $\beta_1$ quickly converge to their true value; the \gls{VB}
posteriors center around their true values as long as $N \geq 1000$.
The convergence of \gls{VB} posteriors of slopes for continuous
variables ($\beta_{11}, \beta_{12}$) are generally faster than those
for binary ones ($\beta_{13}, \beta_{14}$). The \gls{VB} posterior of
$\sigma^2$ shares a similarly fast convergence rate. The \gls{VB}
posterior of the intercept $\beta_0$, however, struggles; it is away
from the true value until the data set size hits $N = 20000$. This
aligns with the convergence rate inferred in \Cref{corollary:glmm},
$\sqrt{mn}$ for $\beta_1$ and $\sqrt{m}$ for $\beta_0$ and
$\sigma^2.$

Computation wise, \gls{VB} takes orders of magnitude less time than
\gls{HMC}. The performance of \gls{VB} posteriors is comparable with
that from \gls{HMC} when the sample size is sufficiently large; in
this case, we need $N = 20000.$

\subsection{Latent Dirichlet Allocation}
Latent Dirichlet Allocation (\gls{LDA}) is a generative statistical
model commonly adopted to describe word distributions in documents by
latent topics.

Given $M$ documents, each with $N_m, m = 1, ..., M$ words, composing
a vocabulary of $V$ words, we assume $K$ latent topics. Consider two
sets of latent variables: topic distributions for document $m$,
$(\theta_m)_{K\times 1}$, $m = 1, ..., M$ and word distributions for
topic $k$, $(\phi_k)_{V\times 1}$, $k = 1, ..., K$. The generative
process is
\begin{align*}
    \theta_m\sim& p_\theta,& 
    m= 1, ..., M,\\
    \phi_k\sim& p_\phi,& 
    k= 1, ..., K,\\
    z_{m,j}\sim& \textrm{Mult}(\theta_m), &
    j = 1, ..., N_m, m= 1, ..., M, \\
    w_{m,j}\sim& \textrm{Mult}(\phi_{z_{m,j}}), &
    j = 1, ..., N_m, m= 1, ..., M.
\end{align*}
The first two rows are assigning priors assigned to the latent
variables. $w_{m,j}$ denotes word $j$ of document $m$ and $z_{m,j}$
denotes its assigned topic.

We simulate a data set with $V = 100$ sized vocabulary and $K = 10$
latent topics in $M =$ (10, 20, 50, 100, 200, 500, 100) documents.
Each document has $N_m$ words where
$N_m\stackrel{iid}{\sim}$Poi(100). As the number of documents $M$
grows, the number of document-specific topic vectors $\theta_m$ grows
while the number of topic-specific word vectors $\phi_k$ stays the
same. In this sense, we consider $\theta_m, m = 1, ..., M$ as local
latent variables and $\phi_k, k = 1, ..., K$ as global latent
variables. We are interested in the \gls{VB} posteriors of global
latent variables $\phi_k, k=1, ..., K$ here. We generate the data
sets with true values of $\theta$ and $\phi$, where they are random
draws from $\theta_m\stackrel{iid}{\sim}$Dir$((1/K)_{K\times 1})$ and
$\phi_k\stackrel{iid}{\sim}$Dir$((1/V)_{V\times 1})$.

\begin{figure}
\begin{subfigure}{.5\textwidth}
    \centering
    \includegraphics[width=\linewidth]{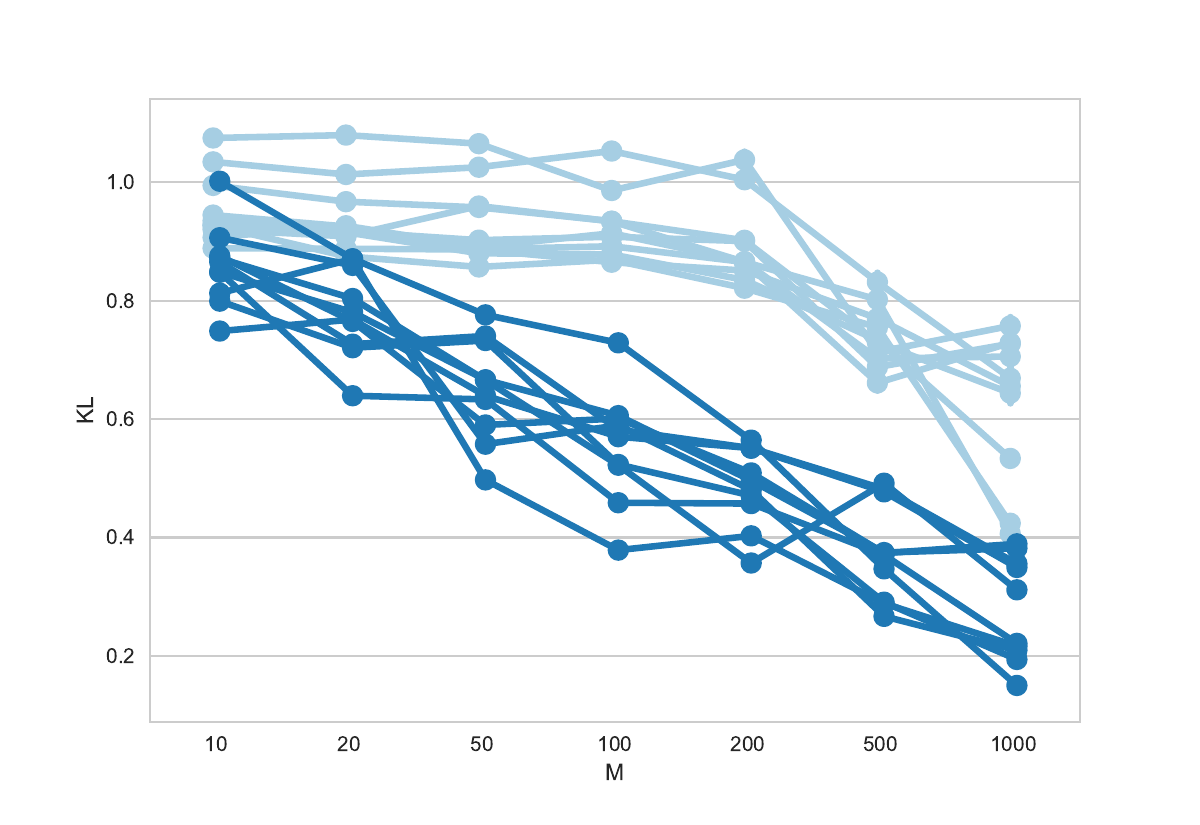}
    \caption{Posterior mean \gls{KL} divergence of the \\$K=10$ topics}
    \label{fig:lda_M_vbkl}
\end{subfigure}%
\begin{subfigure}{.5\textwidth}
    \centering
    \includegraphics[width=\linewidth]{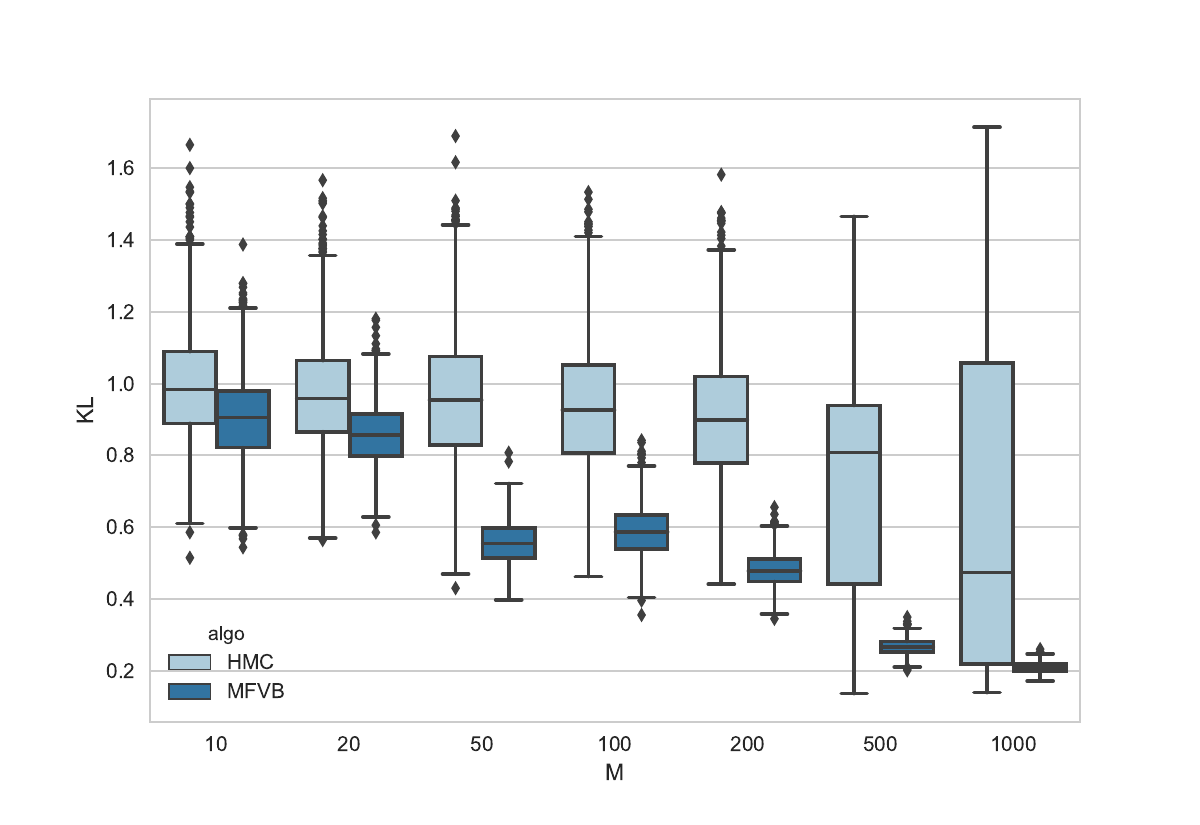}
    \caption{Boxplots of posterior \gls{KL} divergence of Topic 2}
    \label{fig:lda_M_hmckl}
\end{subfigure}
\caption{Mean of \gls{KL} divergence between the true topics and the
fitted \gls{VB} and \gls{HMC} posterior topics versus size of
datasets. (a) \gls{VB} posteriors (dark blue) converge to the truth;
they are very close to the truth as we hit $M=1000$ documents. (b)
\gls{VB} posteriors are consistent but underdispersed compared to
\gls{HMC} posteriors (light blue). These align with our conclusions
in \Cref{thm:main}. \label{fig:ldasim}}
\end{figure}

\Cref{fig:ldasim} presents the \gls{KL} divergence between the $K=10$
topic-specific word distributions induced by the true $\phi_k$'s and
the fitted $\phi_k$'s by \gls{VB} and \gls{HMC}. This \gls{KL}
divergence equals to \gls{KL}(Mult($\phi_k^0$)||Mult($\hat{\phi}_k))
= \sum^V_{i=1}\phi_{ki}^0(\log \phi_{ki}^0 - \log \hat{\phi}_{ki})$,
where $\phi_{ki}^0$ is the $i$th entry of the true $k$th topic and
$\hat{\phi}_{ki}$ is the $i$th entry of the fitted $k$th topic.

\Cref{fig:ldasim}a shows that \gls{VB} posterior (dark blue) mean
\gls{KL} divergences of all $K=10$ topics get closer to 0 as the
number of documents $M$ increase, faster than \gls{HMC} (light blue).
We become very close to the truth as the number of documents $M$ hits
1000. \Cref{fig:ldasim}b\footnote{We only show boxplots for Topic 2
here. The boxplots of other topics look very similar.} shows that the
boxplots of \gls{VB} posterior mean \gls{KL} divergences get closer
to 0 as $M$ increases. They are underdispersed compared to \gls{HMC}
posteriors. These align with our understanding of how \gls{VB}
posterior behaves in \Cref{thm:main}.

Computation wise, again \gls{VB} is orders of magnitude faster than
\gls{HMC}. In particular, optimization in \gls{VB} in our simulation
studies converges within 10,000 steps.

%%% Local Variables:
%%% mode: latex
%%% TeX-master: "main"
%%% End:

%% file: sec_conclusion.tex
% !TEX root = main.tex
\section{Discussion}
\label{sec:discussion}

Variational Bayes (\gls{VB}) methods
\glsresetall
are a fast alternative to \gls{MCMC} for posterior inference in
Bayesian modeling. However, few theoretical guarantees have been
established. This work proves consistency and asymptotic normality for
\gls{VB} posteriors. The convergence is in the sense of \gls{TV}
distance converging to zero in probability. In addition, we establish
consistency and asymptotic normality of
\gls{VBE}. The result is frequentist in the sense that we assume a
data generating distribution driven by some fixed nonrandom true value
for global latent variables.

These results rest on ideal variational Bayes and its connection to
frequentist variational approximations. Thus this work bridges the gap
in asymptotic theory between the frequentist variational
approximation, in particular the \gls{VFE}, and variational Bayes. It
also assures us that variational Bayes as a popular approximate
inference algorithm bears some theoretical soundness.

We present our results in the classical \gls{VB} framework but the
results and proof techniques are more generally applicable. Our
results can be easily generalized to more recent developments of
\gls{VB} beyond \gls{KL} divergence, $\alpha$-divergence or
$\chi$-divergence for example
\citep{li2016renyi,dieng2017variational}. They are also applicable to
more expressive variational families, as long as they contain the mean
field family. We could also allow for model misspecification, as long
as the variational loglikelihood $M_n(\theta\s x)$ under the
misspecified model still enjoys local asymptotic normality.

There are several interesting avenues for future work. The variational
Bernstein--von Mises theorem developed in this work applies to
parametric and semiparametric models. One direction is to study the
\gls{VB} posteriors in nonparametric settings. A second direction is
to characterize the finite-sample properties of \gls{VB} posteriors.
Finally, we characterized the asymptotics of an optimization problem,
assuming that we obtain the global optimum.  Though our simulations
corroborated the theory, \gls{VB} optimization typically finds a local
optimum. Theoretically characterizing these local optima requires
further study of the optimization loss surface.

%%% Local Variables:
%%% mode: latex
%%% TeX-master: "main"
%%% End:

%% file: sec_supp.tex
% !TEX root = main.tex
\appendix
{\Large\textbf{Appendix}}
\section{Proof of \Cref{lemma:vbvm}}
\label{sec:vbvmproof}

What we need to show here is that our consistent testability
assumption implies assumption (2.3) in \citet{kleijn2012bernstein}:
\[\int_{\tilde{\theta}>M_n} \pi^*_{\tilde{\theta}}(\tilde{\theta}\mid x) \dif
\tilde{\theta}\stackrel{P_{\theta_0}}{\rightarrow} 0\] for every sequence of
constants $M_n \rightarrow \infty$, where $\tilde{\theta} = \delta_n^{-1}(\theta-\theta_0).$

This is a consequence of a slight generalization of
Theorem 3.1 of \citet{kleijn2012bernstein}. That theorem shows this
implication for the iid case with a common $\sqrt{n}$-convergence
rate for all dimensions of $\theta$. Specifically, they rely on a
suitable test sequence under misspecification of uniform exponential
power around the true value $\theta_0$ to split the posterior
measure.

To show this implication in our case, we replace all $\sqrt{n}$ by
$\delta_n^{-1}$ in the proofs of Theorem 3.1, Theorem 3.3, Lemma 3.3,
Lemma 3.4 of \citet{kleijn2012bernstein}. We refer the readers to
\citet{kleijn2012bernstein} and omit the proof here.

\section{Proof of \Cref{lemma:ivbconsist}}
\label{sec:ivbconsistproof}

We first perform a change of variable step regarding the mean field
variational family. In light of \Cref{lemma:vbvm}, we know that the
\gls{VB} ideal degenerates to a point mass at the rate of
$\delta_n^{-1}$. We need to assume a variational family that
degenerates to points masses at the same rate as the ideal \gls{VB}
posterior. This is because if the variational distribution converges
to a point mass faster than $\pi(\theta\mid x)$, then the \gls{KL}
divergence between them will converge to $+\infty$. This makes the
\gls{KL} minimization meaningless as $n$ increases.
\footnote{\Cref{eq:canceldelta1} and \Cref{eq:canceldelta2} in the
proof exemplify this claim.}

To avoid this pathology, we assume a variational family for the
rescaled and re-centered $\theta$, $\check{\theta}:=
\delta_n^{-1}(\theta - \mu)$, for some $\mu \in \Theta$.  This is a
centered and scaled transformation of $\theta$, centered to an
arbitrary $\mu$.  (In contrast, the previous transformation
$\tilde{\theta}$ was centering $\theta$ at the true $\theta_0$.)
With this transformation, the variational family is
\begin{align}
  q_{\check{\theta}}(\check{\theta}) = q(\mu + \delta_n
  \check{\theta}) | \textrm{det}(\delta_n) |,
\end{align}
where $q(\cdot)$ is the original mean field variational family.  We
will overload the notation in \Cref{sec:introduction} and write this
transformed family as $q(\theta)$ and the corresponding family
$\mathcal{Q}^d$.

In this family, for each fixed $\mu$, $\theta$ degenerates
to a point mass at $\mu$ as $n$ goes to infinity.
$\mu$ is not necessarily equal to the true value
$\theta_0$. We allow $\mu$ to vary throughout the
parameter space $\Theta$ so that $\mathcal{Q}^d$ does not restrict
what the distributions degenerate to. $\mathcal{Q}^d$ only constrains
that the variational distribution degenerates to some point mass at
the rate of $\delta_n$. This step also does not restrict the
applicability of the theoretical results. In practice, we always have
finite samples with fixed $n$, so assuming a fixed variational family
for $\check{\theta}$ as opposed to $\theta$ amounts to a change of
variable $\theta = \mu + \delta_n \check{\theta}$.

Next we show consistency of the \gls{KL} minimizer of the \gls{VB}
ideal.

To show consistency, we need to show that the mass of the \gls{KL}
minimizer 
\[q^\ddagger :=
\argmin_{q(\theta)\in\mathcal{Q}^d}
        \gls{KL}(q(\theta)||\pi^*(\theta\mid x))\]
concentrates near $\theta_0$ as $n\rightarrow \infty$. That is,
\begin{align}
\int_{B(\theta_0, \xi_n)} q^\ddagger (\theta)\dif 
\theta\stackrel{P_{\theta_0}}{\rightarrow} 1,
\label{eq:intconv}
\end{align}
for some $\xi_n\rightarrow 0$ as $n\rightarrow \infty$. This implies
\[q^\ddagger (\theta) \stackrel{d}{\rightarrow} \delta_{\theta_0}\]
in $P_{\theta_0}$-probability.\\

To begin with, we first claim that 
\begin{align}
\label{claim:upbdkl}
\limsup_{n\rightarrow\infty}\min\gls{KL}
(q(\theta)||\pi^*(\theta\mid x)) \leq M,
\end{align}
for some constant $M >0$,
and 
\begin{align}
\label{claim:compact}
\int_{\mathbb{R}^d\backslash K}q^{\ddagger}(\theta)\dif 
\theta\stackrel{}{\rightarrow}0,
\end{align}
where $K$ is the compact set assumed in the local asymptotic
normality condition.

The first claim says that the limiting minimum \gls{KL} divergence is
upper bounded. The intuition is that a choice of $q(\theta)$ with
$\mu = \theta_0$ would have a finite \gls{KL} divergence
in the limit. This is because (rougly) $\pi^*(\theta\mid x)$
converges to a normal distribution centered at $\theta_0$ with rate
$\delta_n$, so it suffices to have a $q(\theta)$ that shares the same
center and the same rate of convergence.

The second claim says that the restriction of $q^\ddagger(\theta)$ to
the compact set $K$, due to the set compactness needed in the \gls{LAN}
condition, will not affect our conclusion in the limit. This is
because the family of $\mathcal{Q}^d$ we assume has a 
shrinking-to-zero scale. In this way, as long as $\mu$ resides within
$K$, $q^\ddagger(\theta)$ will eventually be very close to its
renormalized restriction to the compact set $K$, $q^{\ddagger,
K}(\theta)$, where
\[q^{\ddagger, K}(\theta) = \frac{q^\ddagger(\theta)\cdot
\mathbb{I}_\theta(K)}{\int q^\ddagger(\theta)\cdot
\mathbb{I}_\theta(K)\dif  \theta}.\] We will prove these claims at
the end.

To show $\int_{B(\theta_0, \xi_n)} q^{\ddagger, K} (\theta)\dif
\theta\stackrel{P_{\theta_0}}{\rightarrow} 1,$ we both upper bound
and lower bound this integral. This step mimicks the Step 2 in the
proof of Lemma 3.6 along with Lemma 3.7 in \citet{lu2016gaussapprox}.

We first upper bound the integral using the \gls{LAN} condition,
\begin{align}
&\int q^{\ddagger,K}(\theta)M_n(\theta\s x)\dif \theta\\
=&\int q^{\ddagger,K}(\theta)\nonumber\\
&\times\left[M_n(\theta_0\s x)+
\delta_n^{-1}(\theta-\theta_0)^\top 
V_{\theta_0}\Delta_{n,\theta_0} - 
\frac{1}{2}[\delta_n^{-1}(\theta-\theta_0)]^\top 
V_{\theta_0}[\delta_n^{-1}(\theta-\theta_0)]+o_P(1)\right]\dif \theta\\
\leq& M_n(\theta_0\s x) - C_1\sum_{i=1}^d\frac{\eta^2}{\delta_{n,ii}^2}
\int_{B(\theta_0, \eta)^c} q^{\ddagger,K}(\theta)\dif \theta +o_P(1),
\label{eq:klupperbd}
\end{align}
for large enough $n$ and $\eta << 1$ and some constant $C_1 > 0$. The
first equality is due to the \gls{LAN} condition. The second
inequality is due to the domination of quadratic term for large $n$.

Then we lower bound the integral using our first claim. By \[\limsup_
{n\rightarrow\infty}\gls{KL}(q^{\ddagger,K}(\theta)||\pi^*(\theta\mid
x)) \leq M,\] we can have
\begin{align}
&\int q^{\ddagger,K}(\theta)M_n(\theta\s x)
\dif \theta\geq M_n(\theta_0\s x)-M_0,
\end{align}
for some large constant $M_0>M$.

This step is due to a couple of steps of technical calculation and
the \gls{LAN} condition. To show this implication, we first rewrite the
\gls{KL} divergence as follows.
\begin{align}
&\gls{KL}(q^{\ddagger,K}(\theta)||\pi^*(\theta\mid x))\\
=&\int q^{\ddagger,K}(\theta)\log q^{\ddagger,K}(\theta)\dif \theta - 
\int q^{\ddagger,K}(\theta)\log \pi^*(\theta\mid x)\dif \theta\\
=&\sum_{i=1}^d\int[\delta_{n,ii}^{-1}q_{h,i}^{\ddagger,K}(h)]
\log [\delta_{n,ii}^{-1}q_{h,i}^{\ddagger,K}(h)]\delta_{n,ii}\dif h - 
\int q^{\ddagger,K}(\theta)\log \pi^*(\theta\mid x)\dif \theta\\
=& \log |\det(\delta_n)|^{-1} + 
\sum^d_{i=1} \mathbb{H}(q_{h,i}^{\ddagger,K}(h)) 
- \int q^{\ddagger,K}(\theta)\log \pi^*(\theta\mid x)\dif \theta,
\label{eq:rewritekl}
\end{align}
where this calculation is due to the form of the $\mathcal{Q}^d$
family we assume and a change of variable of $h =
\delta_n^{-1}(\theta-\mu)$; $h$ is in the same spirit as
$\check{\theta}$ above. Notation wise, $\mu$ is the location
parameter specific to $q^{\ddagger,K}(\theta)$ and
$\mathbb{H}(q_h(h))$ denotes the entropy of distribution $q_h$.

We further approximate the last term by the \gls{LAN} condition.
\begin{align}
&\int q^{\ddagger,K}(\theta)\log \pi^*(\theta\mid x)\dif \theta\\
=& \int q^{\ddagger,K}(\theta)\log\frac{p(\theta)\exp(M_n(\theta\s x))}
{\int p(\theta)\exp(M_n(\theta\s x))\dif \theta}\dif \theta\\
=&\int q^{\ddagger,K}(\theta)\log p(\theta)\dif \theta + 
\int q(\theta)M_n(\theta\s x)\dif \theta 
-\log \int p(\theta)\exp(M_n(\theta\s x))\dif \theta\\
=& \int q^{\ddagger,K}(\theta)\log p(\theta)\dif \theta + 
\int q^{\ddagger,K}(\theta)M_n(\theta\s x)\dif \theta \nonumber\\
&- \left[\frac{d}{2}\log(2\pi) - \frac{1}{2}\log \det V_{\theta_0}+
\log\det(\delta_n)+M_n(\theta_0\s x)+\log p(\theta_0)+o_P(1)\right].
\label{eq:normalizerapprox}
\end{align}
This first equality is due to the definition of $\pi^*(\theta\mid
x)$. The second equality is due to $\int q^{\ddagger,K}(\theta)\dif
\theta = 1$. The third equality is due to Laplace approximation and
the \gls{LAN} condition.

Going back to the \gls{KL} divergence, this approximation gives
\begin{align}
&\gls{KL}(q^{\ddagger,K}(\theta)||\pi^*(\theta\mid x))\\
=& \log |\det(\delta_n)|^{-1} + 
\sum^d_{i=1} \mathbb{H}(q_{h,i}^{\ddagger,K}(h)) - 
\int q^{\ddagger,K}(\theta)\log \pi^*(\theta\mid x)\dif \theta\\
=& \log |\det(\delta_n)|^{-1} + 
\sum^d_{i=1} \mathbb{H}(q_{h,i}^{\ddagger,K}(h)) -  
\int q^{\ddagger,K}(\theta)\log p(\theta)\dif \theta - 
\int q^{\ddagger,K}(\theta)M_n(\theta\s x)\dif \theta \nonumber\\
&+\left[\frac{d}{2}\log(2\pi) - \frac{1}{2}\log \det V_{\theta_0}+
\log\det(\delta_n)+M_n(\theta_0\s x)+
\log p(\theta_0)+o_P(1)\right]\label{eq:canceldelta1}\\
=&\sum^d_{i=1} \mathbb{H}(q_{h,i}^{\ddagger,K}(h)) -  
\int q^{\ddagger,K}(\theta)\log p(\theta)\dif \theta - 
\int q^{\ddagger,K}(\theta)M_n(\theta\s x)\dif \theta\nonumber\\
&+\frac{d}{2}\log(2\pi) - \frac{1}{2}\log \det V_{\theta_0}+
M_n(\theta_0\s x)+\log p(\theta_0)+o_P(1).
\label{eq:klapprox}
\end{align}

The first equality is exactly \Cref{eq:rewritekl}. The second
equality is due to \Cref{eq:normalizerapprox}. The third equality is
due to the cancellation of the two $\log\det(\delta_n)$ terms. This
exemplifies why we assumed the convergence rate of the $\mathcal{Q}^d$
family in the first place; we need to avoid the \gls{KL} divergence
going to infinity.

By \[\limsup_{n\rightarrow\infty}\gls{KL}(q^{\ddagger,K}
(\theta)||\pi^*(\theta\mid x)) \leq M,\]
we have
\begin{align}
&\int q^{\ddagger,K}(\theta)M_n(\theta\s x)\dif \theta\\
\geq &-M +\sum^d_{i=1} \mathbb{H}(q_{h,i}^{\ddagger,K}(h)) -  
\int q^{\ddagger,K}(\theta)\log p(\theta)\dif \theta \nonumber\\
&+ \frac{d}{2}\log(2\pi) - \frac{1}{2}\log \det V_{\theta_0}+
M_n(\theta_0\s x)+\log p(\theta_0)+o_P(1)\\
\geq &-M_0 + M_n(\theta_0\s x) + o_P(1)
\label{eq:kllowerbd}
\end{align}
for some constant $M_0>0$. This can be achieved by choosing a large
enough $M_0$ to make the last inequality true. This is doable because
all the terms does not change with $n$ except $\int
q^{\ddagger,K}(\theta)\log p(\theta)\dif \theta$. And we have
$\limsup_{n\rightarrow \infty}\int q^{\ddagger,K}(\theta)\log
p(\theta)\dif \theta <\infty$ due to our prior mass condition.

Now combining \Cref{eq:kllowerbd} and \Cref{eq:klupperbd}, we have 
\begin{align*}
M_n(\theta_0\s x) - C_1\sum_{i=1}^d\frac{\eta^2}{\delta_{n,ii}^2}
\int_{B(\theta_0, \eta)^c} q^{\ddagger,K}(\theta)\dif \theta +o_P(1)
\geq &-M_0 + M_n(\theta_0\s x).
\end{align*}
This gives
\begin{align*}
\int_{B(\theta_0, \eta)^c} q^{\ddagger,K}(\theta)\dif \theta +o_P(1)
\leq \frac{M_0\cdot (\min_i\delta_{n,ii})^2}{C_2\eta^2},
\end{align*}
for some constant $C_2>0$. The right side of the inequality will go
to zero as $n$ goes to infinity if we choose $\eta =
\sqrt{M_0(\min_i\delta_{n,ii})/C_2}\rightarrow 0$. That is, we just
showed \Cref{eq:intconv} with $\xi_n = \eta$.

We are now left to show the two claims we made at the beginning.

To show \Cref{claim:upbdkl}, it suffices to show that there exists a
choice of $q(\theta)$ such that
\begin{align*}
\limsup_{n\rightarrow\infty}\gls{KL}(q(\theta)||\pi^*(\theta\mid x)) 
<\infty.
\end{align*}
We choose $\tilde{q}(\theta) = \prod_{i=1}^d N(\theta_i;\theta_{0,i}, 
\delta^2_{n,ii}v_i)$ for $v_i>0, i = 1, ..., d$. We thus have
\begin{align}
&\gls{KL}(\tilde{q}(\theta)||\pi^*(\theta\mid x))\\
=&\sum^d_{i=1} \frac{1}{2}\log(v_i)+\frac{d}{2} + d\log(2\pi)-  
\int \tilde{q}(\theta)\log p(\theta)\dif \theta - 
\int \tilde{q}(\theta)M_n(\theta\s x)\dif \theta\nonumber\\
& - \frac{1}{2}\log \det V_{\theta_0}+M_n(\theta_0\s x)+
\log p(\theta_0)+o_P(1)\\
=&\sum^d_{i=1} \frac{1}{2}\log(v_i)+\frac{d}{2} + d\log(2\pi)-  
\log p(\theta_0) -M_n(\theta_0\s x)\nonumber\\
& - \frac{1}{2}\log \det V_{\theta_0}+M_n(\theta_0\s x)+
\log p(\theta_0)+o_P(1)\label{eq:normalintapprox}\\
=&\sum^d_{i=1} \frac{1}{2}\log(v_i)+\frac{d}{2} + d\log(2\pi)
- \frac{1}{2}\log \det V_{\theta_0}+C_6+o_P(1),
\label{eq:relaxcompact}
\end{align}
for some constant $C_6 > 0.$ The finiteness of limsup is due to the
last term being bounded in the limit. The first equality is due to
the same calculation as in \Cref{eq:normalintapprox}. The third
equality is due to the cancellation of the two $M_n(\theta_0\s x)$
terms and the two $p(\theta_0)$ terms; this renders the whole term
independent of $n$. The second equality is due to the limit of
$\tilde{q}(\theta)$ concentrating around $\theta_0$. Specifically, we
expand $\log p(\theta)$ to the second order around $\theta_0$,
\begin{align}
&\int \tilde{q}(\theta)\log p(\theta)\dif \theta\\
=& \log p(\theta_0) + \int \tilde{q}(\theta)\left[(\theta-\theta_0)(\log p(\theta_0))'+
\frac{(\theta-\theta_0)^2}{2}
\int_0^1(\log p(\xi\theta+(1-\xi)\theta_0))''(1-\xi)^2\dif \xi\right]
\dif \theta\\
\leq & \log p(\theta_0) + 
\frac{1}{2!}\max_{\xi\in[0,1]}
\left\{\int \tilde{q}(\theta)(\theta-\theta_0)^2
(\log p(\xi\theta+(1-\xi)\theta_0))''
\dif \theta\right\}\\
\leq&\log p(\theta_0) +\frac{M_p}{\sqrt{(2\pi)^d
\det(\delta_n^2)\prod_iv_i}}\int_{\mathbb{R}^d}|\theta|^2e^{(|\theta|+
|\theta_0|)^2}\cdot e^{-\frac{1}{2}
\theta^\top(\delta_nV\delta_n)^{-1}\theta}
\dif \theta\\
\leq & \log p(\theta_0) +
\frac{M_p}{\sqrt{(2\pi)^d\det(\delta_n^2)\prod_iv_i}} e^{\theta_0^2}
\int_{\mathbb{R}^d}|\theta|^2e^{-\frac{1}{2}\theta^\top
[(\delta_nV\delta_n)^{-1}-2I_d]\theta}\\
\leq & \log p(\theta_0) + C_3M_pe^{\theta_0^2}\max_d(\delta_{n,ii}^2)
\det(V^{-1}-2\delta_n^2)^{-1}\\
\leq & \log p(\theta_0) + C_4\max_d(\delta_{n,ii}^2)
\end{align}
where $\max_d(\delta_{n,ii}^2)\rightarrow 0$ and $C_3, C_4 >0$. The
first equality is due to Taylor expansion with integral form
residuals. The second inequality is due to the first order derivative
terms equal to zero and taking the maximum of the second order
derivative. The third inequality is due to the prior mass condition
where we assume the second derivative of $\log p(\theta)$ is bounded
by $M_pe^{|\theta|^2}$ for some constant $M_p>0$.  The fourth
inequality is pulling $e^{\theta_0^2}$ out of the integral. The fifth
inequality is due to rescaling $\theta$ by its covariance matrix and
appealing to the mean of a Chi-squared distribution with $d$ degrees
of freedom. The sixth (and last) inequality is due to
$\det(V^{-1}-2\delta_n^2)^{-1} > 0$ for large enough $n$.

We apply the same Taylor expansion argument to the $\int
\tilde{q}(\theta)M_n(\theta\s x)\dif \theta$.
\begin{align}
&\int_{K_n} \tilde{q}(\theta)M_n(\theta\s x)\dif \theta\\
=& M_n(\theta_0\s x) + \int_{K_n} \tilde{q}(\theta)
\left[\delta_n^{-1}(\theta-\theta_0)^\top V_{\theta_0}
\Delta_{n,\theta_0}+\frac{1}{2}(\delta_n^{-1}(\theta-\theta_0))^\top 
V_{\theta_0} \delta_n^{-1}(\theta-\theta_0)+o_P(1)\right]\dif \theta\\
\leq &  M_n(\theta_0\s x) + \frac{1}{2}Tr(V_{\theta_0}V) + o_P(1)\\
\leq& M_n(\theta_0\s x) + C_6  + o_P(1)
\end{align}
where $K_n$ is a compact set. The first equality is due to the
\gls{LAN} condition. The second inequality is due to
$\tilde{q}(\theta)$ centered at $\theta_0$ with covariance
$\delta_nV\delta_n$. The third inequalities are true for $C_6 > 0$.

For the set outside of this compact set $K_n$, we consider for a
general choice of $q$ distribution, $\tilde{q}(\theta) =
\cN(\theta;\theta_0+\Delta_{n,\theta_0},
\delta_nV_{\theta_0}\delta_n),$ of which $\tilde{q}(\theta) =
\prod_{i=1}^d \cN(\theta_i;\theta_{0,i},
\delta^2_{n,ii}v_i)$ we work with is a special case.
\begin{align}
&\int_{\mathbb{R}^d\backslash K_n} \tilde{q}(\theta)(\log p(\theta) + 
M_n(\theta\s x))\dif \theta\\
\leq &C_7 \int_{\mathbb{R}^d\backslash K_n} \cN(\theta;\theta_0+
\Delta_{n,\theta_0}, \delta_nV_{\theta_0}\delta_n)(\log p(\theta) + 
M_n(\theta\s x))\dif \theta\\
\leq &C_8 [\det(\delta_n)^{-1}\log(\det(\delta_n)^{-1})] 
\int_{\mathbb{R}^d\backslash K_n} \cN(\tilde{\theta}; \Delta_{n,\theta_0}, 
V_{\theta_0})
\log \pi^*(\tilde{\theta}\mid x) \det(\delta_n)\dif \tilde{\theta}\\
\leq &C_9 \log(\det(\delta_n)^{-1})]\int_{\mathbb{R}^d\backslash K_n}
[\pi^*(\tilde{\theta}\mid x)+o_P(1)]
\log \pi^*(\tilde{\theta}\mid x), V_{\theta_0})\dif \tilde{\theta}\\
\leq & C_{10} \log(\det(\delta_n)^{-1})]
\int_{\mathbb{R}^d\backslash K_n}
[\cN(\tilde{\theta}; \Delta_{n,\theta_0}, V_{\theta_0})+o_P(1)]
\log \cN(\tilde{\theta}; \Delta_{n,\theta_0}, 
V_{\theta_0})\dif \tilde{\theta}\\
\leq &o_P(1)
\end{align}
for some $C_7, C_8, C_9, C_{10} > 0$. The first inequality is due to
$\tilde{q}(\theta)$ centered at $\theta_0$ and with rate of
convergence $\delta_n$. The second inequality is due to a change of
variable $\tilde{\theta} = \delta_n^{-1}(\theta-\theta_0)$. The third inequality
is due to \Cref{lemma:vbvm}. The fourth inequality is due to
\Cref{lemma:vbvm} and Theorem 2 in \citet{piera2009convergence}. The
fifth inequality is due to a choice of fast enough increasing
sequence of compact sets $K_n$.

The lower bound of $\int \tilde{q}(\theta)(\log p(\theta) +
M_n(\theta\s x))\dif \theta$ can be derived with exactly the same
argument. Our first claim \Cref{claim:upbdkl} is thus proved.

To show our second claim \Cref{claim:compact}, we first denote
$B(\mu, M)$ as the largest ball centered at $\mu$ and contained in
the compact set $K$. We know by the construct of $\mathcal{Q}^d$ ---
$\mathcal{Q}^d$ has a shrinking-to-zero scale --- that for each
$\epsilon>0$, there exists an $N$ such that for all $n> N$ we have
$\int_{{||\theta -\mu||>M}}q(\theta)\dif \theta <\epsilon.$
Therefore, we have
\[\int_{\mathbb{R}^d\backslash K}q^{\ddagger}(\theta)\dif \theta\leq
\int_{\mathbb{R}^d\backslash B(\mu,
M)}q^{\ddagger}(\theta)\dif \theta\leq \epsilon.\]

\section{Proof of \Cref{lemma:ivbnormal}}
\label{sec:ivbnormalproof}
To show the convergence of optimizers from two minimization problems,
we invoke $\Gamma$-convergence. It is a classical technique in
characterizing variational problems. A major reason is that if two
functionals $\Gamma-$converge, then their minimizer also converge.

We recall the definition of $\Gamma$-convergence
\citep{dal2012introduction,braides2006handbook}.\\
\begin{defn}
Let $\mathcal{X}$ be a metric space and
$F_\epsilon:\mathcal{X}\rightarrow \mathbb{R}$ a family of
functionals indexed by $\epsilon > 0.$ Then the existence of a
limiting functional $F_0$, the $\Gamma-$limit of $F_\epsilon$, as
$\epsilon \rightarrow 0$, relies on two conditions:
\begin{enumerate}
  \item (liminf inequality) for every $x\in\mathcal{X}$ and for every
  $x_\epsilon\rightarrow x$, we have
  \[F_0(x)\leq\liminf_{\epsilon\rightarrow 0}F_\epsilon(x_\epsilon),\]
  \item (limsup inequality / existence of a recovery sequence) for
  every $x\in\mathcal{X}$ we can find a sequence
  $\bar{x}_\epsilon\rightarrow x$ such that
  \[F_0(x)\geq\limsup_{\epsilon\rightarrow 0}
  F_\epsilon(\bar{x}_\epsilon).\\\]
\end{enumerate}
\end{defn}
The first condition says that $F_0$ is a lower bound for the sequence
$F_\epsilon$, in the sense that $F_0(x) \leq F_\epsilon(x_\epsilon) +
o(1)$ whenever $x_\epsilon\rightarrow x$. Together with the first
condition, the second condition implies that $F_0(x) =
\lim_{\epsilon\rightarrow 0} F_\epsilon(\bar{x}_\epsilon),$ so that
the lower bound is sharp.

$\Gamma$-convergence is particularly useful for variational problems
due to the following fundamental theorem. Before stating the theorem,
we first define equi-coerciveness. \\

\begin{defn} (Equi-coerciveness of functionals)
A sequence $F_\epsilon:\mathcal{X}\rightarrow\bar{\mathbb{R}}$ is
equi-coercise if for all $\epsilon_j\rightarrow 0$ and $x_j$ such
that $F_{\epsilon_j}(x_j)\leq t$ there exist a subsequence of $j$
(not relabeled) and a converging sequence $x'_j$ such that
$F_{\epsilon_j}(x'_j)\leq F_{\epsilon_j}(x_j)+o(1)$.\\
\end{defn}

Equi-coerciveness of functionals ensures that we can find a
precompact minimizing sequence of $F_\epsilon$ such that the
convergence $x_\epsilon\rightarrow x$ can take place. Now we are
ready to state the fundamental theorem.\\

\begin{thm} (Fundamental theorem of $\Gamma$-convergence) Let
$\mathcal{X}$ be a metric space. Let $(F_\epsilon)$ be an equi-
coercise sequence of functions on $\mathcal{X}$. Let $F =
\Gamma-\lim_{\epsilon\rightarrow0}F_\epsilon$, then
\[\argmin_\mathcal{X}F = \lim_{\epsilon\rightarrow0}
\argmin_\mathcal{X}F_\epsilon.\\\]
\label{thm:gammaconv}
\end{thm}

The above theorem implies that if all functions $F_\epsilon$ admit a
minimizer $x_\epsilon$ then, up to subsequences, $x_\epsilon$
converge to a minimum point of $F$. We remark that the converse is
not true; we may have minimizers of $F$ which are not limits of
minimizers of $F_\epsilon$, e.g. $F_\epsilon(t)=\epsilon t^2$
\citep{braides2006handbook}.

In this way, $\Gamma$-convergence is convenient to use when we would
like to study the asymptotic behavior of a family of problem
$F_\epsilon$ through defining a limiting problem $F_0$ which is a
`good approximation' such that the minimizers converge:
$x_\epsilon\rightarrow x_0$, where $x_0$ is a minimizer of $F_0$.
Conversely, we can characterize solutions of a difficult $F_0$ by
finding easier approximating $F_\epsilon$
\citep{braides2006handbook}.

We now prove \Cref{lemma:ivbnormal} for the general mean field
family. The family is parametric as in \Cref{sec:introduction}, so we
assume it is indexed by some finite dimensional parameter $m$. We
want to show that the functionals
\[F_n(m):=\gls{KL}(q(\theta; m)||\pi^*(\theta\mid x))\]
$\Gamma$-converge to
\[F_0(m):=\gls{KL}(q(\theta; m)||\cN(\theta;\theta_0+
\delta_n\Delta_{n,\theta_0},\delta_nV_{\theta_0}^{-1}\delta_n)) 
- \Delta_{n,\theta_0}^\top V_{\theta_0}\Delta_{n,\theta_0}\]
in probability as $n\rightarrow 0$. Recall that the mean field family has density
\[q(\theta) = \prod^d_{i=1}\delta_{n,ii}^{-1}q_{h,i}(h),\]
where $h = \delta_n^{-1}(\theta - \theta_0).$

We need the following mild technical conditions on $\mathcal{Q}^d$.

Following the change-of-variable step detailed in the beginning of
\Cref{sec:ivbconsistproof}, we consider the mean field variational
family with densities $q(\theta) =
\prod^d_{i=1}\delta_{n,ii}^{-1}q_{h,i}(h),$ where $h =
\delta_n^{-1}(\theta - \mu)$ for some $\mu\in\Theta$.\\

\begin{assumption} We assume the following conditions on $q_{h,i}$:
\label{assumption:scoreint}
\begin{enumerate}
\item $q_{h,i}, i = 1, ..., d$ have continuous densities.
\item $q_{h,i}, i = 1, ..., d$ have positive and finite entropies.
\item $\int q_{h,i}'(h)\dif h < \infty, i = 1, ..., d.$\\
\end{enumerate}
\end{assumption}

The last condition ensures that convergence in finite dimensional
parameters imlied convergence in \gls{TV} distance. This is due to a
Taylor expansion argument:
\begin{align}
&\gls{KL}(q(\theta;m)||q(\theta;m+\delta))\\
=&\int q(\theta;m)\cdot(\log q(\theta;m) - 
\log q(\theta;m+\delta))\dif \theta\\
=&\int q(\theta;m)\cdot (\delta \cdot (\log q(\theta;m))')\dif \theta 
+ o(1)\\
=& \delta \int (q(\theta;m))'\dif \theta\\
< &\epsilon
\end{align}
The last step is true if \Cref{assumption:scoreint} is true for $q$.
We also notice that convergence in \gls{KL} divergence implies
convergence in \gls{TV} distance. Therefore,
\Cref{assumption:scoreint} implies that convergence in finite
dimensional parameter implies convergence in
\gls{TV} distance.

Together with \Cref{thm:gammaconv}, the $\Gamma$-convergence of the
two functionals implies $m_n \stackrel{P_{\theta_0}}{\rightarrow}
m_0$ where $m_n$ is the minimizer of $F_n$ for each $n$ and $m_0$ is
the minimizer of $F_0$. This is due to the last term of $F_0$ --
$\Delta_{n,\theta_0}^\top V_{\theta_0}\Delta_{n,\theta_0}$ is a
constant bounded in $P_{\theta_0}$ probability and independent of
$m$. The convergence in total variation then follows from
\Cref{assumption:scoreint} and our argument above.

Lastly, we prove the $\Gamma$-convergence of the two functionals for
the mean field family. 

We first rewrite $F_n(m,\mu)$.
\begin{align}
&F_n(m,\mu):=\gls{KL}(q(\theta;m,\mu)||\pi^*(\theta\mid x))\\
=&\log|\det(\delta_n)|^{-1} +\sum^d_{i=1}\mathbb{H}(q_{h,i}(h;m)) - 
\int q(\theta;m,\mu) \log \pi^*(\theta\mid x))\dif \theta\\
=& \log|\det(\delta_n)|^{-1} +\sum^d_{i=1}\mathbb{H}(q_{h,i}(h;m)) - 
\int q(\theta;m,\mu)\log p(\theta)\dif \theta-\int q(\theta;m,\mu)
M_n(\theta\s x)\dif \theta\nonumber\\
& + \log \int p(\theta)\exp(M_n(\theta\s x))\dif \theta\\
=&\log|\det(\delta_n)|^{-1} +\sum^d_{i=1}\mathbb{H}(q_{h,i}(h;m)) - 
\int q(\theta;m,\mu)\log p(\theta)\dif \theta-\int q(\theta;m,\mu)
M_n(\theta\s x)\dif \theta\nonumber\\
& + \left[\frac{d}{2}\log(2\pi) - \frac{1}{2}\log \det V_{\theta_0}+
\log\det(\delta_n)+M_n(\theta_0\s x)+\log p(\theta_0)+o_P(1)\right]\\
=&\sum^d_{i=1}\mathbb{H}(q_{h,i}(h;m)) - 
\int q(\theta;m,\mu)\log p(\theta)\dif \theta-\int q(\theta;m,\mu)
M_n(\theta\s x)\dif \theta\nonumber\\
& + \left[\frac{d}{2}\log(2\pi) - \frac{1}{2}\log \det V_{\theta_0}+
M_n(\theta_0\s x)+\log p(\theta_0)+o_P(1)\right]
\label{eq:canceldelta2}\\
=&\sum^d_{i=1}\mathbb{H}(q_{h,i}(h;m)) -\int q(\theta;m,\mu)
M_n(\theta\s x)\dif \theta +\log p(\theta_0)
- \log p(\mu)\nonumber\\
&+\left[\frac{d}{2}\log(2\pi) - 
\frac{1}{2}\log \det V_{\theta_0}+M_n(\theta_0\s x) + o_P(1)\right]\\
=&\sum^d_{i=1}\mathbb{H}(q_{h,i}(h;m))-\int q(\theta;m,\mu)
[ M_n(\theta_0\s x) + \delta_n^{-1}(\theta-\theta_0)^\top 
V_{\theta_0}\Delta_{n,\theta_0}+\log p(\theta_0)
- \log p(\mu)\nonumber\\
&-\frac{1}{2}(\delta_n^{-1}(\theta-\theta_0))^\top V_{\theta_0} 
\delta_n^{-1}(\theta-\theta_0)+o_P(1)]\dif \theta - \left[\frac{d}{2}
\log(2\pi) - \frac{1}{2}\log \det V_{\theta_0}+M_n(\theta_0\s x)+o_P(1)
\right]\\
=& \sum^d_{i=1}\mathbb{H}(q_{h,i}(h;m))- \int \delta_n^{-1}
(\theta-\theta_0)^\top V_{\theta_0}\Delta_{n,\theta_0}\cdot 
q(\theta;m,\mu)\dif \theta\nonumber \\
& + \int \frac{1}{2}(\delta_n^{-1}(\theta-\theta_0))^\top 
V_{\theta_0} \delta_n^{-1}(\theta-\theta_0)\cdot q(\theta;m,\mu)
\dif\theta - \left[\frac{d}{2}\log(2\pi) - \frac{1}{2}\log \det 
V_{\theta_0}+o_P(1)\right]
\label{eq:fn1}
\end{align}
The first equality is by the definition of \gls{KL} divergence. The
second equality is by the definition of the \gls{VB} ideal. The third
equality is due to the Laplace approximation of the normalizer like
we did in \Cref{eq:normalizerapprox}. The fourth equality is due to
the cancellation of the two $\log\det(\delta_n)$ terms. This again
exemplifies why we assume a fixed variational family on the rescale
variable $\check{\theta}$. The fifth equality is due to a similar
argument as in \Cref{eq:normalintapprox}. The sixth equality is due
to the \gls{LAN} condition of $M_n(\theta\s x)$. The seventh equality
is due to the computation of each term in the integral as an
expectation under the distribution $q(\theta)$. To extend the
restriction to some compact set $K$ to the whole space $\mathbb{R}^d$
in the sixth equality, we employ the same argument as in
\Cref{eq:relaxcompact}.

We notice that when $\mu\ne\theta_0$, we will have
$F_n(m)\rightarrow\infty$. On the other hand, we have $\limsup F_n
<\infty$. This echoes our consistency result in
\Cref{lemma:ivbconsist}.

Now we rewrite $F_0(m,\mu)$.
\begin{align}
&\gls{KL}(q(\theta;m,\mu)||\cN(\theta;\theta_0+\delta_n
\Delta_{n,\theta_0},
\delta_nV_{\theta_0}^{-1}\delta_n))\\
=&\log|\det(\delta_n)|^{-1} +\sum^d_{i=1}\mathbb{H}(q_{h,i}(h;m)) 
+\int q(\theta;m,\mu) \log \cN(\theta;\theta_0+\delta_n
\Delta_{n,\theta_0},
\delta_nV_{\theta_0}^{-1}\delta_n)\dif \theta\\
=&\log|\det(\delta_n)|^{-1} +\sum^d_{i=1}\mathbb{H}(q_{h,i}(h;m)) 
+\frac{d}{2}\log(2\pi) - \frac{1}{2}\log \det V_{\theta_0}
+\log\det(\delta_n)\nonumber\\
&+ \int q(\theta;m,\mu)\cdot (\theta - \theta_0
-\delta_n\Delta_{n,\theta_0})^\top \delta_n^{-1}
V_{\theta_0}\delta_n^{-1}(\theta - 
\theta_0-\delta_n\Delta_{n,\theta_0})\dif \theta\\
=&\sum^d_{i=1}\mathbb{H}(q_{h,i}(h;m)) +\frac{d}{2}\log(2\pi) 
- \frac{1}{2}\log \det V_{\theta_0} + 
\Delta_{n,\theta_0}^\top V_{\theta_0}\Delta_{n,\theta_0} \nonumber\\
&- \int \delta_n^{-1}(\theta-\theta_0)^\top V_{\theta_0}
\Delta_{n,\theta_0}\cdot q(\theta;m,\mu)\dif \theta + 
\int \frac{1}{2}(\delta_n^{-1}(\theta-\theta_0))^\top V_{\theta_0} 
\delta_n^{-1}(\theta-\theta_0)\cdot q(\theta;m,\mu)\dif\theta.
\end{align}

This gives 
\begin{align}
&F_0(m,\mu) - \Delta_{n,\theta_0}^\top V_{\theta_0}\Delta_{n,\theta_0}\\
= &\sum^d_{i=1}\mathbb{H}(q_{h,i}(h;m)) -\frac{d}{2}\log(2\pi) 
+ \frac{1}{2}\log \det V_{\theta_0} \nonumber\\
&- \int \delta_n^{-1}(\theta-\theta_0)^\top V_{\theta_0}
\Delta_{n,\theta_0}\cdot q(\theta;m,\mu)\dif \theta\nonumber\\
& + \int \frac{1}{2}(\delta_n^{-1}(\theta-\theta_0))^\top 
V_{\theta_0} \delta_n^{-1}(\theta-\theta_0)\cdot q(\theta;m,\mu)
\dif\theta\\
=&+\infty\cdot(1-\mathbb{I}_{\mu}(\theta_0)) + 
[\sum^d_{i=1}\mathbb{H}(q_{h,i}(h;m)) -\frac{d}{2}\log(2\pi) 
+ \frac{1}{2}\log \det V_{\theta_0}\nonumber\\
&- \int \delta_n^{-1}(\theta-\theta_0)^\top V_{\theta_0}
\Delta_{n,\theta_0}\cdot q(\theta;m,\mu)\dif \theta \nonumber\\
&+ \int \frac{1}{2}(\delta_n^{-1}(\theta-\theta_0))^\top 
V_{\theta_0} \delta_n^{-1}(\theta-\theta_0)\cdot q(\theta;m,\mu)
\dif\theta]\cdot 
\mathbb{I}_{\mu}(\theta_0).
\label{eq:f01}
\end{align}
The last step is due to our definition of our variational family
$q(\theta;m,\mu) = \prod^d_{i=1}\delta_{n,ii}^{-1}q_{h,i}(h;m),$
where $h = \delta_n^{-1}(\theta - \mu)$ for some $\mu\in\Theta$. The
last step is true as long as the $q_{h,i}$ distributions are not
point masses at zero. This is ensured by positive entropy in
\Cref{assumption:scoreint}.

Comparing \Cref{eq:fn1} and \Cref{eq:f01}, we can prove the $\Gamma$
convergence. Let $m_n\rightarrow m$.
When $\mu\ne\theta_0$, $\liminf_{n\rightarrow \infty}F_n(m_n,\mu)
= +\infty$. The limsup inequality is automatically satisfied. When $\mu
= \theta_0$, we have $F_n(m,\mu) =
F_0(m,\mu)-\Delta_{n,\theta_0}^\top
V_{\theta_0}\Delta_{n,\theta_0}+o_P(1)$. This implies $F_0(m,\mu)
\leq \lim_{n\rightarrow \infty}F_n(m_n,\mu)$ in $P_{\theta_0}$
probability by the continuity of $F_n$ ensured by 
\Cref{assumption:scoreint}.

We then show the existence of a recovery sequence. When
$\mu\ne\theta_0$, $F_0(m,\mu) = +\infty$. The limsup inequality is
automatically satisfied. When $\mu = \theta_0$, we can simply choose
$m_n = \theta_0$. The limsup inequality is
again ensured by $F_0(m,\mu) \leq \lim_{n\rightarrow
\infty}F_n(m_n,\mu)$ in $P_{\theta_0}$ probability and the
continuity of $F_n$. The $\Gamma$-convergence of the $F$ functionals
is shown.

We notice that $\Delta_{n,\theta_0}^\top
V_{\theta_0}\Delta_{n,\theta_0}$ does not depend on $m$ or $\mu$ 
so that $\argmin F_0 = $ $\argmin F_0 - \Delta_{n,\theta_0}^\top
V_{\theta_0}\Delta_{n,\theta_0}$. The convergence of the \gls{KL}
minimizers is thus proved.

\section{Proof of \Cref{lemma:vb_post}}
\label{sec:vbpostproof}

Notice that the mean field variational families
$\mathcal{Q}^d=\{q:q(\theta) =
\prod^d_{i=1}\delta_{n,ii}^{-1}q_{h,i}(h),$ where $h =
\delta_n^{-1}(\theta - \mu)$ for some $\mu\in\Theta\}$, or the
Gaussian family $\{q:q(\theta) = N(m, \delta_n\Sigma\delta_n)\}$ can
be written in the form of
\[q(\theta) = |\det \delta_n|^{-1} q_h(\delta_n^{-1}(\theta-\mu))\]
for some $\mu\in\mathbb{R}^d$, and $\int q_h(h)\dif h = 0.$ This form
is due to a change-of-variable step we detailed in the beginning of
\Cref{sec:ivbconsistproof}.

We first specify the mild technical conditions on $\mathcal{Q}^d$.\\
\begin{assumption} We assume the following conditions on $q_h$.
\label{assumption:vbpost_tech}
\begin{enumerate}
  \item If $q_h$ is has zero mean, we assume $\int h^2\cdot
  q_h(h)\dif h<\infty$ and $\sup_{z,x} |(\log p(z,x\mid\theta))''
  |\leq C_{11}\cdot q_h(\theta)^{-C_{12}} $ for some $C_{11}, C_{12}
  > 0;$ $|M_n(\theta\s x)''| \leq C_{13}\cdot q_h(\theta)^{-C_{14}}$
  for some $C_{13}, C_{14} > 0.$

  \item If $q_h$ has nonzero mean, we assume $\int h\cdot q_h(h)\dif
  h<\infty$ and $\sup_{z,x} |(\log p(z,x\mid\theta))' |\leq
  C_{11}\cdot q_h(\theta)^{-C_{12}} $ for some $C_{11}, C_{12} > 0$;
  $|M_n(\theta\s x)' \leq C_{13}|\cdot q_h(\theta)^{-C_{14}}$ for
  some $C_{13}, C_{14} > 0.$\\
\end{enumerate}
\end{assumption}

The assumption first assumes finite moments for $q_h$ so that we can
properly apply a Taylor expansion argument. The second part of this
assumption makes sure the derivative of $\log p(z,x\mid\theta)$ does
not increase faster than the tail decrease of $q_h(\cdot)$. For
example, if $q_h(\cdot)$ is normal, then the second part writes
$\sup_{z,x} |(\log p(z,x\mid\theta))'' |\leq C_{15}\exp(\theta^2)$
for some $C_{15} > 0$, and $|M_n(\theta\s x)''| \leq
C_{13}\exp(\theta^2)$ for some $C_{13} > 0$. The latter is satisfied
by the \gls{LAN} condition. This is in general a rather weak
condition. We usually would not expect the derivative of $\log
p(z,x\mid\theta)$ and $M_n(\theta\s x)$ to increase this fast as
$\theta$ increases.

Now we are ready to prove the lemma.

We first approximate the profiled \gls{ELBO}, $\gls{ELBO}_p(q(\theta))$.
\begin{align}
&\gls{ELBO}_p(q(\theta))\\
:=&\sup_{q(z)}
  \int q(\theta)
  \left(
  \log \left[p(\theta)
  \exp\left\{
  \int q(z) \log \frac{p(x, z|\theta)}{q(z)}
  \dif z\right\}\right]
  - \log q(\theta) \right) \dif \theta,\\
=&\int q(\theta)\log p(\theta)\dif \theta - \int q(\theta)\log q(\theta)
\dif \theta + \sup_{q(z)}\int q(\theta)\int q(z) 
\log \frac{p(x, z|\theta)}{q(z)}
  \dif z\dif \theta\\
=&\int q(\theta)\log p(\theta)\dif \theta - \int q(\theta)
\log q(\theta)\dif \theta\nonumber\\
& + \sup_{q(z)}\int q(\theta)[\int q(z) \log 
\frac{p(x, z|\mu)}{q(z)}\dif z + (\theta-\mu) \left(\int q(z) 
\log \frac{p(x, z|\mu)}{q(z)}\dif z\right)'\nonumber\\
=&\int q(\theta)\log p(\theta)\dif \theta - \int q(\theta)
\log q(\theta)\dif \theta + \sup_{q(z)}\int q(z) \log 
\frac{p(x, z|\mu)}{q(z)}\dif z\nonumber\\
&+\int q(\theta)\frac{1}{2}|\theta-\mu|^2\left(\int q(z) 
\log \frac{p(x, z|\tilde{\theta}^\dagger)}{q(z)}\dif z\right)'']
\dif \theta\\
\geq & \int q(\theta)\log p(\theta)\dif \theta - \int q(\theta)
\log q(\theta)\dif \theta + \sup_{q(z)}\int q(z) 
\log \frac{p(x, z|\mu)}{q(z)}\dif z\nonumber\\
&-C_{15}\int q(\theta)\frac{1}{2}|\theta-\mu|^2 
q_h(\theta)^{-C_{12}}\dif \theta\\
= & \int q(\theta)\log p(\theta)\dif \theta - 
\int q(\theta)\log q(\theta)\dif \theta + 
\sup_{q(z)}\int q(z) \log \frac{p(x, z|\mu)}{q(z)}\dif z\nonumber\\
&-C_{15}\int q_h(h)\frac{1}{2}|\delta_nh|^2 
q_h(\mu+\delta_nh)^{-C_{12}}\dif h\\
\geq & \int q(\theta)\log p(\theta)\dif \theta - 
\int q(\theta)\log q(\theta)\dif \theta + 
\sup_{q(z)}\int q(z) \log \frac{p(x, z|\mu)}{q(z)}\dif z\nonumber\\
&-C_{16}\min_i(\delta_{n,ii}^2)\\
=& \int q(\theta)\log p(\theta)\dif \theta - 
\int q(\theta)\log q(\theta)\dif \theta + 
\sup_{q(z)}\int q(z) \log \frac{p(x, z|\mu)}{q(z)}\dif z + o(1),
\end{align}
for some constant $C_{16} > 0.$ The first equality is by the definition
of $\gls{ELBO}_p(q(\theta))$. The second equality is rewriting the
integrand. The third equality is due to mean value theorem where
$\tilde{\theta}^\dagger$ is some value between $\mu$ and $\theta$. (A
very similar argument for $q_h$ with nonzero means can be made
starting from here, that is expanding only to the first order term.)
The fourth equality is due to $q_h(\cdot)$ having zero mean. The
fifth inequality is due to the second part of
\Cref{assumption:vbpost_tech}. The sixth inequality is due to a
change of variable $h = \delta_n(\theta-\mu)$. The seventh inequality
is due to $q_h(\cdot)$ residing within exponential family with finite
second moment (the first part of \Cref{assumption:vbpost_tech}). The
eighth equality is due to $\delta_n\rightarrow 0$ as
$n\rightarrow\infty.$

We now approximate $-\gls{KL}(q(\theta)||\pi^*(\theta|x))$ in a
similar way.
\begin{align}
&-\gls{KL}(q(\theta)||\pi^*(\theta|x))\\
=&\int q(\theta)\log\frac{p(\theta)\exp(M_n(\theta\s x))}{q(\theta)}
\dif \theta\\
=&  \int q(\theta)
  \left(
  \log \left[p(\theta)
  \exp\left\{\sup_{q(z)}
  \int q(z) \log \frac{p(x, z|\theta)}{q(z)}
  \right\}\right]
  - \log q(\theta) \right) d\theta\\
=&\int q(\theta)\log p(\theta)\dif \theta - 
\int q(\theta)\log q(\theta)\dif \theta + 
\int q(\theta)[\sup_{q(z)}\int q(z) \log 
\frac{p(x, z|\mu)}{q(z)}\dif z\nonumber\\
& + (\theta-\mu) \left(\sup_{q(z)}\int q(z) 
\log \frac{p(x, z|\mu)}{q(z)}\dif z\right)' +
\frac{1}{2}|\theta-\mu|^2\left(\sup_{q(z)}\int q(z) 
\log \frac{p(x, z|\tilde{\tilde{\theta}}^\dagger)}{q(z)}
\dif z\right)'']\dif \theta\\
=&\int q(\theta)\log p(\theta)\dif \theta - \int q(\theta)
\log q(\theta)\dif \theta + \sup_{q(z)}\int q(z) 
\log \frac{p(x, z|\mu)}{q(z)}\dif z\nonumber\\
&+\int q(\theta)\frac{1}{2}|\theta-\mu|^2\left(\sup_{q(z)}
\int q(z) \log \frac{p(x, z|\tilde{\tilde{\theta}}^\dagger)}{q(z)}
\dif z\right)'']\dif \theta\\
\leq & \int q(\theta)\log p(\theta)\dif \theta - \int q(\theta)
\log q(\theta)\dif \theta + \sup_{q(z)}\int q(z) 
\log \frac{p(x, z|\mu)}{q(z)}\dif z\nonumber\\
&+C_{17}\int q(\theta)\frac{1}{2}|\theta-\mu|^2 q_h(\theta)^{-C_{14}}
\dif \theta\\
= & \int q(\theta)\log p(\theta)\dif \theta - 
\int q(\theta)\log q(\theta)\dif \theta + 
\sup_{q(z)}\int q(z) \log \frac{p(x, z|\mu)}{q(z)}\dif z\nonumber\\
&+C_{17}\int q_h(h)\frac{1}{2}|\delta_nh|^2 
q_h(\mu+\delta_nh)^{-C_{14}}\dif h\\
\leq & \int q(\theta)\log p(\theta)\dif \theta - 
\int q(\theta)\log q(\theta)\dif \theta + \sup_{q(z)}\int q(z) 
\log \frac{p(x, z|\mu)}{q(z)}\dif z+C_{17}\max_i(\delta_{n,ii}^2)\\
=& \int q(\theta)\log p(\theta)\dif \theta - \int q(\theta)
\log q(\theta)\dif \theta + \sup_{q(z)}\int q(z) 
\log \frac{p(x, z|\mu)}{q(z)}\dif z + o(1)
\end{align}
for some constant $C_{17} > 0$. The first equality is by the definition
of \gls{KL} divergence. The second equality is rewriting the
integrand. The third equality is due to mean value theorem where
$\tilde{\tilde{\theta}}^\dagger$ is some value between $\mu$ and
$\theta$. (A very similar argument for $q_h$ with nonzero means can
be made starting from here, that is expanding only to the first order
term.) The fourth equality is due to $q_h(\cdot)$ having zero mean.
The fifth inequality is due to the third part of
\Cref{assumption:vbpost_tech}. The sixth inequality is due to a
change of variable $h = \delta_n(\theta-\mu)$. The seventh inequality
is due to $q_h(\cdot)$ residing within exponential family with finite
second moment (the first part of \Cref{assumption:vbpost_tech}). The
eighth equality is due to $\delta_n\rightarrow 0$ as
$n\rightarrow\infty.$

Combining the above two approximation, we have
$-\gls{KL}(q(\theta)||\pi^*(\theta|x))\leq
\gls{ELBO}_p(q(\theta))+o(1)$. On the other hand, we know that
$\gls{ELBO}_p(q(\theta))\leq -\gls{KL}(q(\theta)||\pi^*(\theta|x))$
by definition. We thus conclude $\gls{ELBO}_p(q(\theta))=
-\gls{KL}(q(\theta)||\pi^*(\theta|x))+o_P(1).$

\section{Proof of \Cref{thm:main} and \Cref{thm:estimate}}
\label{sec:mainproof}
\Cref{thm:main} is a direct consequence of \Cref{lemma:ivbconsist},
\Cref{lemma:ivbnormal}, and \Cref{lemma:vb_post}.
\Cref{lemma:ivbconsist} and \Cref{lemma:ivbnormal} characterizes the
consistency and asymptotic normality of the \gls{KL} minimizer of the
\gls{VB} ideal. \Cref{lemma:vb_post} says the
\gls{VB} posterior shares the same asymptotic properties as the
\gls{VB} ideal. All of them together give the consistency and
asymptotic normality of \gls{VB} posteriors.

\Cref{thm:estimate} is a consequence of a slight generalization of
Theorem 2.3 of \citet{kleijn2012bernstein}. The theorem characterizes
the consistency and asymptotic normality of the posterior mean
estimate under model specification with a common
$\sqrt{n}$-convergence rate. We only need to replace all $\sqrt{n}$
by $\delta_n^{-1}$ in their proof to obtain the generalization.

Specifically, we first show that, for any $M_n\rightarrow\infty$,
\[\int_{||\tilde{\theta}||>M_n}||\tilde{\theta}||^2 
q^*(\tilde{\theta})\dif\tilde{\theta}
\stackrel{P_{\theta_0}}{\rightarrow}0.\]
This is ensured by \Cref{assumption:vbpost_tech}.1.

We then consider three stochastic processes: fix some compact set $K$
and for given $M>0$,
\begin{align}
t&\mapsto Z_{n,M}(t) = \int_{||\tilde{\theta}||\leq M} 
(t-\tilde{\theta})^2\cdot q^*_{\tilde{\theta}}(\tilde{\theta})
\dif \tilde{\theta},\\
t&\mapsto W_{n,M}(t) = \int_{||\tilde{\theta}||\leq M} 
(t-\tilde{\theta})^2\cdot \cN(\tilde{\theta};\Delta_{n,\theta_0}, 
V_{\theta_0}^{-1})\dif \tilde{\theta},\\
t&\mapsto W_{M}(t) = \int_{||\tilde{\theta}||\leq M} 
(t-\tilde{\theta})^2\cdot \cN(\tilde{\theta};X, V_{\theta_0}^{-1})
\dif \tilde{\theta}.
\end{align}

We note that $\theta_n^*$ is the minimizer of $t\mapsto
Z_{n,\infty}(t)$ and $\int \tilde{\theta} \cdot
\argmin_{q\in\mathcal{Q}^d}
\gls{KL}(q(\tilde{\theta})|| \cN(\tilde{\theta} \s X,
V^{-1}_{\theta_0} ))
\dif\tilde{\theta}$ is the minimizer of $t\mapsto W_{\infty}(t)$.

By $\sup_{t\in K, ||h||\leq M}(t-h)^2<\infty$, we have
$Z_{n,M}-W_{n,M}=o_{P_{\theta_0}}(1)$ in $\ell^\infty(K)$ by
\Cref{thm:main}. Since
$\Delta_{n,\theta_0}\stackrel{d}{\rightarrow}X$, the continuous
mapping theorem implies that $W_{n,M} - W_{M}=o_{P_{\theta_0}}(1)$ in
$\ell^\infty(K)$. By $\int \theta \cdot q^*(\theta)<\infty$, we have
$W_M - W_{\infty} = o_{P_{\theta_0}}(1)$ as $M\rightarrow\infty.$ We
conclude that there exists a sequence $M_n\rightarrow\infty$ such
that $Z_{n,M_n}-W_\infty = o_{P_{\theta_0}}(1)$. We also have from
above that $Z_{n,M_n} - Z_{n,\infty} = o_{P_{\theta_0}}(1)$ in
$\ell^\infty(K)$. We conclude that $Z_{n,\infty} - W_\infty =
o_{P_{\theta_0}}(1)$ in $\ell^\infty(K)$. By the continuity and
convexity of the squared loss, we invoke the argmax theorem and
conclude that $\theta_n^*$ converges weakly to $\int \tilde{\theta}
\cdot \argmin_{q\in\mathcal{Q}^d}
\gls{KL}(q(\tilde{\theta})|| \cN(\tilde{\theta} 
\s X, V^{-1}_{\theta_0} ))
\dif\tilde{\theta}$.

\section{Proof of \Cref{corollary:fullrankgaussian}}
\label{sec:fullrankgaussianproof}

We prove \Cref{lemma:ivbnormal} for the Gaussian family. We want
to show that the functionals
\[F_n(m,\Sigma):=\gls{KL}(\cN(\theta;m,\delta_n\Sigma\delta_n)||
\pi^*(\theta\mid x))\]
$\Gamma$-converge to
\[F_0(m,\Sigma):=\gls{KL}(\cN(\theta;m,\delta_n\Sigma\delta_n)||
\cN(\theta;\theta_0+\delta_n\Delta_{n,\theta_0},
\delta_nV_{\theta_0}^{-1}\delta_n)) - \Delta_{n,\theta_0}^\top 
V_{\theta_0}\Delta_{n,\theta_0}\] in probability as $n\rightarrow 0$.
We note that this is equivalent to the $\Gamma$-convergence of the
functionals of $q$:
$\gls{KL}(q(\cdot)||\pi^*_{\tilde{\theta}}(\cdot\mid x))$ to
$\gls{KL}(q(\cdot)||
\cN(\cdot \s\Delta_{n,\theta_0}, V^{-1}_{\theta_0}))$. This is
because the second statement is the same as the first up to a change
of variable step from $\theta$ to $\tilde{\theta}$.

Together with
\Cref{thm:gammaconv}, this implies $m_n, \Sigma_n
\stackrel{P_{\theta_0}}{\rightarrow} m_0, \Sigma_0$ where $(m_n,
\Sigma_n)$ is the minimizer of $F_n$ for each $n$ and $(m_0,
\Sigma_0)$ is the minimizer of $F_0$. This is due to the last term of
$F_0$ -- $\Delta_{n,\theta_0}^\top V_{\theta_0}\Delta_{n,\theta_0}$
is a constant bounded in $P_{\theta_0}$ probability and independent
of $m, \Sigma$. The convergence in total variation then follows
from Lemma 4.9 of \citet{klartag2007central}, which gives an upper
bound on \gls{TV} distance between two Gaussian distributions.

Now we prove the $\Gamma$-convergence.

We first rewrite $F_n(m,\Sigma)$.
\begin{align}
&F_n(m,\Sigma)\\
:=&\gls{KL}(\cN(\theta;m,\delta_n\Sigma\delta_n)||
\pi^*(\theta\mid x))\\
=&\int \cN(\theta;m,\delta_n\Sigma\delta_n)\log 
\cN(\theta;m,\delta_n\Sigma\delta_n)\dif \theta - 
\int \cN(\theta;m,\delta_n\Sigma\delta_n) \log 
\pi^*(\theta\mid x))\dif \theta\\
=& -\frac{d}{2} - \frac{d}{2}\log(2\pi) - \log 
\det(\delta_n) - \frac{1}{2}\log \det(\Sigma) - 
\int \cN(\theta;m,\delta_n\Sigma\delta_n)\log p(\theta)
\dif \theta\nonumber\\
&-\int \cN(\theta;m,\delta_n\Sigma\delta_n)M_n(\theta\s x)
\dif \theta + \log \int p(\theta)\exp(M_n(\theta\s x))\dif \theta\\
=&-\frac{d}{2} - \frac{d}{2}\log(2\pi) - 
\log \det(\delta_n) - \frac{1}{2}\log \det(\Sigma) - 
\int \cN(\theta;m,\delta_n\Sigma\delta_n)\log p(\theta)
\dif \theta\nonumber\\
&-\int \cN(\theta;m,\delta_n\Sigma\delta_n)M_n(\theta\s x)
\dif \theta\nonumber\\
& + \left[\frac{d}{2}\log(2\pi) - \frac{1}{2}\log \det 
V_{\theta_0}+\log\det(\delta_n)+M_n(\theta_0\s x)+
\log p(\theta_0)+o_P(1)\right]\label{eq:canceldelta2}\\
=&-\frac{d}{2} - \frac{1}{2}\log \det(\Sigma)- 
\int \cN(\theta;m,\delta_n\Sigma\delta_n)\log p(\theta)
\dif \theta\nonumber\\
&-\int \cN(\theta;m,\delta_n\Sigma\delta_n)M_n(\theta\s x)
\dif \theta+ \left[- \frac{1}{2}\log \det V_{\theta_0}+
M_n(\theta_0\s x)+\log p(\theta_0)+o_P(1)\right]\\
=&-\frac{d}{2} - \frac{1}{2}\log \det(\Sigma)-
\int \cN(\theta;m,\delta_n\Sigma\delta_n)M_n(\theta\s x)
\dif \theta\nonumber\\
& + \left[- \frac{1}{2}\log \det V_{\theta_0}+
M_n(\theta_0\s x)+o_P(1)\right]\\
=&-\frac{d}{2} - \frac{1}{2}\log \det(\Sigma)-
\int \cN(\theta;m,\delta_n\Sigma\delta_n)
[ M_n(\theta_0\s x) + \delta_n^{-1}(\theta-\theta_0)^\top 
V_{\theta_0}\Delta_{n,\theta_0}\nonumber\\
&-\frac{1}{2}(\delta_n^{-1}(\theta-\theta_0))^\top V_{\theta_0} 
\delta_n^{-1}(\theta-\theta_0)+o_P(1)]\dif \theta - 
\left[ \frac{1}{2}\log \det V_{\theta_0}-
M_n(\theta_0\s x)+o_P(1)\right]\\
=& -\frac{d}{2} - \frac{1}{2}\log \det(\Sigma)- 
\delta_n^{-1}(m-\theta_0)^\top V_{\theta_0}\Delta_{n,\theta_0} + 
\frac{1}{2}(\delta_n^{-1}(m-\theta_0))^\top V_{\theta_0} 
\delta_n^{-1}(m-\theta_0) \nonumber\\
&+ \frac{1}{2}Tr(V_{\theta_0}\cdot \Sigma) -  
\frac{1}{2}\log \det V_{\theta_0}+o_P(1).
\label{eq:fn}
\end{align}
The first equality is by the definition of \gls{KL} divergence. The
second equality is calculating the entropy of multivariate Gaussian
distribution. The third equality is due to the Laplace approximation
of the normalizer like we did in \Cref{eq:normalizerapprox}. The
fourth equality is due to the cancellation of the two
$\log\det(\delta_n)$ terms. This again exemplifies why we assume the
family to have variance $\delta_n\Sigma\delta_n$. The fifth equality
is due to a similar argument as in \Cref{eq:normalintapprox}. The
intuition is that $\cN(m, \delta_n\Sigma\delta_n)$ converges to a
point mass as $n\rightarrow \infty.$ The sixth equality is due to the
\gls{LAN} condition of $M_n(\theta\s x)$. The seventh equality is due
to the computation of each term in the integral as an expectation
under the Gaussian distribution $\cN(m, \delta_n\Sigma\delta_n)$. To
extend the restriction to some compact set $K$ to the whole space
$\mathbb{R}^d$ in the sixth equality, we employ the same argument as
in \Cref{eq:relaxcompact}.

We notice that when $m\ne\theta_0$, we will have
$F_n(m,\Sigma)\rightarrow\infty$. On the other hand, we have $\limsup
F_n <\infty$. This echoes our consistency result in
\Cref{lemma:ivbconsist}.

Now we rewrite $F_0(m,\Sigma)$.
\begin{align}
&\gls{KL}(\cN(\theta;m,\delta_n\Sigma\delta_n)||
\cN(\theta;\theta_0+\delta_n\Delta_{n,\theta_0},
\delta_nV_{\theta_0}^{-1}\delta_n))\\
=&-\frac{d}{2}+\frac{1}{2}Tr(V_{\theta_0}\cdot\Sigma) + 
(m - \theta_0-\delta_n\Delta_{n,\theta_0})^\top 
\delta_n^{-1}V_{\theta_0}\delta_n^{-1}(m - 
\theta_0-\delta_n\Delta_{n,\theta_0})\nonumber\\
& - \frac{1}{2}\log \det(\Sigma) -  
\frac{1}{2}\log \det V_{\theta_0}\\
=&-\frac{d}{2} - \frac{1}{2}\log \det(\Sigma)- 
\delta_n^{-1}(m-\theta_0)^\top V_{\theta_0}
\Delta_{n,\theta_0} + \frac{1}{2}(\delta_n^{-1}
(m-\theta_0))^\top V_{\theta_0} \delta_n^{-1}
(m-\theta_0) \nonumber\\
&+ \frac{1}{2}Tr(V_{\theta_0}\cdot \Sigma) -  
\frac{1}{2}\log \det V_{\theta_0} + 
\Delta_{n,\theta_0}^\top V_{\theta_0}\Delta_{n,\theta_0}.
\end{align}

This gives 
\begin{align}
&F_0(m,\Sigma) - \Delta_{n,\theta_0}^\top 
V_{\theta_0}\Delta_{n,\theta_0}\\
= & -\frac{d}{2} - \frac{1}{2}\log \det(\Sigma)- 
\delta_n^{-1}(m-\theta_0)^\top V_{\theta_0}
\Delta_{n,\theta_0} + \frac{1}{2}(\delta_n^{-1}(m-\theta_0))^\top 
V_{\theta_0} \delta_n^{-1}(m-\theta_0) \nonumber\\
&+ \frac{1}{2}Tr(V_{\theta_0}\cdot \Sigma) -  
\frac{1}{2}\log \det V_{\theta_0}\\
=&+\infty\cdot(1-\mathbb{I}_{m}(\theta_0)) + 
[-\frac{d}{2} - \frac{1}{2}\log \det(\Sigma)+ 
\frac{1}{2}Tr(V_{\theta_0}\cdot \Sigma) -  
\frac{1}{2}\log \det V_{\theta_0}]\cdot \mathbb{I}_{m}(\theta_0).
\label{eq:f0}
\end{align}

This equality is due to the \gls{KL} divergence between two
multivariate Gaussian distributions.

Comparing \Cref{eq:fn} and \Cref{eq:f0}, we can prove the $\Gamma$
convergence. Let $m_n\rightarrow m$ and $\Sigma_n\rightarrow \Sigma$.
When $m\ne\theta_0$, $\liminf_{n\rightarrow \infty}F_n(m_n,\Sigma_n)
= +\infty$. The limsup inequality is automatically satisfied. When $m
= \theta_0$, we have $F_n(m,\Sigma) =
F_0(m,\Sigma)-\Delta_{n,\theta_0}^\top
V_{\theta_0}\Delta_{n,\theta_0}+o_P(1)$. This implies $F_0(m, \Sigma)
\leq \lim_{n\rightarrow \infty}F_n(m_n,\Sigma_n)$ in $P_{\theta_0}$
probability by the continuity of $F_n$.

We then show the existence of a recovery sequence. When
$m\ne\theta_0$, $F_0(m,\Sigma) = +\infty$. The limsup inequality is
automatically satified. When $m = \theta_0$, we can simply choose
$\Sigma_m = \Sigma$ and $m_n = \theta_0$. The limsup inequality is
again ensured by $F_0(m, \Sigma) \leq \lim_{n\rightarrow
\infty}F_n(m_n,\Sigma_n)$ in $P_{\theta_0}$ probability and the
continuity of $F_n$. The $\Gamma$-convergence of the $F$ functionals
is shown.

We notice that $\Delta_{n,\theta_0}^\top
V_{\theta_0}\Delta_{n,\theta_0}$ does not depend on any of $m,
\Sigma$ so that $\argmin F_0 = \argmin F_0 - \Delta_{n,\theta_0}^\top
V_{\theta_0}\Delta_{n,\theta_0}$. The convergence of the \gls{KL}
minimizers is thus proved.

\section{Proof of \Cref{lemma:normalkl}}
\label{sec:normalklproof}
 We first note that 
 \begin{align}
 &\gls{KL}(\cN(\cdot;\mu_0,\Sigma_0||\cN(\cdot;\mu_1,\Sigma_1))\\
 =& \frac{1}{2}[Tr(\Sigma_1^{-1}\Sigma_0)+(\mu_1-\mu_0)^\top 
 \Sigma_1^{-1}(\mu_1-\mu_0)-d+\log\frac{\det(\Sigma_1)}{\det(\Sigma_0)}].
 \end{align}

 Clearly, the optimal choice of $\mu_0$ is $\hat{\mu}_0=\mu_1$. Next,
 we write $\Sigma_0 = \diag(\lambda_1, ..., \lambda_d)$. The \gls{KL}
 divergence minimization objective thus becomes
 \[\frac{1}{2}[\sum_{i=1}^d (\Sigma_1^{-1})_{ii}\lambda_i + 
 \log \det(\Sigma_1)-\sum_{i=1}^d\log \lambda_i].\]
 Taking its derivative with respect to each $\lambda_i$ and setting
 it to zero, we have
 \[(\Sigma_1^{-1})_{ii} = \lambda_i^{-1}.\]
 The optimal $\Sigma_0$ thus should be diagonal with
 $\hat{\Sigma}_{0, ii} = ((\Sigma_1^{-1})_{ii})^{-1}$ for $i = 1, 2,
 ..., d$. In this sense, mean field (factorizable) approximation
 matches the precision matrix at the mode.

 Moreover, by the inequality 
 \citep{amir1969product,beckenbach2012inequalities}
 \[\det(\Sigma_1^{-1})\leq \prod_i (\Sigma_1^{-1})_{ii} = 
 \det(\hat{\Sigma}_0^{-1}),\]
we have \[\mathbb{H}(\hat{\Sigma}_0^{-1}) = \frac{1}{2}
\log((2\pi e)^d\cdot \det(\hat{\Sigma}_0))\leq \frac{1}{2}
\log((2\pi e)^d\cdot \det(\Sigma_1)) = \mathbb{H}(\Sigma_1).\]

\section{Proof of \Cref{prop:variationallan}}
\label{sec:lanproof}

For simplicity, we prove this proposition for the case when the local
latent variables $z$ are discrete. The continuous case is easily
adapted by replacing the joint probability of $(x, z)$ with the marginal probability of $x$ with $z$ constrained over a neighborhood around $z_{\text{profile}}$ shrinking to a point mass.

The proof relies on the following inequality:
\begin{align}
\label{eq:boundvii}
\log p(x, z_{\text{profile}}\mid \theta)\leq M_n(\theta\, ;\, x) \leq \log \int p(x, z\mid \theta)\dif z,
\end{align}
where $z_{\text{profile}}$ is the maximum profile likelihood estimate,
$z_{\text{profile}} := \argmax_z p(x, z\mid \theta_0)$. The lower bound is due to choosing the variational distribution as a point mass at $z_{\text{profile}}$. The upper bound is due to the Jensen's inequality.

Let $x$ be generated by $\int p(x, z\mid\theta=\theta_0)\dif z$. 
Condition 1 implies that the posterior of the
local latent variables given the true global latent variables
concentrates around $z_{\text{profile}}$: in $P_{\theta_0}$-probability,
\[p(z \mid x, \theta =\theta_0) \stackrel{d}{\rightarrow} \delta_{z_{\text{profile}}}.\]

This convergence result implies
\begin{align}
\label{eq:marginal}
\log p(x, z_{\text{profile}}\mid \theta_0) =\log \int p(x, z\mid \theta_0)\dif z + o_{P_{\theta_0}}(1),
\end{align}
when the local latent variables $z$ are discrete.
Hence,
\begin{align}
\label{eq:variational}
\log p(x, z_{\text{profile}}\mid \theta_0) = M_n(\theta_0\, ;\, x) + o_{P_{\theta_0}}(1).
\end{align}
Therefore, subtracting \Cref{eq:boundvii} by $M_n(\theta_0\, ;\, x)$ gives
\begin{align}
\label{eq:lanbound}
\log \frac{p(x, z_{\text{profile}}\mid \theta)}{p(x, z_{\text{profile}}\mid \theta_0)}  +
o_{P_{\theta_0}}(1)\leq  M_n(\theta\, ;\, x) - M_n(\theta_0\, ;\, x)
\leq \log \frac{\int p(x, z\mid \theta)\dif z}{\int p(x, z\mid
\theta_0)\dif z} + o_{P_{\theta_0}}(1).
\end{align}
The left inequality is due to \Cref{eq:variational}. The right
inequality is due to \Cref{eq:marginal} and \Cref{eq:variational}.

Finally recall that, with the two conditions, Theorem 4.2 of
\citet{bickel2012semiparametric} shows that the log marginal
likelihood has the same \gls{LAN} expansion as the complete log
likelihood in a Hellinger neighborhood around $z_{\text{profile}}$,
\begin{align}
\label{eq:samelan}
\log \frac{p(x, z_{\text{profile}}\mid \theta)}{p(x, z_{\text{profile}}\mid \theta_0)} = \log
\frac{\int p(x, z\mid \theta)\dif z}{\int p(x, z\mid \theta_0)\dif z}
+ o_{P_{\theta_0}}(1).
\end{align}
Together with \Cref{eq:lanbound}, we conclude the variational log
likelihood $M_n( \theta\, ;\,x)$ and the complete log likelihood
$\log p(x, z\mid \theta)$ have the same \gls{LAN} expansion around
$\theta = \theta_0.$

\section{Proof of \Cref{corollary:gmm}}
\label{sec:gmmproof}
We only need to verify the local asymptotic normality of $L_n(\mu\s
x)$ here. By Equation (2) of \citet{westling2015establishing}, we
know the variational log likelihood writes $L_n(\mu\s x) = \sum_i
m(\mu\s x_i)$. We Taylor-expand it around the true value $\mu_0$:
\begin{align}
&L_n(\mu_0 + \frac{s}{\sqrt{n}}\s x)\\
=&\sum_i m(\mu_0 + \frac{s}{\sqrt{n}}\s x_i) \\
=&\sum_i m(\mu_0\s x_i) + \frac{s}{\sqrt{n}}\sum_i 
D_\mu m(\mu_0\s x_i) + \frac{1}{n}s^\top 
[\sum_i D^2_\mu m(\mu_0\s x_i)]s.
\end{align}

Due to $X_i$'s being independent and identically distributed, we have
\[\frac{1}{\sqrt{n}}\sum_i D_\mu m(\mu_0\s x_i) = 
\sqrt{n}\cdot \frac{\sum_i D_\mu m(\mu_0\s x_i)}{n}
\stackrel{d}{\rightarrow}\cN(0, B(\mu))\]
under $P_{\mu_0}$, where $B(\mu) = 
\mathbb{E}_{P_{\mu_0}}[D_\mu m(\theta\s x)D_\mu m(\mu\s x)^\top]$.
The convergence in distribution is due to central limit theorem. The
mean zero here is due to conditions B2, B4, and B5 of
\citet{westling2015establishing} (See point 4 in the first paragraph
of Proof of Theorem 2 in \citet{westling2015establishing} for
details.)

By strong law of large numbers, we also have
\[\frac{1}{n}\sum_i D^2_\mu m(\mu_0\s x_i) 
\stackrel{P_{\theta_0}}{\rightarrow} 
\mathbb{E}_{P_{\theta_0}}[D^2_\mu m(\mu_0\s x_i)].\] 

This gives the local asymptotic normality, for $s$ in a compact set,
\[L_n(\mu_0 + \frac{s}{\sqrt{n}}\s x)=
L_n(\mu_0\s x) + s^\top \Sigma Y -\frac{1}{2}s^\top\Sigma s+o_P(1),\]
where 
\[Y\sim\cN(0, A(\mu_0)^{-1}B(\mu_0)A(\mu_0)^{-1}),\]
and
\[\Sigma = A(\mu_0) = \mathbb{E}_{P_{\theta_0}}[D^2_\mu m(\mu_0\s x_i)]\]
is a positive definite matrix.

The consistent testability assumption is satisfied by the existence
of consistent estimators. This is due to Theorem 1 of
\citet{westling2015establishing}.

The corollary then follows from \Cref{thm:main} and
\Cref{thm:estimate} in \Cref{sec:vbconsistency}.

\section{Proof of \Cref{corollary:glmm}}
\label{sec:glmmproof}
We first verify the local asymptotic normality of the variational log
likelihood:
\begin{align}
&\underline{\ell}(\beta, \sigma^2):=\sup_{\mu,\lambda}
\ell(\beta,\sigma^2,\mu, \lambda)\\
=& \sup_{\mu,\lambda}\sum^m_{i=1}\sum^n_{j=1}
\{Y_{ij(\beta_0+\beta_1X_{ij}+\mu_i) - 
\exp(\beta_0+\beta_1X_{ij}+\mu_i+\lambda_i/2)} -\log (Y_{ij}!)\}\nonumber\\
&-\frac{m}{2}\log(\sigma^2) + \frac{m}{2} - 
\frac{1}{2\sigma^2}\sum^n_{i=1}(\mu_i^2+\lambda_i)+
\frac{1}{2}\sum^m_{i=1}\log(\lambda_i).
\end{align}

We take the Taylor expansion of the variational log likelihood at the
true parameter values $\beta^0_0,\beta^0_1, (\sigma^2)^0$:
\begin{align}
&\underline{\ell}(\beta^0_0 + \frac{s}{\sqrt{m}}, 
\beta^0_1 + \frac{t}{\sqrt{mn}}, (\sigma^2)^0 + \frac{r}{\sqrt{m}})\\
=& \underline{\ell}(\beta^0_0,\beta^0_1, (\sigma^2)^0) +
\frac{s}{\sqrt{m}}\frac{\partial}{\partial\beta_0}
\underline{\ell}(\beta^0_0,\beta^0_1, (\sigma^2)^0) +  
\frac{t}{\sqrt{mn}}\frac{\partial}{\partial\beta_1}
\underline{\ell}(\beta^0_0,\beta^0_1, (\sigma^2)^0)\nonumber\\
& + \frac{r}{\sqrt{m}}\frac{\partial}{\partial\sigma^2}
\underline{\ell}(\beta^0_0,\beta^0_1, (\sigma^2)^0)+ 
\frac{s^2}{m}\frac{\partial^2}{\partial\beta_0^2}
\underline{\ell}(\beta^0_0,\beta^0_1, (\sigma^2)^0) +  
\frac{t^2}{mn}\frac{\partial^2}{\partial\beta_1^2}
\underline{\ell}(\beta^0_0,\beta^0_1, (\sigma^2)^0)\nonumber\\
& + \frac{r^2}{m}\frac{\partial^2}{\partial(\sigma^2)^2}
\underline{\ell}(\beta^0_0,\beta^0_1, (\sigma^2)^0) + o_P(1).
\end{align}

Denote $\hat{\mu}_i$ and $\hat{\lambda}_i$ as the optimal $\mu$ and
$\lambda$ at the true values $(\beta^0_0,\beta^0_1, (\sigma^2)^0)$.
Also write $Y_{i\cdot} = \sum^n_{j=1}Y_{ij}$ and
$B_i=\sum^n_{j=1}\exp(\beta_0+\beta_1X_{ij})$. Let $\hat{\beta}_1$,
$\hat{\beta}_0$, $\hat{\sigma}^2$ be the maximizers of the
$\underline{\ell}$. Hence, we write
$\hat{B}_i=\sum^n_{j=1}\exp(\hat{\beta}_0+\hat{\beta}_1X_{ij})$.
Finally, the moment generating function of $X$ writes $\phi(t) =
\mathbb{E}\{\exp(tX)\}$.

The cross terms are zero due to Equation (5.21), Equation (5.29),
Equation (5.37), and Equation (5.50) of \citet{hall2011asymptotic}.

We next compute each of the six derivatives terms.

For the first term $\frac{\partial}{\partial\beta_0}
\underline{\ell}(\beta^0_0,\beta^0_1, (\sigma^2)^0)$, we have
\begin{align}
&\frac{1}{\sqrt{m}}\frac{\partial}{\partial\beta_0}
\underline{\ell}(\beta^0_0,\beta^0_1, (\sigma^2)^0)\\
=&\frac{1}{\sqrt{m}}\sum^m_{i=1}[Y_{i\cdot}-B_i
\exp(\hat{\mu}_i+\frac{1}{2}\hat{\lambda}_i)]\\
=&\frac{1}{\sqrt{m}}\sum^m_{i=1}[Y_{i\cdot}-
\hat{B}_i\exp(\hat{\mu}_i+\frac{1}{2}\hat{\lambda}_i)]- 
\frac{1}{\sqrt{m}}(\beta_0 - \hat{\beta}_0)
\sum^m_{i=1}\hat{B}_i\exp(\hat{\mu}_i+\frac{1}{2}\hat{\lambda}_i)\nonumber\\
& - \frac{1}{\sqrt{m}}(\beta_1 - \hat{\beta}_1)
\sum^m_{i=1}\hat{B}_i\exp(\hat{\mu}_i+
\frac{1}{2}\hat{\lambda}_i) + o_P(1)\\
= &{\sqrt{m}}\frac{1}{m}\sum^m_{i=1}
\sum_{j=1}^n((\beta_0 - \hat{\beta}_0)+
(\beta_1 - \hat{\beta}_1)X_{ij})\exp(\hat{\beta}_0+
\hat{\beta}_1X_{ij})\exp(\hat{\mu}_i+
\frac{1}{2}\hat{\lambda}_i) + o_P(1)\\
=&\sqrt{m}\frac{1}{m}\sum^m_{i=1}
\sum_{j=1}^n((\beta_1 - \hat{\beta}_1)(X_{ij}-\gamma(\beta_1^0))+
\bar{U})\exp(\hat{\beta}_0+\hat{\beta}_1X_{ij})\exp(\hat{\mu}_i+
\frac{1}{2}\hat{\lambda}_i) + o_P(1)\\
=&\sqrt{m}\bar{U}\frac{1}{m}\sum^m_{i=1}\hat{B}_i\exp(\hat{\mu}_i+
\frac{1}{2}\hat{\lambda}_i) + o_P(1)\\
=&\sqrt{m}\bar{U}\frac{1}{m}\sum^m_{i=1}\hat{B}_i\exp(\hat{\mu}_i+
\frac{1}{2}\hat{\lambda}_i) + o_P(1)\\
\stackrel{d}{\rightarrow}&\cN(0, (\sigma^2)^0)[\exp(\beta^0_0)
\mathbb{E}\exp(\beta_1X_{ij}+U_i)] + o_P(1)\\
=&\cN(0, (\sigma^2)^0)[\exp(\beta_0^0-\frac{1}{2}(\sigma^2)^0)
\phi(\beta_1^0)].
\end{align}
The first equality is due to differentiation with respect to
$\beta_0$. The second equality is due to Taylor expansion around the
\gls{VFE}. The third equality is due to Equation (3.5) of
\citet{hall2011asymptotic}. The fourth equation is due to Equation
(5.21) of \citet{hall2011asymptotic}. The fifth equation is due to
Equation (5.1) of \citet{hall2011asymptotic}. The sixth equation is
due to the weak law of law numbers and Slutsky's theorem together with
Equation (3.4), Equation (5.16), and Equation (5.18) of
\citet{hall2011asymptotic}. The seventh equality is due to the
equation below Equation (5.80) of \citet{hall2011asymptotic}.

We then compute the fourth term.
\begin{align}
&\frac{1}{m}\frac{\partial^2}{\partial\beta_0^2}
\underline{\ell}(\beta^0_0,\beta^0_1, (\sigma^2)^0)\\
 = &\frac{1}{m}\sum^m_{i=1}\hat{B}_i\exp(\hat{\mu}_i+
 \frac{1}{2}\hat{\lambda}_i) + o_P(1)\\
 \stackrel{P}{\rightarrow} & \exp(\beta_0-\frac{1}{2}
 (\sigma^2)^0)\phi(\beta_1^0).
\end{align}
This step is due to a similar computation to above. The last step is
due to weak law of large numbers and the equation below Equation
(5.80) of \citet{hall2011asymptotic}.

We now compute the third term.
\begin{align}
&\frac{1}{\sqrt{m}}\frac{\partial}{\partial\sigma^2}
\underline{\ell}(\beta^0_0,\beta^0_1, (\sigma^2)^0)\\
=&\frac{1}{\sqrt{m}}[-\frac{m}{2(\sigma^2)^0} + 
\frac{1}{2\{(\sigma^2)^0\}^2}\sum^m_{i=1}
(\hat{\mu}^2_i + \hat{\lambda}_i)]\\
=&-\sqrt{m}(\frac{1}{2(\sigma^2)^0})(1 - 
\frac{\hat{\sigma}^2}{(\sigma^2)^0})\\
=& \frac{1}{2\{(\sigma^2)^0\}^2}\sqrt{m}
(\hat{\sigma}^2 - (\sigma^2)^0)\\
\stackrel{d}{\rightarrow}&\cN(0, 2\{(\sigma^2)^0\}^2)
\frac{1}{2\{(\sigma^2)^0\}^2}.
\end{align}
The first equality is due to differentiation with respect to
$\sigma^2$. The second equality is due to Equation (5.3) of
\citet{hall2011asymptotic}. The third equality is rearranging
the terms. The fourth equation is due to Equation (3.6) of
\citet{hall2011asymptotic}.

We then compute the sixth term.
\begin{align}
&\frac{1}{m}\frac{\partial^2}{\partial(\sigma^2)^2}
\underline{\ell}(\beta^0_0,\beta^0_1, (\sigma^2)^0)\\
=&[-\frac{1}{2}\frac{1}{\{(\sigma^2)^0\}^2} + 
\frac{1}{\{(\sigma^2)^0\}^3}\frac{1}{m}\sum^m_{i=1}(\hat{\mu}^2_i + \hat{\lambda}_i)]\\
=&[-\frac{1}{2}\frac{1}{\{(\sigma^2)^0\}^2} + 
\frac{\hat{\sigma}^2}{\{(\sigma^2)^0\}^3}]+o_P(1)\\
\stackrel{P}{\rightarrow} & \frac{1}{2\{(\sigma^2)^0\}^2}.
\end{align}
This is due to a similar computation to above. The last step is due
to Equation (3.6) of \citet{hall2011asymptotic} and the weak law of
large numbers.

We next compute the second term.
\begin{align}
&\frac{1}{\sqrt{mn}}\frac{\partial}{\partial\beta_1}
\underline{\ell}(\beta^0_0,\beta^0_1, (\sigma^2)^0)\\
=& \frac{1}{\sqrt{mn}}\sum_{i=1}^m\sum_{j=1}^nX_{ij}
(Y_{ij}-\exp(\beta_0^0+\beta_1^0X_{ij}+\hat{\mu}_i+
\frac{1}{2}\hat{\lambda}_i))\\
=& \frac{1}{\sqrt{mn}}\sum_{i=1}^m\sum_{j=1}^nX_{ij}
(Y_{ij}-\exp(\hat{\beta}_0+\hat{\beta}_1X_{ij}+
\hat{\mu}_i+\frac{1}{2}\hat{\lambda}_i))\nonumber\\
&+(\beta_0^0-\hat{\beta}_0)\frac{1}{\sqrt{mn}}
\sum_{i=1}^m\sum_{j=1}^nX_{ij}(-\exp(\hat{\beta}_0+
\hat{\beta}_1X_{ij}+\hat{\mu}_i+\frac{1}{2}
\hat{\lambda}_i))\nonumber\\
&+(\beta_1^0-\hat{\beta}_1)\frac{1}{\sqrt{mn}}
\sum_{i=1}^m\sum_{j=1}^nX_{ij}^2(-\exp(\hat{\beta}_0+
\hat{\beta}_1X_{ij}+\hat{\mu}_i+\frac{1}{2}\hat{\lambda}_i)) + o_P(1)\\
=& \sqrt{mn}\frac{1}{mn}\sum_{i=1}^m\sum_{j=1}^n((\beta^0_0 - 
\hat{\beta}_0)+(\beta_1^0 - \hat{\beta}_1)X_{ij})X_{ij}
(-\exp(\hat{\beta}_0+\hat{\beta}_1X_{ij}+\hat{\mu}_i+
\frac{1}{2}\hat{\lambda}_i))\nonumber\\
& + o_P(1)\\
=& \sqrt{mn}\frac{1}{mn}\sum_{i=1}^m\sum_{j=1}^n
((\beta_1^0 - \hat{\beta}_1)(X_{ij}-\gamma(\beta_1^0))+
\bar{U})X_{ij}(-\exp(\hat{\beta}_0+\hat{\beta}_1X_{ij}+
\hat{\mu}_i+\frac{1}{2}\hat{\lambda}_i))\nonumber\\
& + o_P(1)\\
\stackrel{d}{\rightarrow}&\cN(0,\tau^2)\exp(\beta_0^0-
\frac{1}{2}\sigma^2)\phi''(\beta_1^0).
\end{align}

The first equality is due to differentiation with respect to
$\beta_1$. The second equality is due to Taylor expansion around
\gls{VFE}. The third equality is due to Equation (3.4) of
\citet{hall2011asymptotic}. The fourth equality is due to Equation
(5.16), Equation (5.18), and Equation (5.21) of
\citet{hall2011asymptotic}. The fifth equation is due to the weak law of
law numbers and Slutsky's theorem together with Equation (3.5) and
the equation below Equation (5.80) of \citet{hall2011asymptotic}.

We lastly compute the fifth term.
\begin{align}
&\frac{1}{mn}\frac{\partial^2}{\partial\beta_1^2}
\underline{\ell}(\beta^0_0,\beta^0_1, (\sigma^2)^0)\\
=&\frac{1}{mn}\sum_{i=1}^m\sum_{j=1}^nX_{ij}^2
(\exp(\beta_0^0+\beta_1^0X_{ij}+\hat{\mu}_i+
\frac{1}{2}\hat{\lambda}_i))\\
=&\frac{1}{mn}\sum_{i=1}^m\sum_{j=1}^nX_{ij}^2
(\exp(\beta_0^0+\beta_1^0X_{ij}+U_i))\\
\stackrel{P}{\rightarrow} &\exp(\beta_0^0-
\frac{1}{2}\sigma^2)\phi''(\beta_1^0).
\end{align}
This is due to a similar computation to above. The last step is due
to the weak law of large numbers.

The calculation above gives the full local asymptotic expansion of
$\ell(\beta_0, \beta_1, \sigma^2)$.

The consistent testability assumption is satisfied by the existence
of consistent estimators. This is due to Theorem 3.1 of
\citet{hall2011asymptotic}. The corollary then follows directly from
\Cref{thm:main} and \Cref{thm:estimate} in \Cref{sec:vbconsistency}.

\section{Proof of \Cref{corollary:sbm}}
\label{sec:sbmproof}

We verify the local asymptotic normality of the variational log likelihood \\
$M_n(\theta\s A) = \sup_{q(z)\in\mathcal{Q}^K}
    \int q(z)\log\frac{p(A, z\mid \nu, 
    \omega)}{q(z)}\dif z,$
where $\theta = (\nu, \omega).$

We first notice that, for any $(A, Z)$ pair,
\begin{align}
\label{eq:boundvi}
\log p(A, Z\mid \theta)\leq M_n(\theta\s A) \leq \log \int p(A, z\mid \theta)\dif z.
\end{align}
The lower bound is due to choosing the variational distribution as a point mass at $z$. The upper bound is due to the Jensen's inequality. 

From now on, assume $(A, Z)$ is generated by $\theta_0$. Lemma 3 of \citet{bickel2013asymptotic} shows that 
\begin{align}
\log p(A, Z\mid \theta_0) = \log \int p(A, z\mid \theta_0)\dif z + o_P(1),
\end{align}
Together with the above sandwich inequality, this result gives us 
\begin{align}
\label{eq:sbmvariational}
\log p(A, Z\mid \theta_0) = M_n(\theta_0\s A) + o_{P_{\theta_0}}(1).
\end{align}

Theorem 3 of \citet{bickel2013asymptotic} says that, up to label switching,
\begin{align}
M_n(\theta\s A) - \log \int p(A, z\mid \theta_0)\dif z = \log \frac{f}{f_0}(A, Z, \theta) + o_P(1).
\end{align}

Together with \Cref{eq:sbmvariational}, the above equality implies
\begin{align}
M_n(\theta\s A) - M_n(\theta_0\s A) = \log \frac{f}{f_0}(A, Z, \theta) + o_P(1).
\end{align}
Finally, Lemma 2 of \citet{bickel2013asymptotic} shows that 
\begin{align}
\log \frac{f}{f_0}(A, Z, \nu_0+\frac{t}{\sqrt{n^2\rho_n}}, 
    \omega_0+\frac{s}{\sqrt{n}}) = s^\top Y_1 + t^\top Y_2 -\frac{1}{2}s^\top\Sigma_1 s - 
    \frac{1}{2}t^\top \Sigma_2 t + o_P(1),
\end{align}
where $\Sigma_1, \Sigma_2$ are functions of $\omega_0$ and $\nu_0$, and $Y_1, Y_2$ are asymptotically normal distributed with zero mean and covariances $\Sigma_1, \Sigma_2$. Therefore, we have the \gls{LAN} expansion for the variational log likelihood $M_n(\theta\s A)$:

\begin{multline}
    M_n(\theta\s A) := \sup_{q(z)\in\mathcal{Q}^K}
    \int q(z)\log\frac{p(A, z\mid \nu_0+\frac{t}{\sqrt{n^2\rho_n}}, 
    \omega_0+\frac{s}{\sqrt{n}})}{q(z)}\dif z \\
    = \sup_{q(z)\in\mathcal{Q}^K}
    \int q(z)\log\frac{p(A, z\mid \nu_0, \omega_0)}{q(z)}\dif z + 
    s^\top Y_1 + t^\top Y_2 -\frac{1}{2}s^\top\Sigma_1 s - 
    \frac{1}{2}t^\top \Sigma_2 t + o_P(1), 
\end{multline}
for $(\nu_0, \omega_0)\in\mathcal{T}$ for compact $\mathcal{T}$ with
$\rho_n = \frac{1}{n}\mathbb{E}(\text{degree of each node})$.